\documentclass{article}

\usepackage{microtype}
\usepackage{graphicx}
\usepackage{booktabs} %

\usepackage{amsmath,amsfonts,bm}

\def\eqref#1{equation~\ref{#1}}

\def\1{\bm{1}}

\DeclareMathAlphabet{\mathsfit}{\encodingdefault}{\sfdefault}{m}{sl}
\SetMathAlphabet{\mathsfit}{bold}{\encodingdefault}{\sfdefault}{bx}{n}

\usepackage[pagebackref]{hyperref}

\usepackage{amsfonts}       %
\usepackage{amsmath}
\usepackage{amssymb}
\usepackage{nicefrac}       %
\usepackage{microtype}      %
\usepackage{enumitem}
\usepackage{xcolor}
\usepackage{subcaption}

\usepackage{floatrow}
\newfloatcommand{capbtabbox}{table}[][\FBwidth]

\usepackage[accepted]{icml2020}

\newcommand{\eg}{e.g.\ }

\icmltitlerunning{Generalization to New Actions in Reinforcement Learning}

\begin{document}

\twocolumn[
\icmltitle{Generalization to New Actions in Reinforcement Learning}
\icmlsetsymbol{equal}{*}

\begin{icmlauthorlist}
\icmlauthor{Ayush Jain}{equal,to}
\icmlauthor{Andrew Szot}{equal,to}
\icmlauthor{Joseph J. Lim}{to}
\end{icmlauthorlist}

\icmlaffiliation{to}{Department of Computer Science, University of Southern California, California, USA}

\icmlcorrespondingauthor{Ayush Jain}{ayushj@usc.edu}
\icmlcorrespondingauthor{Andrew Szot}{szot@usc.edu}

\icmlkeywords{Machine Learning, ICML, Generalization, New Actions, Reinforcement Learning}

\vskip 0.3in
]

\printAffiliationsAndNotice{\icmlEqualContribution} %

\begin{abstract}
A fundamental trait of intelligence is the ability to achieve goals in the face of novel circumstances, such as making decisions from new action choices.~\nocite{legg2007universal}
However, standard reinforcement learning assumes a fixed set of actions and requires expensive retraining when given a new action set.
To make learning agents more adaptable, we introduce the problem of zero-shot generalization to new actions.
We propose a two-stage framework where the agent first infers action representations from action information acquired separately from the task. A policy flexible to varying action sets is then trained with generalization objectives.
We benchmark generalization on sequential tasks, such as selecting from an unseen tool-set to solve physical reasoning puzzles and stacking towers with novel 3D shapes.
Videos and code are available at \url{https://sites.google.com/view/action-generalization}.
\end{abstract}

\section{Introduction}
\begin{figure}[ht!]
    \centering
    \includegraphics[width=1\linewidth]{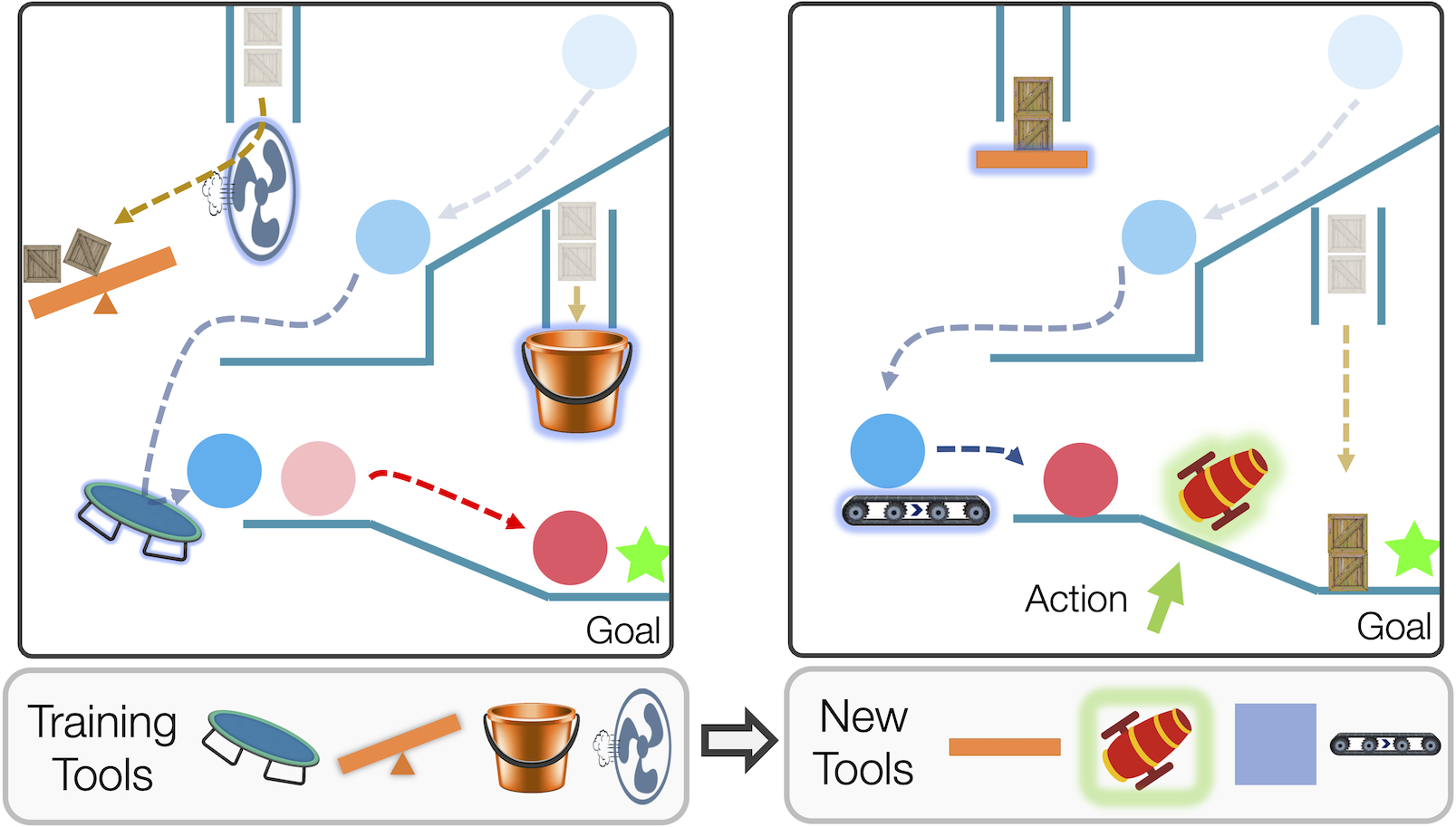}
    \caption{
    An illustration of zero-shot generalization to new actions in a sequential decision-making task, CREATE. (Left) Learning to select and place the right tools for reaching the goal. (Right) Generalizing the learned policy to a previously unseen set of tools.
    }
    \label{fig:teaser}
    \vspace{-10pt}
\end{figure}
Imagine making a salad with an unfamiliar set of tools. Since tools are characterized by their behaviors, you would first inspect the tools by interacting with them. For instance, you can observe a blade has a thin edge and infer that it is sharp. Afterward, when you need to cut vegetables for the salad, you decide to use this blade because you know sharp objects are suitable for cutting.
Like this, humans can make selections from a novel set of choices by observing the choices, inferring their properties, and finally making decisions to satisfy the requirements of the task.

From a reinforcement learning perspective, this motivates an important question of how agents can adapt to solve tasks with previously unseen actions. 
Prior work in deep reinforcement learning has explored generalization of policies over environments~\citep{cobbe2018quantifying, nichol2018gotta}, tasks~\citep{finn2017model, parisi2018continual}, and agent morphologies~\citep{wang2018nervenet, pathak19assemblies}.
However, zero-shot generalization of policies to new discrete actions has not yet been explored. The primary goal of this paper is to propose the problem of generalization to new actions. In this setup, a policy that is trained on one set of discrete actions is evaluated on its ability to solve tasks zero-shot with new actions that were unseen during training.

Addressing this problem can enable robots to solve tasks with a previously unseen toolkit, recommender systems to make suggestions from newly added products, and hierarchical reinforcement learning agents to use a newly acquired skill set.
In such applications, retraining with new actions would require prohibitively costly environment interactions. Hence, zero-shot generalization to new actions without retraining is crucial to building robust agents. 
To this end, we propose a framework and benchmark it on using new tools in the CREATE physics environment~(Figure~\ref{fig:teaser}), stacking of towers with novel 3D shapes, reaching goals with unseen navigation skills, and recommending new articles to users.

We identify three challenges faced when generalizing to new actions. 
Firstly, an agent must observe or interact with the actions to obtain data about their characteristics.
This data can be in the form of videos of a robot interacting with various tools, images of inspecting objects from different viewpoints, or state trajectories observed when executing skills.
In present work, we assume such action observations are given as input since acquiring them is domain-specific.
The second key challenge is to extract meaningful properties of the actions from the acquired action observations, which are diverse and high-dimensional.
Finally, the task-solving policy architecture must be flexible to incorporate new actions and be trained through a procedure that avoids overfitting~\citep{hawkins2004problem} to training actions.

To address these challenges, we propose a two-stage framework of representing the given actions and using them for a task. First, we employ the hierarchical variational autoencoder~\citep{edwards2016towards} to learn action representations by encoding the acquired action observations.
In the reinforcement learning stage, our proposed policy architecture computes each given action's utility using its representation and outputs a distribution.
We observe that naive training leads to overfitting to specific actions. Thus, we propose a training procedure that encourages the policy to select diverse actions during training, hence improving its generalization to unseen actions.

Our main contribution is introducing the problem of generalization to new actions. We propose four new environments to benchmark this setting. We show that our proposed two-stage framework can extract meaningful action representations and utilize them to solve tasks by making decisions from new actions. Finally, we examine the robustness of our method and show its benefits over retraining on new actions.

\section{Related Work}
\label{sec:related}

\textbf{Generalization in Reinforcement Learning}:
Our proposed problem of zero-shot generalization to new discrete action-spaces follows prior research in deep reinforcement learning (RL) for building robust agents.
Previously, state-space generalization has been used to transfer policies to new environments~\citep{cobbe2018quantifying, nichol2018gotta, packer2018assessing}, agent morphologies~\citep{wang2018nervenet, sanchez-gonzalez2018graph, pathak19assemblies}, and visual inputs for manipulation of unseen tools~\citep{fang2018learning, xie2019improvisation}.
Similarly, policies can solve new tasks by generalizing over input task-specifications, enabling agents to follow new instructions~\citep{oh2017zero}, demonstrations~\citep{xu2017neural}, and sequences of subtasks~\citep{andreas2017modular}.
Likewise, our work enables policies to adapt to previously unseen action choices.

\textbf{Unsupervised Representation Learning}:
Representation learning of high-dimensional data can make it easier to extract useful information for downstream tasks~\citep{bengio2013representation}.
Prior work has explored downstream tasks such as classification and video prediction~\citep{denton2017unsupervised}, relational reasoning through visual representation of objects~\citep{steenbrugge2018improving}, domain adaptation in RL by representing image states~\citep{higgins2017darla}, and goal representation in RL for better exploration~\citep{pmlr-v87-laversanne-finot18a} and sample efficiency~\citep{nair2018visual}.
In this paper, we leverage unsupervised representation learning of action observations to achieve generalization to new actions in the downstream RL task.

\textbf{Learning Action Representations}:
In prior work, \citet{chen2019learning, chandak2019learning, pmlr-v97-kim19a} learn a latent space of discrete actions during policy training by using forward or inverse models. \citet{tennenholtz2019natural} use expert demonstration data to extract contextual action representations. However, these approaches require a predetermined and fixed action space.
Thus, they cannot be used to infer representations of previously unseen actions.
In contrast, we learn action representations by encoding action observations acquired independent of the task, which enables zero-shot generalization to novel actions.

\textbf{Applications of Action Representations}:
Continuous representations of discrete actions have been primarily used to ease learning in large discrete action spaces~\citep{dulac2015deep, chandak2019learning} or exploiting the shared structure among actions for efficient learning and exploration~\citep{he2015deep, tennenholtz2019natural, pmlr-v97-kim19a}.
Concurrent work from~\citet{Chandak_2020} learns to predict in the space of action representations, allowing efficient finetuning when new actions are added.
In contrast, we utilize action representations learned separately, to enable zero-shot generalization to new actions in RL.

\begin{figure*}[ht!]
\centering
\includegraphics[width=0.9\linewidth]{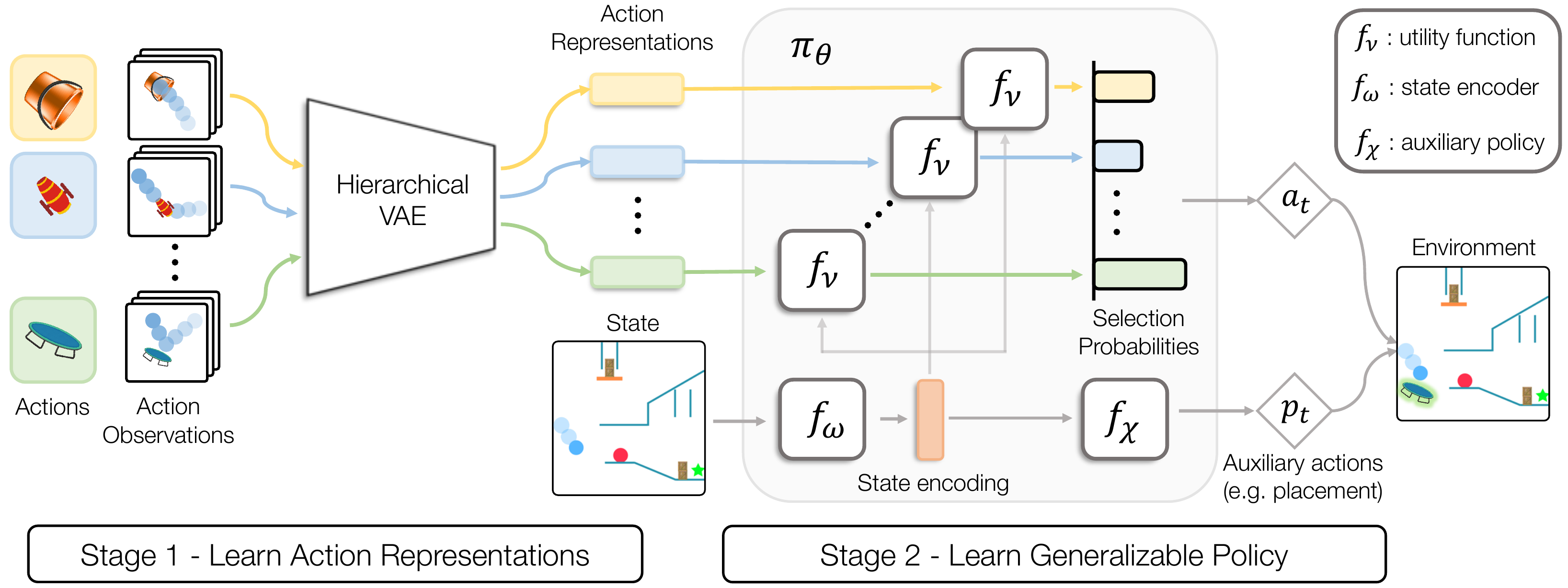}
\caption{
Two-stage framework for generalization to new actions through action representations.
(1) For each available action, a hierarchical VAE module encodes the action observations into action representations and is trained with a reconstruction objective. (2) The policy $\pi_\theta$ encodes the state with state encoder $f_\omega(s)$ and pairs it with each action representation using the utility function $f_\nu$. The utility scores are computed for each action and output to a categorical distribution. The auxiliary network takes the encoded state and outputs environment-specific auxiliary actions such as tool placement in CREATE. The policy architecture is trained with policy gradients.
}
\vspace{-10pt}
\label{fig:method}
\end{figure*}

\section{Problem Formulation}
\label{sec:problem}

In order to build robust decision-making agents, we introduce the problem setting of generalization to new actions. A policy that is trained on one set of actions is evaluated on its ability to utilize unseen actions without additional retraining.
Such zero-shot transfer requires additional input that can illustrate the general characteristics of the actions.
Our insight is that action choices, such as tools, are characterized by their general behaviors.
Therefore, we record a collection of an action's behavior in diverse settings in a separate environment to serve as action observations.
The action information extracted from these observations can then be used by the downstream task policy to make decisions. For instance, videos of an unseen blade interacting with various objects can be used to infer that the blade is sharp. If the downstream task is cutting, an agent can then reason to select this blade due to its sharpness.

\subsection{Reinforcement Learning}
We consider the problem family of episodic Markov Decision Processes (MDPs) with discrete action spaces. MDPs are defined by a tuple $\{\mathcal{S}, \mathcal{A}, \mathcal{T}, \mathcal{R}, \gamma\}$ of states, actions, transition probability, reward function, and discount factor.
At each time step $t$ in an episode, the agent receives a state observation $s_t \in \mathcal{S}$ from the environment and responds with an action $a_t \in \mathcal{A}$. This results in a state transition to $s_{t+1}$ and a state-conditioned reward $\mathcal{R}(s_{t+1})$. The objective of the agent is to maximize the expected discounted reward $R = \sum _{t=1}^{T}\gamma ^{t-1} \mathcal{R}(s_t) $ in an episode of length $T$.

\subsection{Generalization to New Actions}
The setting of generalization to new actions consists of two phases: training and evaluation.
During training, the agent learns to solve tasks with a given set of actions $\mathbb{A} = \{a_{1}, \dots, a_{N} \}$. During each evaluation episode, the trained agent is evaluated on a new action set $\mathcal{A}$ sampled from a set of unseen actions $\mathbb{A}'$.
The objective is to learn a policy $\pi(a | s, \mathcal{A})$, which maximizes the expected discounted reward using any given action set $\mathcal{A} \subset \mathbb{A}'$,
\setlength{\belowdisplayskip}{10pt} \setlength{\belowdisplayshortskip}{10pt}
\setlength{\abovedisplayskip}{10pt} \setlength{\abovedisplayshortskip}{10pt}
\begin{equation}
\label{eq:problem_objective}
    R = \mathbb{E}_{\mathcal{A} \subset \mathbb{A}',\; a\sim\pi(a | s, \mathcal{A})}
    \;\big[
    \textstyle\sum_{t=1}^{T} \gamma^{t-1} \mathcal{R}(s_t)
    \big].
\end{equation}
For each action $a \in \mathbb{A} \cup \mathbb{A}'$, the set of acquired action observations is denoted with $\mathcal{O} = \{o_{1}, \dots, o_{n} \}$. Here, each $o_j \in \mathcal{O}$ is an observation for the action like a state-trajectory, a video, or an image, indicating the action's behavior. For the set of training actions $\mathbb{A}$, we denote the set of associated actions observations as $\mathbb{O} = \{\mathcal{O}_{1}, \dots, \mathcal{O}_{N} \}$. 

\section{Approach}
\label{sec:approach}

Our approach for generalization to new actions is based on the intuition that humans make decisions from new options by exploiting prior knowledge about the options~\citep{gershman2015novelty}.
First, we infer the properties of each action from the action observations given as prior knowledge. Second, a policy learns to make decisions based on these inferred action properties. When a new action set is given, their properties are inferred and exploited by the policy to solve the task. Formally, we propose a two-stage framework:
\begin{enumerate}[leftmargin=*, parsep=5pt, itemsep=0pt, topsep=0pt]
    \item \textbf{Learning Action Representations}:
    We use unsupervised representation learning to encode each set of action observations into an action representation. This representation expresses the latent action properties present in the set of diverse observations (Section~\ref{sec:embedder}).
    \item \textbf{Learning Generalizable Policy}:
    We propose a flexible policy architecture to incorporate action representations as inputs, which can be trained through RL (Section~\ref{sec:policy}). We provide a training procedure to control overfitting to the training action set, making the policy generalize better to unseen actions (Section~\ref{sec:procedure}).
\end{enumerate}

\subsection{Unsupervised Learning of Action Representations}
\label{sec:embedder}
Our goal is to encode each set of action observations into an action representation that can be used by a policy to make decisions in a task.
The main challenge is to extract the shared statistics of the action's behavior from high-dimensional and diverse observations.

To address this, we employ the hierarchical variational autoencoder (HVAE) by~\citet{edwards2016towards}. HVAE first summarizes the entire set of an action's observations into a single action latent. This action latent then conditions the encoding and reconstruction of each constituent observation through a conditional VAE.
Such hierarchical conditioning ensures that the observations for the same action are organized together in the latent space. Furthermore, the action latent sufficiently encodes the diverse statistics of the action. Therefore, this action latent is used as the action's representation in the downstream RL task (Figure~\ref{fig:method}).

Formally, for each training action $a_i \in \mathbb{A}$, HVAE encodes its associated action observations $\mathcal{O}_i \in \mathbb{O}$ into a representation $c_i$ by mean-pooling over the individual observations $o_{i,j} \in \mathcal{O}_i$. We refer to this action encoder as the action representation module $q_\phi(c_i | \mathcal{O}_i)$.
The action latent $c_i$ sampled from the action encoder is used to condition the encoders $q_\psi(z_{i,j} | o_{i,j}, c_i)$ and decoders $p(o_{i,j} | z_{i,j}, c_i)$ for each individual observation $o_{i,j} \in \mathcal{O}_i$.
The entire HVAE framework is trained with reconstruction loss across the individual observations, along with KL-divergence regularization of encoders $q_\phi$ and $q_\psi$ with their respective prior distributions $p(c)$ and $p(z|c_i)$. For additional details on HVAE, refer to Appendix~\ref{supp:training:network:HVAE} and \citet{edwards2016towards}. The final training objective requires maximizing the ELBO:
\setlength{\belowdisplayskip}{10pt} \setlength{\belowdisplayshortskip}{10pt}
\setlength{\abovedisplayskip}{10pt} \setlength{\abovedisplayshortskip}{10pt}
\begin{equation}
\begin{split}
\label{eq:elbo}
\mathcal{L} =
\sum_{\mathcal{O} \in \mathbb{O}}
\Big[
\mathbb{E}_{q_\phi(c|\mathcal{O})}
\Big[\sum_{o \in \mathcal{O}} {\mathbb{E}}_{q_\psi(z|o, c)}
~{\log p} (o|z, c)\\
- D_{KL}(q_\psi || p(z|c)) \Big]
- D_{KL}(q_\phi ||  p(c))
\Big].
\end{split}
\end{equation}

For action observations consisting of sequential data, $o = \{x_0, \dots, x_m\}$ like state trajectories or videos, we augment HVAE to extract temporally extended behaviors of actions.
We accomplish this by incorporating insights from trajectory autoencoders~\citep{NIPS2017_7116, co2018self} in HVAE.
Bi-LSTM~\citep{schuster1997bidirectional} is used in the encoders and LSTM is used as the decoder $p(x_1, \dots, x_m \lvert z, c, x_0)$ to reconstruct the trajectory given the initial state $x_0$.
Explicitly for video observations, we also incorporated temporal skip connections~\citep{ebert2017self} by predicting an extra mask channel to balance contributions from the predicted and first frame of the video.

We set the representation for an action as the mean of the inferred distribution $q_\phi(c_i | \mathcal{O}_i)$ as done in~\citet{higgins2017beta, steenbrugge2018improving}. 

\subsection{Adaptable Policy Architecture}
\label{sec:policy}
To enable decision-making with new actions, we develop a policy architecture that can adapt to any available action set~$\mathcal{A}$ by taking the list of action representations as input. Since the action representations are learned independently of the downstream task, a task-solving policy must learn to extract the relevant task-specific knowledge.

\setlength{\textfloatsep}{0pt}
\begin{algorithm}[t]
    \caption{Two-stage Training Framework}
    \label{alg:train}
    \begin{algorithmic}[1]
        \STATE \textbf{Inputs:} Training actions $ \mathbb{A}$,
        action observations $ \mathbb{O} $
        \STATE Randomly initialize HVAE and policy parameters
        \FOR{epoch = 1, 2, \dots}
            \STATE Sample batch of action observations $\mathcal{O}_i \sim \mathbb{O}$
            \STATE Train HVAE parameters with gradient ascent on Eq.~\ref{eq:elbo}
        \ENDFOR
        \STATE Infer action representations: $ c_{i} = q_{\phi}^\mu(\mathcal{O}_i), \forall a_i \in \mathbb{A} $ 
        \FOR{iteration = 1, 2, \dots}
            \WHILE{episode not done}
                \STATE Subsample action set $\mathcal{A} \subset \mathbb{A}$ of size $m$
                \STATE Sample action $ a_t \sim \pi_\theta (s, \mathcal{A})$ using Eq.~\ref{eq:policy}
                \STATE $s_{t+1}, r_t \leftarrow$ ENV$(s_t, a_t)$
                \STATE Store experience $(s_t, a_t, s_{t+1}, r_t)$ in replay buffer
            \ENDWHILE
            \STATE Update and save policy $ \theta $ using PPO on Eq.~\ref{eq:final_obj}
        \ENDFOR
        \STATE Select $\theta$ with best validation performance
    \end{algorithmic}
\end{algorithm}
\setlength{\textfloatsep}{10pt}
\begin{algorithm}[t]
    \caption{Generalization to New Actions}
    \label{alg:inference}
    \begin{algorithmic}[1]
        \STATE \textbf{Inputs:} New actions $\mathcal{A}= \{ a_{1}, \dots a_{M} \}$, observations $\{ \mathcal{O}_{1}, \dots \mathcal{O}_{M} \} $. Trained networks $q_\phi$ and $\pi_\theta$
        \STATE Infer action representations: $ c_{i} = q_{\phi}^\mu(\mathcal{O}_i), \forall a_i \in \mathcal{A} $
        \WHILE{not done}
             \STATE Sample action $ a_t \sim \pi_\theta (s, \mathcal{A})$ using Eq.~\ref{eq:policy}
            \STATE $s_{t+1}, r_t \leftarrow$ ENV$(s_t, a_t)$
        \ENDWHILE
    \end{algorithmic}
\end{algorithm}

The policy $\pi(a | s, \mathcal{A})$ receives a set of available actions~$\mathcal{A}=\{a_1, \dots, a_k \}$ as input, along with the action representations$\{c_1, \dots, c_k \}$. As shown in Figure~\ref{fig:method}, the policy architecture starts with a state encoder $f_\omega$. The utility function $f_\nu$ is applied to each given action's representation $c_i$ and the encoded state~$f_\omega(s)$ (Eq.~\ref{eq:policy}). The utility function estimates the score of an action at the current state, through its action representation, just like a Q-function~\citep{watkins1992q}. Action utility scores are converted into a probability distribution through a softmax function:
\setlength{\belowdisplayskip}{10pt} \setlength{\belowdisplayshortskip}{10pt}
\setlength{\abovedisplayskip}{10pt} \setlength{\abovedisplayshortskip}{10pt}
\begin{equation}
\label{eq:policy}
    \pi(a_i | s, \mathcal{A}) = \frac{e^{f_\nu[c_i, f_\omega(s)]}}{
    \sum_{j=1}^{k} e^{f_\nu[c_j, f_\omega(s)]}}.
\end{equation}
In many physical environments, the choice of a discrete action is associated with auxiliary parameterizations, such as the intended position of tool usage or a binary variable to determine episode termination. We incorporate such hybrid action spaces~\citep{hausknecht2015deep}, through an auxiliary network $f_\chi$, which takes the encoded state and outputs a distribution over the auxiliary actions~\footnote{Alternatively, the auxiliary network can take the discrete selection as input as tested in Appendix~\ref{supp:exp:auxiliary_alternate}}.
An environment action is taken by sampling the auxiliary action from this distribution and the discrete action from Eq.~\ref{eq:policy}. The policy parameters $\theta = \{\nu, \omega, \chi \}$ are trained end-to-end using policy gradients~\citep{sutton2000policy}.

\subsection{Generalization Objective and Training Procedure}
\label{sec:procedure}
Our final objective is to find policy parameters $\theta$ to maximize reward on held-out action sets $\mathcal{A} \subset \mathbb{A}'$ (Eq.~\ref{eq:problem_objective}), while being trained on a limited set of actions $\mathbb{A}$.
We study this generalization problem based on statistical learning theory~\citep{vapnik1998statistical, vapnik2013nature} in supervised learning.
Particularly, generalization of machine learning models is expected when their training inputs are independent and identically distributed~\citep{bousquet2003introduction}. However, in RL, a policy typically acts in the environment to collect its own training data.
Thus when a policy overexploits a specific subset of the training actions, this skews the policy training data towards those actions. 
To avoid this form of overfitting and be robust to diverse new action sets, we propose the following regularizations to approximate i.i.d. training:
\begin{itemize}[leftmargin=*, parsep=5pt, itemsep=0pt, topsep=0pt]
\item \textbf{Subsampled action spaces}:
To limit the actions available in each episode of training, we randomly subsample action sets, $\mathcal{A} \subset \mathbb{A}$ of size $m$, a hyperparameter.
This avoids overfitting to any specific actions by forcing the policy to solve the task with diverse action sets.
\item \textbf{Maximum entropy regularization}:
We further diversify the policy's actions during training using the maximum entropy objective~\citep{ziebart2008maximum}. We add the entropy of the policy $\mathcal{H}[\pi_\theta(a|s)]$ to the RL objective with a hyperparameter weighting $\beta$.
While this objective has been widely used for exploration, we find it useful to enable generalization to new actions.
\item \textbf{Validation-based model selection}: During training, the models are evaluated on held-out validation sets of actions, and the best performing model is selected. Just like supervised learning, this helps to avoid overfitting the policy during training. Note that the validation set is also used to tune hyperparameters such as entropy coefficient $\beta$ and subsampled action set size $m$. There is no overlap between test and validation sets, hence the test actions are still completely unseen at evaluation.
\end{itemize}
The final policy training objective is:
\setlength{\belowdisplayskip}{10pt} \setlength{\belowdisplayshortskip}{10pt}
\setlength{\abovedisplayskip}{10pt} \setlength{\abovedisplayshortskip}{10pt}
\begin{equation}
\label{eq:final_obj}
    \max_\theta {\textrm{{$\mathbb{E}$}}}_{
    \mathcal{A} \subset \mathbb{A},
    a \sim \pi_\theta(.|s, \mathcal{A})}
    [
    R(s) + 
    \beta \mathcal{H}[\pi_\theta(a|s, \mathcal{A})]
    ].
\end{equation}
The training procedure is described in Algorithm~\ref{alg:train}. The HVAE is trained using RAdam optimizer~\citep{liu2019radam}, and we use PPO~\citep{PPO} to train the policy with Adam Optimizer~\citep{kingma2014adam}. Additional implementation and experimental details, including the hyperparameters searched, are provided in Appendix~\ref{supp:training}. The inference process is described in Algorithm~\ref{alg:inference}.
When given a new set of actions, we can infer the action representations with the trained HVAE module. The policy can also generalize to utilize these actions since it has learned to map a list of action representations to an action probability distribution.

\begin{figure*}[ht]
\centering
\includegraphics[width=0.85\linewidth]{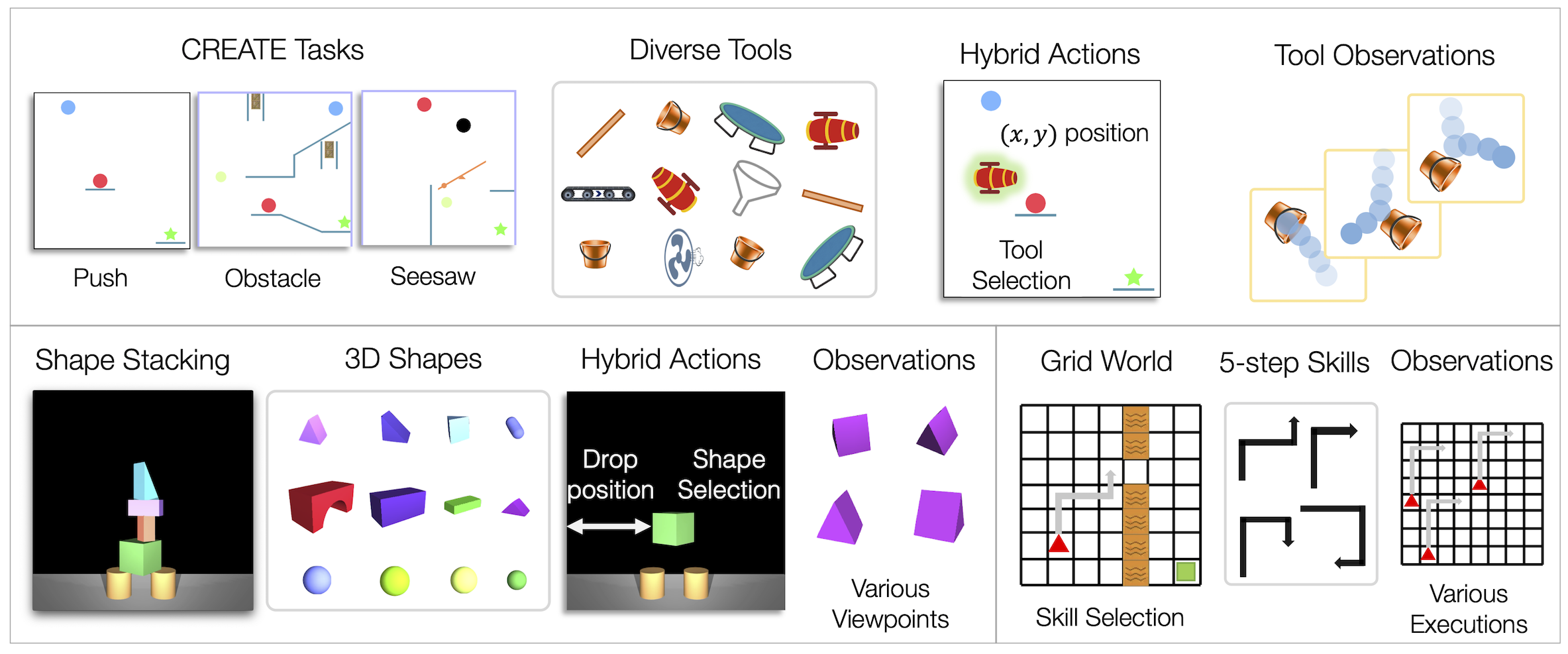}
\vspace{-10pt}
\caption{
Benchmark environments for evaluating generalization to new actions. (Top) In CREATE, an agent selects and places various tools to move the red ball to the goal. Other moving objects can serve as help or obstacles. Some tasks also have subgoals to help with exploration (Appendix~\ref{supp:exp:no_subgoal} shows results with no subgoal rewards). The tool observations consist of trajectories of a test ball interacting with the tool. (Left) In Shape Stacking, an agent selects and places 3D shapes to stack a tower. The shape observations are images of the shape from different viewpoints. (Right) In Grid World, an agent reaches the goal by choosing from 5-step navigation skills. The skill observations are collected on an empty grid in the form of agent trajectories resulting from skill execution from random locations.
}
\label{fig:envs}
\end{figure*}

\section{Experimental Setup}
\label{setup}

\subsection{Environments}
\label{environments}
We propose four sequential decision-making environments with diverse actions to evaluate and benchmark the proposed problem of generalization to new actions.
These test the action representation learning method on various types of action observations.
The long-horizon nature of the environments presents a challenge to use new actions correctly to solve the given tasks consistently. Figure~\ref{fig:envs} provides an overview of the task, types of actions, and action observations in three environments. In each environment, the train-test-validation split is approximately 50-25-25\%. Complete details on each environment, action observations, and train-validation-test splits can be found in Appendix~\ref{supp:sec:env_details}.

\subsubsection{Grid World}
In the Grid world environment~\citep{gym_minigrid}, an agent navigates a 2D lava maze to reach a goal using predefined skills.
Each skill is composed of a 5-length sequence of left, right, up or down movement.
The total number of available skills is $4^5$. Action observations consist of state sequences of an agent observed by applying the skill in an empty grid. 
This environment acts as a simple demonstration of generalization to unseen skill sets.

\subsubsection{Recommender System}
The Recommender System environment~\citep{rohde2018recogym} simulates users responding to product recommendations. Every episode, the agent makes a series of recommendations for a new user to maximize their click-through rate (CTR). With a total of 10,000 products as actions, the agent is evaluated on how well it can recommend previously unseen products to users. The environment specifies predefined action representations. Thus we only evaluate our policy framework on it, not the action encoder.

\subsubsection{CREATE}
We develop the Chain REAction Tool Environment (CREATE) as a challenging benchmark to test generalization to new actions\footnote{CREATE environment: \url{https://clvrai.com/create}}.
It is a physics-based puzzle where the agent must place tools in real-time to manipulate a specified ball's trajectory to reach a goal position (Figure~\ref{fig:envs}). The environment features 12 different tasks and 2,111 distinct tools. Moreover, it tests physical reasoning since every action involves selecting a tool and predicting the 2D placement for it, making it a hybrid action-space environment. Action observations for a tool consist of a test ball's trajectories interacting with the tool from various directions and speeds. CREATE tasks evaluate the ability to understand complex functionalities of unseen tools and utilize them for various tasks. 
We benchmark our framework on all 12 CREATE tasks with the extended results in Appendix~\ref{supp:exp:additional_create}.

\subsubsection{Shape Stacking}
We develop a MuJoCo-based~\citep{todorov2012mujoco} Shape Stacking environment, where the agent drops blocks of different shapes to build a tall and stable tower.
Like in CREATE, the discrete selection of shape is parameterized by the coordinates of where to place the selected shape and a binary action to decide whether to stop stacking. 
This environment evaluates the ability to use unseen complex 3D shapes in a long horizon task and contains 810 shapes.

\subsection{Experiment Procedure}
\label{experiment_procedure}
We perform the following procedure for each action generalization experiment\footnote{Complete code available at \url{https://github.com/clvrai/new-actions-rl}}.
\begin{enumerate}[leftmargin=*, noitemsep, parsep=3pt, topsep=0pt]
    \item \textit{Collect action observations} for all the actions using a supplemental play environment that is task-independent.
    \item \textit{Split the actions} into train, validation, and test sets.
    \item \textit{Train HVAE} on the train action set by autoencoding the collected action observations.
    \item \textit{Infer action representations} for all the actions using the trained HVAE encoder on their action observations.
    \item \textit{Train policy} on the task environment with RL. In each episode, an action set is randomly sampled from the train actions. The policy acts by using the list of inferred action representations as input.
    \item \textbf{Evaluation}: In each episode, an action set is subsampled from the test (or validation) action set. The trained policy uses the inferred representations of these actions to act in the environment zero-shot. The performance metric (\eg success rate) is averaged over multiple such episodes.
    \begin{enumerate}[leftmargin=*, noitemsep, parsep=3pt, topsep=0pt]
        \item Perform hyperparameter tuning and model selection by evaluating on the \textit{validation action set}.
        \item Report final performance on the \textit{test action set}.
    \end{enumerate}
\end{enumerate}

\begin{figure*}[ht]
\centering
\includegraphics[width=0.8\linewidth]{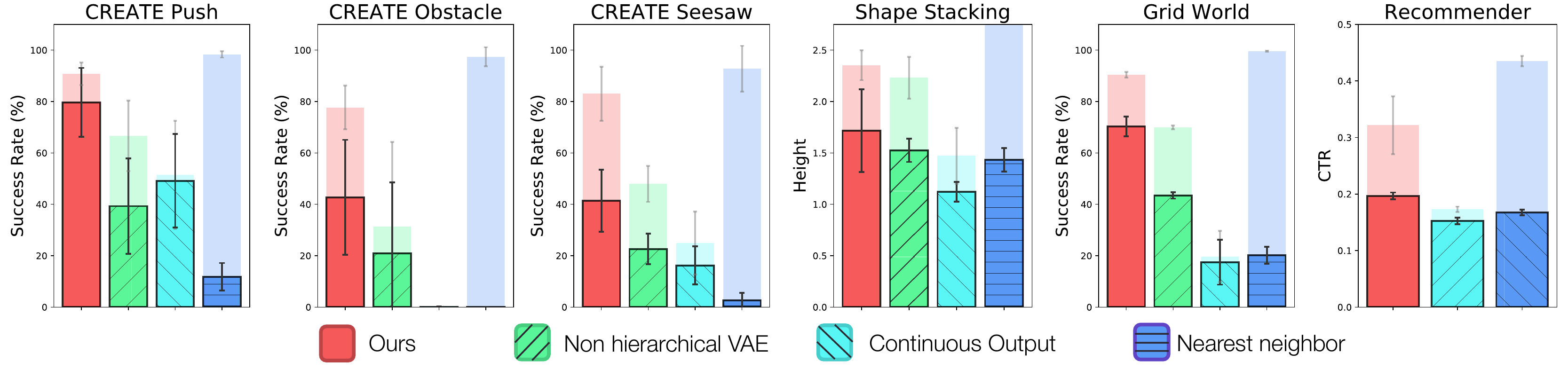}
\vspace{-15pt}
\caption{
Comparison against baseline action representation and policy architectures on 6 environments, 3 of which are CREATE tasks. The solid bar denotes the test performance and the transparent bar the training performance, to observe the generalization gap. The results are averaged over 5000 episodes across 5 random seeds, and the error bars indicate the standard deviation (8 seeds for Grid World). All learning curves are present in Figure~\ref{supp:fig:train_curves}. Results on 9 additional CREATE tasks can be found in Appendix~\ref{supp:exp:additional_create}.
}
\label{fig:baselines}
\end{figure*}

\subsection{Baselines}
\label{sec:baselines}
We validate the design choices of the proposed action encoder and policy architecture. For action encoder, we compare with a policy using action representations from a non-hierarchical encoder.
For policy architecture, we consider alternatives that select actions using distances in the action representation space instead of learning a utility function.
\begin{itemize}[leftmargin=*, parsep=5pt, itemsep=0pt, topsep=0pt]
    \item \textbf{Non-hierarchical VAE}:
    A flat VAE is trained over the individual action observations. An action's representation is taken as the mean of encodings of the constituent action observations.
    \item \textbf{Continuous-output}: The policy architecture outputs a continuous vector in the action representation space, following~\citet{dulac2015deep}. From any given action set, the action closest to this output is selected.
    \item \textbf{Nearest-Neighbor}: A standard discrete action policy is trained. The representation of this policy's output action is used to select the nearest neighbor from new actions.
\end{itemize}

\subsection{Ablations}
\label{sec:ablations}
We individually ablate the two proposed regularizations:
\begin{itemize}[leftmargin=*, parsep=5pt, itemsep=0pt, topsep=0pt]
    \item
    \textbf{Ours without subsampling}: Trained over the entire set of training actions without any action space sampling.
    \item \textbf{Ours without entropy}: Trained without entropy regularization, by setting the entropy coefficient to zero.
\end{itemize}

\section{Results and Analysis}
\label{sec:experiments}
Our experiments aim to answer the following questions about the proposed problem and framework: 
(1) Can the HVAE extract meaningful action characteristics from the action observations?
(2) What are the contributions of the proposed action encoder, policy architecture, and regularizations for generalization to new actions?
(3) How well does our framework generalize to varying difficulties of test actions and types of action observations?
(4) How inefficient is finetuning to a new action space as compared to zero-shot generalization?

\subsection{Visualization of Inferred Action Representations}
\begin{figure}[h]
    \centering
    \includegraphics[width=\textwidth]{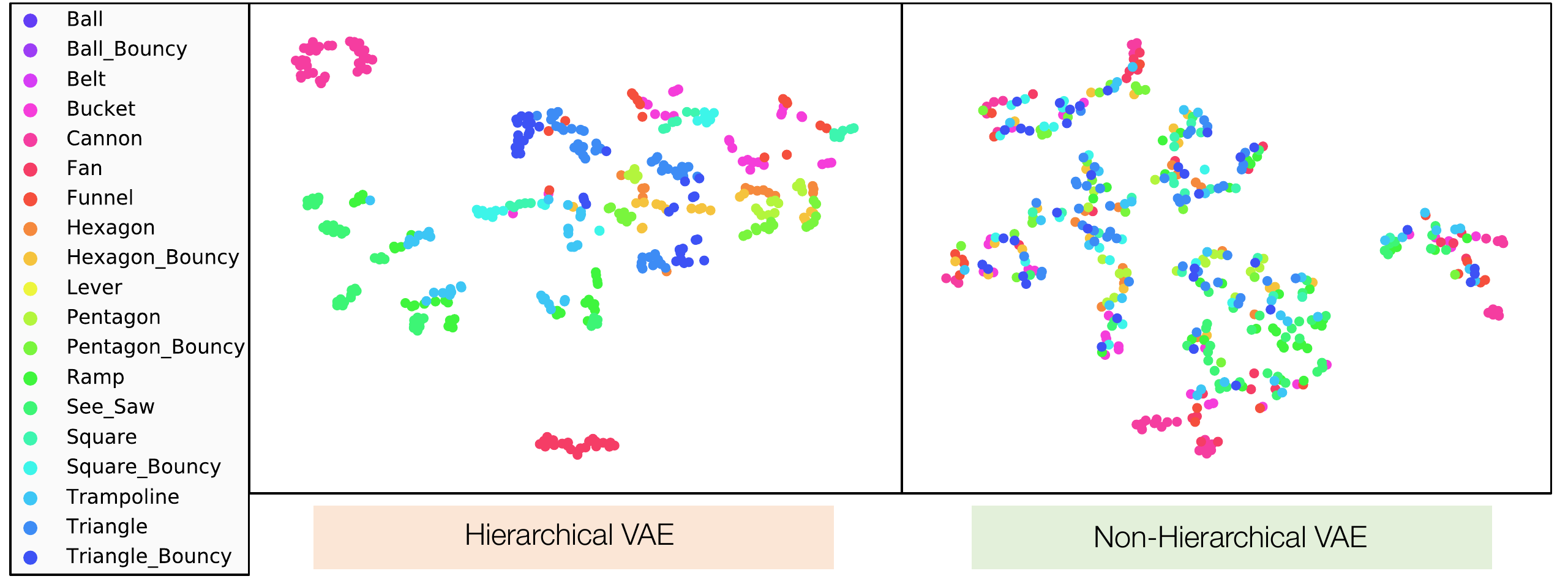}
    \vspace{-20pt}
    \caption{
        t-SNE visualization of action representations for held-out tools in CREATE inferred using a trained HVAE (left) and a VAE (right). The color indicates the tool class (\eg cannons, buckets). The HVAE encoder learns to organize semantically similar tools together, in contrast to the flat VAE, which shows less structure.
    }
    \label{fig:embedding}
\end{figure}
To investigate if the HVAE can extract important characteristics from
observations of new actions, we visualize the inferred action representations for unseen CREATE tools. In Figure~\ref{fig:embedding}, we observe that tools from the same class are clustered together in the HVAE representations.
Whereas in the absence of hierarchy, the action representations are less organized.
This shows that encoding action observations independently, and averaging them to obtain a representation can result in the loss of semantic information, such as the tool's class. In contrast, hierarchical conditioning on action representation enforces various constituent observations to be encoded together. This helps to model the diverse statistics of the action's observations into its representation.

\begin{figure*}[ht]
\centering
\includegraphics[width=0.8\linewidth]{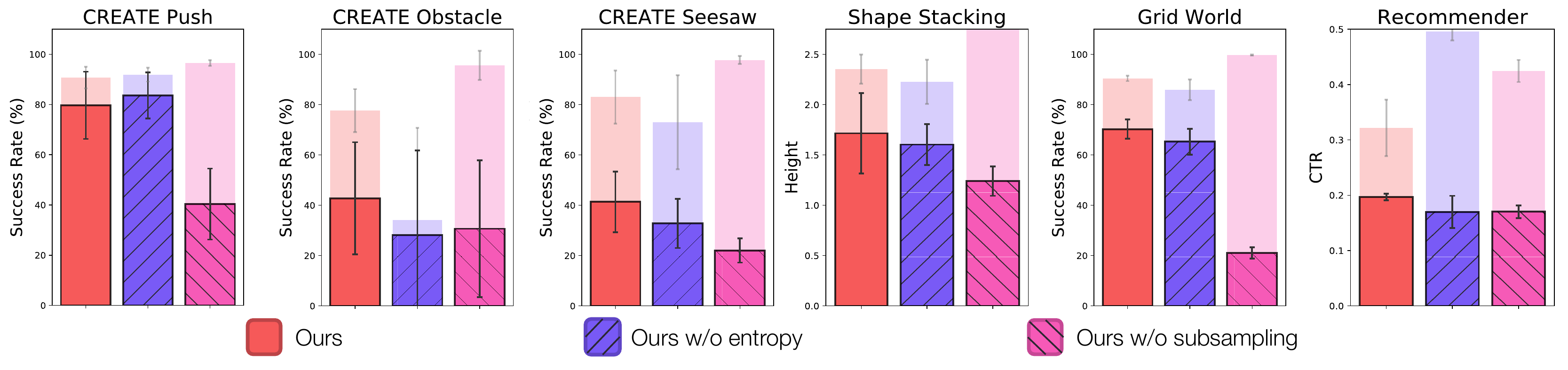}
\vspace{-15pt}
\caption{
Analyzing the importance of the proposed action space subsampling and entropy regularization in our method. The training and evaluation details are the same as Figure~\ref{fig:baselines}.
}
\label{fig:ablations}
\end{figure*}
\begin{figure*}[ht]
    \vspace{-6pt}
    \centering
    \begin{subfigure}[t]{0.29\textwidth}
    	\centering
        \includegraphics[width=0.6\linewidth]{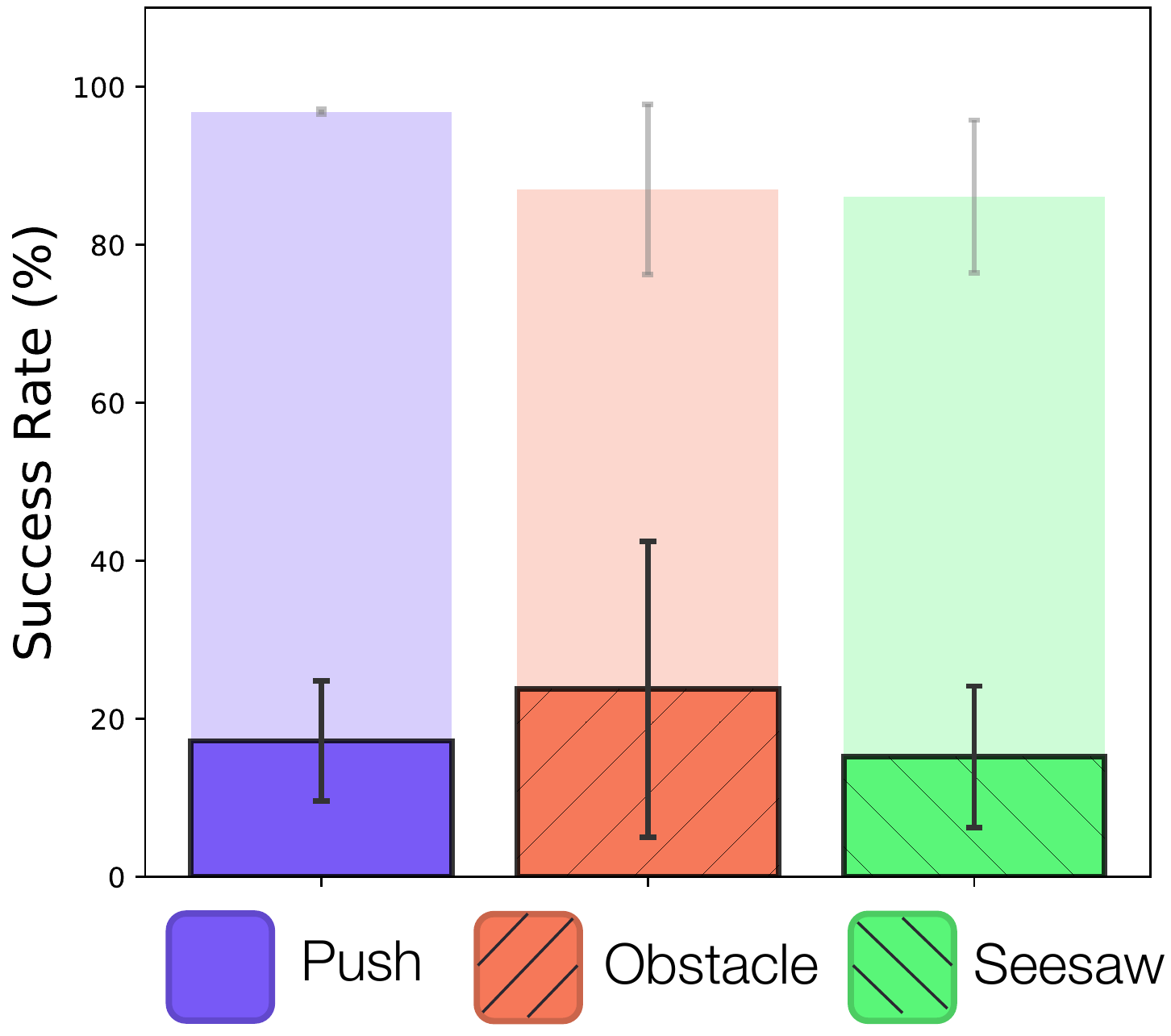}
        \caption{Unseen tool classes in CREATE}
        \label{fig:create_full}
    \end{subfigure}
    \begin{subfigure}[t]{0.66\textwidth}
    	\centering
    	\includegraphics[width=\textwidth]{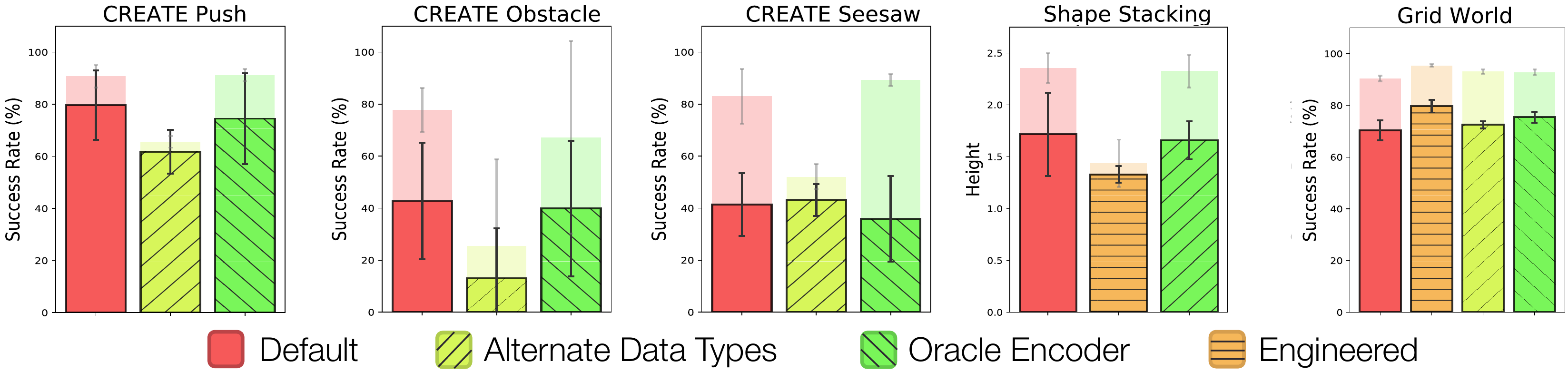}
    	\caption{Alternate action representations}
    	\label{fig:alternate_representations}
    \end{subfigure}
    \vspace{-15pt}
    \caption{
        Additional analyses.
        (a) Our method achieves decent performance on out-of-distribution tools in 3 CREATE tasks, but the generalization gap is more pronounced.
        (b) Various action representations can be successfully used with our policy architecture.
    }
    \label{fig:full_oracle}
\end{figure*}

\subsection{Results and Comparisons}
\subsubsection{Baselines}
Figure~\ref{fig:baselines} shows that our framework outperforms the baselines (Section~\ref{sec:baselines}) in zero-shot generalization to new actions on six tasks.
The non-hierarchical VAE baseline has lower policy performance in both training and testing. This shows that HVAE extracts better representations from action observations that facilitate easier policy learning.

The continuous-output baseline suffers in training as well as testing performance. This is likely due to the complex task of indirect action selection.
The distance metric used to find the closest action does not directly correspond to the task relevance.
Therefore the policy network must learn to adjust its continuous output, such that the desired discrete action ends up closest to it.
Our method alleviates this through the utility function, which first extracts task-relevant features to enable an appropriate action decision.
The nearest-neighbor baseline achieves high training performance since it is merely discrete-action RL with a fixed action set. However, at test time, the simple nearest-neighbor in action representation space does not correspond to the actions' task-relevance. This results in poor generalization performance.

\subsubsection{Ablations}
Figure~\ref{fig:ablations} assesses the contribution of the proposed regularizations to avoid overfitting to training actions. Entropy regularization usually leads to better training and test performance due to better exploration. In the recommender environment, the generalization gap is more substantial without entropy regularization. Without any incentive to diversify, the policy achieved high training performance by overfitting to certain products. We observe a similar effect in the absence of action subsampling across all tasks. It achieves a higher training performance, due to the ease of training in non-varying action space. However, its generalization performance is weak because it is easy to overfit when the policy has access to all the actions during training.

\subsection{Analyzing the Limits of Generalization}
\subsubsection{Generalization to Unseen Action Classes}

Our method is expected to generalize when new actions are within the distribution of those seen during training. However, what happens when we test our approach on completely unseen action classes? Generalization is still expected because the characteristic action observations enable the representation of actions in the same space. Figure~\ref{fig:create_full} evaluates our approach on held-out tool classes in the CREATE environment. Some tool classes like trampolines and cannons are only seen during training, whereas others like fans and conveyor belts are only used during testing. While the generalization gap is more substantial than before, we still observe reasonable task success across the 3 CREATE tasks. The performance can be further improved by increasing the size and diversity of training actions. Appendix~\ref{supp:exp:additional_shape} shows a similar experiment on Shape Stacking.

\begin{figure*}[!ht]
    \centering
    \includegraphics[width=0.9\linewidth]{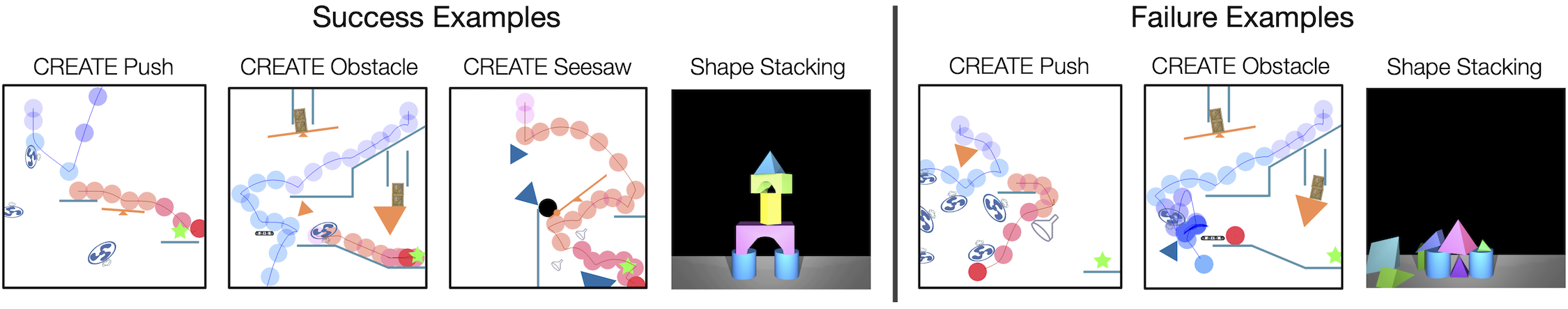}
    \vspace{-15pt}
    \caption{Evaluation results showing the trajectories of objects in CREATE and the final tower in Shape Stacking.
    Our framework is generally able to infer the dynamic properties of tools and geometry of shapes and subsequently use them to make the right decisions.}
    \label{fig:qual_results}
\end{figure*}

\subsubsection{Alternate Action Representations}
In Figure~\ref{fig:alternate_representations}, we study policy performance for various action representations. See Appendix~\ref{supp:sec:action_rep_vis} for t-SNE visualizations.
\begin{itemize}[leftmargin=*, parsep=5pt, itemsep=0pt, topsep=0pt]
\item
\textbf{Alternate Data Types} of action observations are used to learn representations. For CREATE, we use video data instead of the state trajectory of the test ball (see Figure~\ref{fig:envs}). For Grid World, we test with a one-hot vector of agent location instead of $(x,y)$ coordinates. The policy performance using these representations is comparable to the default. This shows that HVAE is suitable for high-dimensional action observations, such as videos.

\item
\textbf{Oracle} HVAE is used to get representations by training on the test actions. The performance difference between default and oracle HVAE is negligible.
This shows that HVAE generalizes well to unseen action observations.

\item \textbf{Hand-Engineered} action representations are used for Stacking and Grid World, by exploiting ground-truth information about the actions. In Stacking, HVAE outperforms these representations, since it is hard to specify the information about shape geometry manually. In contrast, it is easy to specify the complete skill in Grid World. Nevertheless, HVAE representations perform comparably.

\end{itemize}

\subsubsection{Varying the Difficulty of Generalization}
\begin{figure}[ht]
    \vspace{-6pt}
    \centering
    \begin{subfigure}[t]{0.32\textwidth}
    	\centering
    	\includegraphics[width=\textwidth]{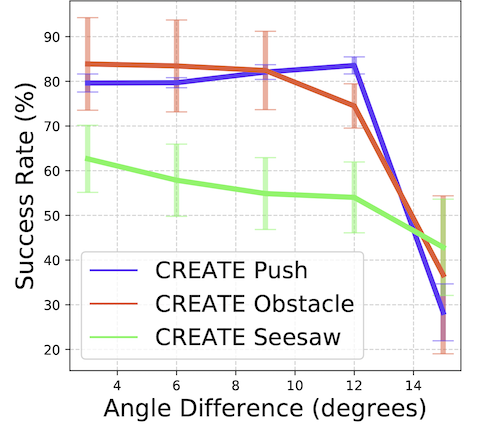}
    	\caption{Tool Angle}
        \label{fig:analysis_gran}
    \end{subfigure}
    \begin{subfigure}[t]{0.32\textwidth}
    	\centering
    	\includegraphics[width=\textwidth]{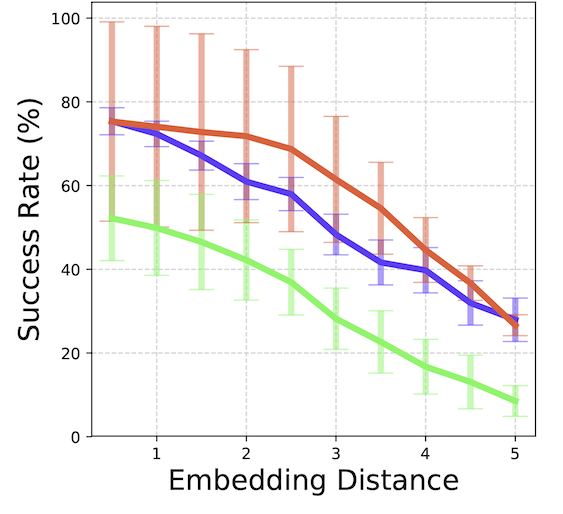}
    	\caption{Tool Embedding}
    	\label{fig:analysis_emb_dist}	
    \end{subfigure}
    \begin{subfigure}[t]{0.32\textwidth}
    	\centering
    	\includegraphics[width=\textwidth]{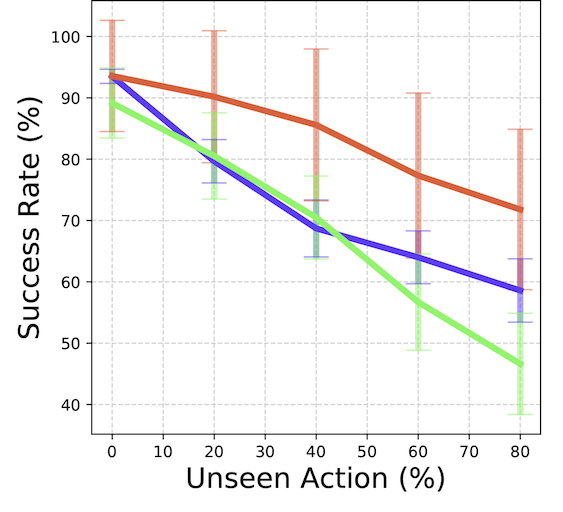}
    	\caption{Unseen Ratio}
      \label{fig:analysis_ratio}	
    \end{subfigure}
    \vspace{-15pt}
    \caption{
        Varying the test action space. An increasing x-axis corresponds to more difficult generalization conditions. Each value plotted is the average test performance over 5 random seeds with the error bar corresponding to the standard deviation.
    }
    \label{fig:result_figure}
\end{figure}
Figure~\ref{fig:result_figure} shows a detailed study of generalization on various degrees of differences between the train and test actions in 3 CREATE tasks. We vary the following parameters:
\begin{enumerate}[leftmargin=*, parsep=5pt, itemsep=0pt, topsep=0pt,label=\alph*)]
\item \textbf{Tool Angle}: Each sampled test tool is at least $\theta$ degrees different from the most similar tool seen during training.
\item \textbf{Tool Embedding}: Each test tool's representation is at least $d$ Euclidean distance away from each training tool.
\item \textbf{Unseen Ratio}: The test action set is a mixture of seen and unseen tools, with $x\%$ unseen.
\end{enumerate}
The results suggest a gradual decrease in generalization performance as the test actions become more different from training actions. We chose the hardest settings for the main experiments: 15$^{\circ}$ angle difference and 100\% unseen actions.

\vspace{-4pt}
\subsubsection{Qualitative Analysis}
Figure~\ref{fig:qual_results} shows success and failure examples when using unseen actions in the CREATE and Stacking environments. In CREATE, our framework correctly infers the directional pushing properties of unseen tools like conveyor belts and fans from their action observations and can utilize them to solve the task. Failure examples include placements being off and misrepresenting the direction of a belt. Collecting more action observations can improve the representations.

In Shape Stacking, the geometric properties of 3D shapes are correctly inferred from image action observations. The policy can act in the environment by selecting the appropriate shapes to drop based on the current tower height. Failures include greedily selecting a tall but unstable shape in the beginning, like a pyramid.

\vspace{-2pt}
\subsection{The Inefficiency of Finetuning on New Actions}
\label{sec:fine_tune}
\begin{figure}[ht]
    \vspace{-10pt}
	\centering
    	\includegraphics[width=0.6\textwidth]{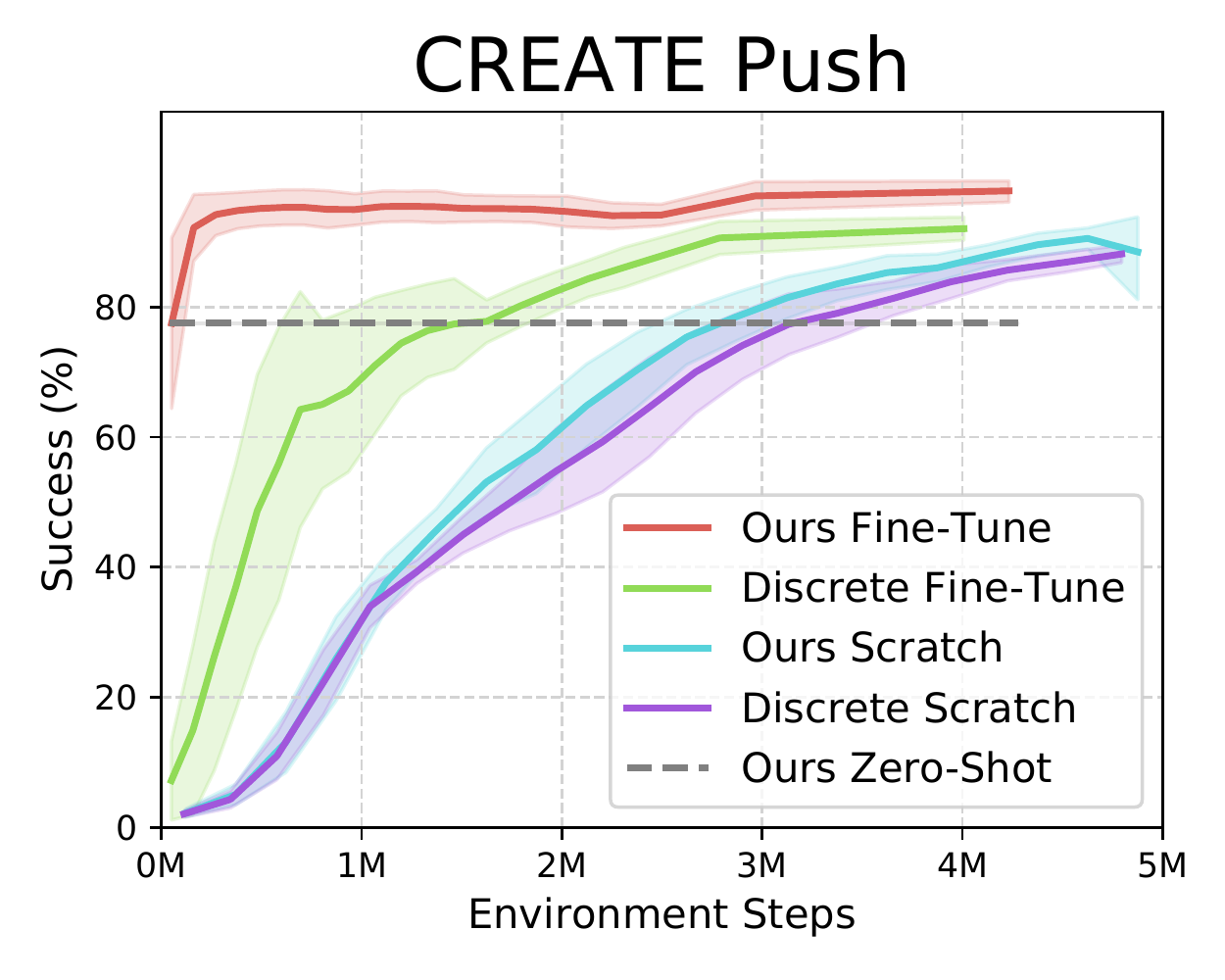}
    \vspace{-15pt}
    \caption{
    Finetuning or training policies from scratch on the new action space. The horizontal line is the zero-shot performance of our method. Each line is the average test performance over 5 random seeds, while the shaded region is the standard deviation. 
    }
    \label{fig:fine_tune}
\end{figure}

In Figure~\ref{fig:fine_tune}, we examine various approaches to continue training on a particular set of new actions in CREATE Push.
First, we train a policy from scratch on the new actions either with our adaptable policy architecture (Ours Scratch) or a regular discrete policy (Discrete Scratch).
These take around 3 million environment steps to achieve our pretrained method's zero-shot performance (Ours Zero-Shot).
Next, we consider ways to transfer knowledge from training actions. We train a regular discrete policy and finetune on new actions by re-initializing the final layer (Discrete Fine-Tune). While this approach transfers some task knowledge, it disregards any relationship between the old and new actions. It still takes over 1 million steps to reach our zero-shot performance. This shows how expensive retraining is on a single action set. Clearly, this retraining process is prohibitive in scenarios where the action space frequently changes. This demonstrates the significance of addressing the problem of zero-shot generalization to new actions.
Finally, we continue training our pretrained policy on the new action set with RL (Ours Fine-Tune). We observe fast convergence to optimal performance, because of its ability to utilize action representations to transfer knowledge from the training actions to the new actions. Finetuning results for all other environments are in Figure~\ref{supp:fig:all_ft}.

\section{Conclusion}
\label{conclusion}
Generalization to novel circumstances is vital for robust agents. We propose the problem of enabling RL policies to generalize to new action spaces.  
Our two-stage framework learns action representations from acquired action observations and utilizes them to make the downstream RL policy flexible.
We propose four challenging benchmark environments and demonstrate the efficacy of hierarchical representation learning, policy architecture, and regularizations.
Exciting directions for future research include building general problem-solving agents that can adapt to new tasks with new action spaces, and autonomously acquiring informative action observations in the physical world.

\section*{Acknowledgements}
This project was funded by SKT. The authors are grateful to Youngwoon Lee and Jincheng Zhou for help with RL experiments and writing. The authors would like to thank Shao-Hua Sun, Karl Pertsch, Dweep Trivedi and many members of the USC CLVR lab for fruitful discussions. The authors appreciate the feedback from anonymous reviewers who helped improve the paper.

\bibliography{references}
\bibliographystyle{icml2020}

\newpage
\twocolumn[
\icmltitle{Generalization to New Actions in Reinforcement Learning - Appendix}
]
\appendix
\section{Environment Details}
\label{supp:sec:env_details}

\subsection{Grid World}
\label{supp:gw}

\begin{figure*}[!ht]
    \centering
	\includegraphics[width=0.8\textwidth]{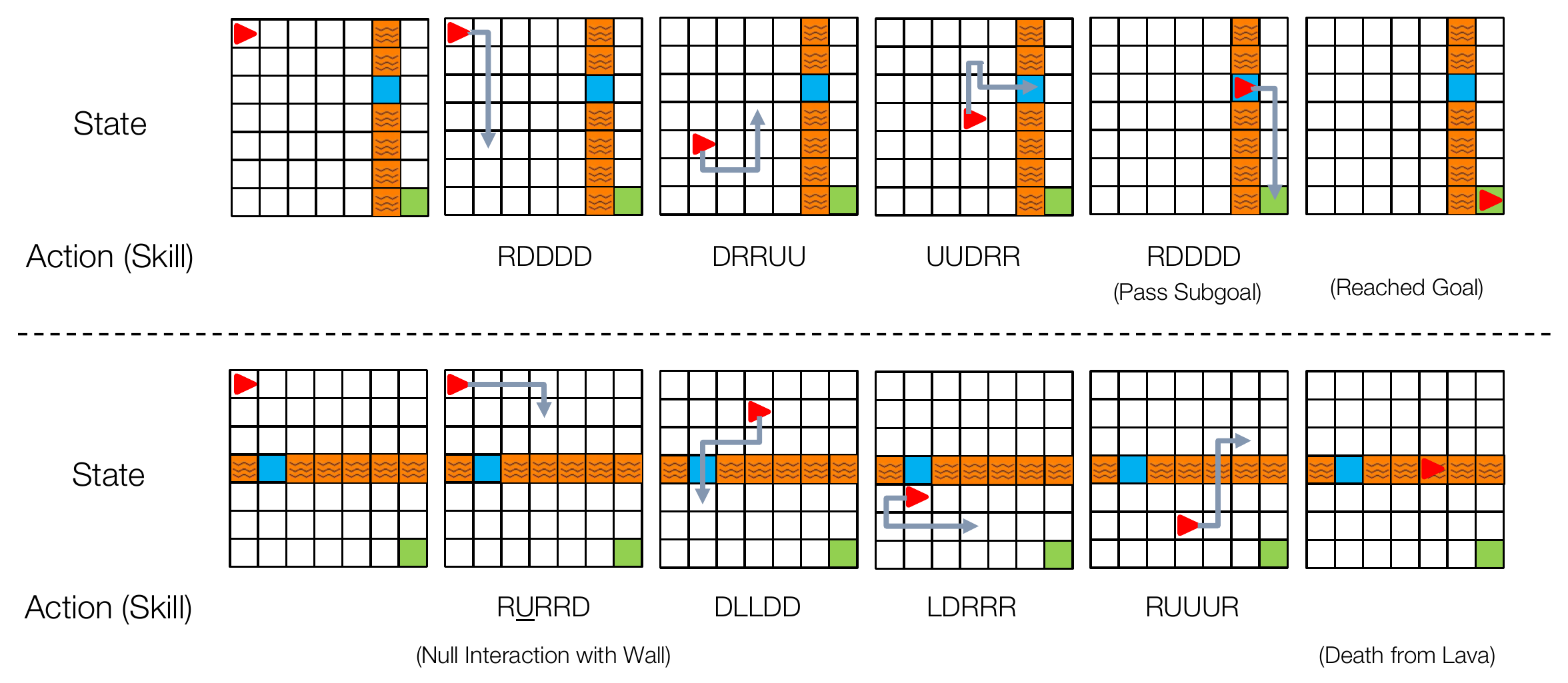}
	\vspace{-5pt}
    \caption{
      Grid World Environment: 9x9 grid navigation task.
      The agent is the red triangle, and the goal is the
      green cell. The environment contains one row or column of lava wall with
      a single opening acting as a subgoal (blue). Each action consists of
      a sequence of 5 consecutive moves in one of the four directions: U(p),
      D(own), R(ight), L(eft).
    }
    \label{supp:fig:gw_env_figure}
\end{figure*}

The Grid World environment, based on~\citet{gym_minigrid}, consists of an agent and a randomly placed lava wall with an opening, as shown in Figure~\ref{supp:fig:gw_env_figure}. 
The lava wall can either be horizontal or vertical. 
The agent spawns in the top left corner, and its objective is
to reach the goal in the bottom-right corner of the grid while avoiding any
path through lava. The agent can move using 5-step skills composed of steps in
one of the four directions (Up, Down, Left, and Right). An episode is
terminated when the agent uses a maximum of 10 actions (50 moves),
or the
agent reaches the goal (success) or lava wall (failure).

\textbf{State}: The state space is a flattened version of the 9x9 grid. Each
element of the 81-dimensional state contains an integer ID based on whether
the cell is empty, wall, agent, goal, lava, or subgoal.

\textbf{Actions}: An action or skill of the agent is a sequence of 5
consecutive moves in 4 directions. Hence, $4^5=1,024$ total actions are
possible. Once the agent selects an action, it executes 5 sequential moves
step-by-step. During a skill execution, if the agent hits the boundary wall,
it will stay in the current cell, making a null interaction. If the agent steps
on lava during any action, the game will be terminated.

\textbf{Reward}: Grid world provides a sparse subgoal reward on passing the subgoal for the first time and a sparse goal reward when the agent reaches the goal. The goal reward is discounted based on the number of actions taken to encourage a shorter path to the goal. More concretely,
\begin{equation}
\begin{split}
    R(s) = \lambda_{Subgoal} \cdot \textbf{1}_{Subgoal} \: + \:
    ( 1 - \lambda_{Goal} \frac{N_{total}}{N_{max}} ) \cdot \textbf{1}_{Goal}
\end{split}
\end{equation}
where $\lambda_{Subgoal} = 0.1, \: \lambda_{Goal} = 0.9, \: N_{max} = 50,$ \\
$N_{total} = \text{number of moves to reach the goal}$.

\textbf{Action Set Split}: The whole action set is randomly divided into a {2:1:1}~split of train, validation, and test action sets.

\textbf{Action Observations}: The observations about each action demonstrate an
agent performing the 5-step skill in an 80x80 grid with no obstacles. Each
observation is a trajectory of states resulting from the skill being applied,
starting from a random initial state on the grid. 
A set of 1024 such trajectories characterizes a single skill. By observing the
effects caused on the environment through a skill, the action representation module
can extract the underlying skill behavior, which is further used in the
actual navigation task. Different types of action representations are
described and visualized in Section~\ref{supp:sec:action_rep_vis}.

\subsection{Recommender System}

\begin{figure}[!ht]
    \centering
	\includegraphics[width=0.8\textwidth]{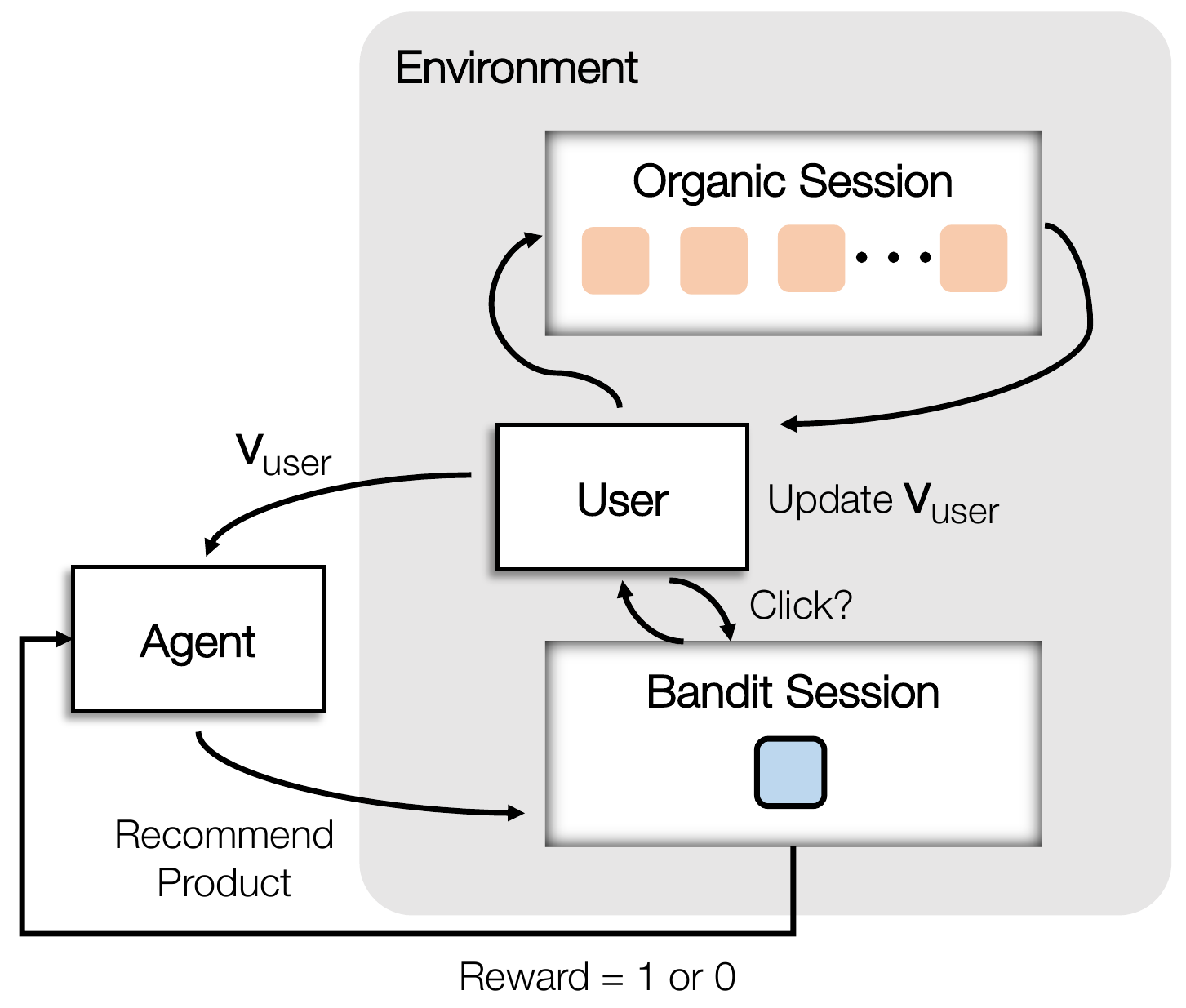}
    \caption{Recommender System schematic:
    The user transitions stochastically between two sessions: organic and
    bandit. Each transition updates the user vector. Organic
    sessions simulate the user independently browsing other products. Bandit sessions
    simulate the agent recommending products to the current user. A reward is
    given if the user clicks on the recommended product. 
    }
    \label{supp:fig:reco_env_figure}
\end{figure}

We adapt the Recommender System environment from \citet{rohde2018recogym} that
simulates users responding to product recommendations (the schematic shown in
Figure~\ref{supp:fig:reco_env_figure}).  Every episode, the agent makes a
series of recommendations for a new user to maximize their
cumulative click-through rate.  Within an episode, there are two types of
states a user can transition between: organic session and bandit session.  
In the bandit session, the agent recommends one of the available products to the
user, which the user may select. After this, the user can transition to an
organic session, where the user independently browses products. The agent takes action (product recommendation) whenever the user transitions to the bandit
session. Every user interaction with organic or bandit sessions varies their
preferences slightly, resulting in a change to the user's vector. As a result,
the agent cannot repetitively recommend the same products in an episode, since the user is unlikely to click it again. The
environment provides engineered action representations, which are also used by the environment to determine the likelihood of a user clicking on the recommendation.  The episode terminates after 100 recommendations or stochastically in between the session transitions.

\textbf{State}: The state is a 16-dimensional vector representing the user, $\textbf{v}_{user}$. Every episode, a new user is created with a vector $\textbf{v}_{user} \sim \mathcal{N} (\bm{0}, \bm{I})$. After each step in the episode, the user transitions between organic and bandit sessions, where the user vector is perturbed by resampling $\textbf{v}_{user} \sim \mathcal{N} (\textbf{v}_{user}, \, \sigma_1 \sigma_2 \bm{I})$, where $\sigma_1 = 0.1$ and $\sigma_2 \sim \mathcal{N} (0, 1)$.

\begin{figure*}[!ht]
    \centering
	\includegraphics[width=\textwidth]{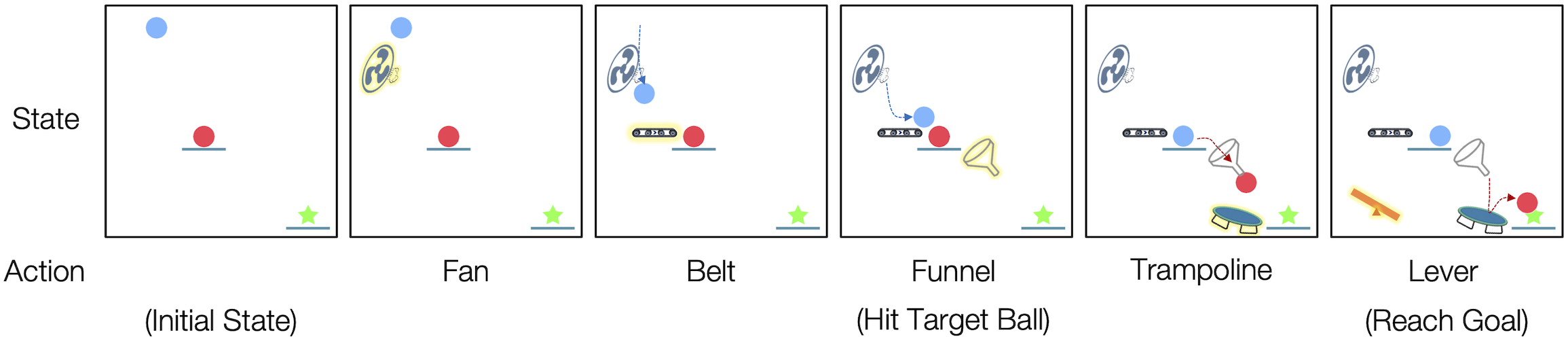}
    \caption{{CREATE Push} Environment: The blue ball falls into the scene and is directed towards the target ball (red), which is pushed towards the goal location (green star).
    This is achieved with the use of various physical tools that manipulate the path of moving objects in peculiar ways.
    At every step, the agent decides which tool to place and the $(x,y)$ position of the tool on the screen.}
    \label{supp:fig:create_env_figure}
\vspace{-5pt}
\end{figure*}

\textbf{Actions}: There are a total of 10,000 actions (products) to recommend to users. Each action is associated with a 16 dimension representation, $c \sim \mathcal{N}(\bm{0}, \bm{I})$. The selected product's representation and the current user vector determine the probability of a click. The agent's objective is to recommend articles that maximize the user's click-through rate. The probability of clicking a recommended product $i$ with action representation $c_i$ is given by:
\begin{equation}
\label{supp:eq:reco_click}
\begin{split}
    p_{click}(\textbf{v}_{user}, \textbf{c}_i) = f(\textbf{c}_i \cdot \textbf{v}_{user} + \mu_i), \text{where} \\
    f(x) = \sigma(a * \sigma(b * \sigma(c * x) - d) - e),
\end{split}
\end{equation}
where $a = 14, b = 2, c = 0.3, d = 2, e = 6, \sigma$ is the sigmoid function, $\cdot$ denotes a vector dot product. Here, $\mu_i$ is an action-specific constant kept hidden from the agent to simulate partial observability, as would be the case in real-world recommender systems. Constants used in the function $f$ make the click-through rate, $p_{click}$, to be a reasonable number, adapted and modified from~\citet{rohde2018recogym}.

In Section~\ref{supp:sec:fully_obs_reco}, we also provide results on the fully observable recommender system environment, where the agent has access to $\mu_i$ as well. Concretely, $\mu_i$ is concatenated to $c_i$ to form the action representation which the learning agent utilizes to generalize.

\textbf{Reward}: There is a dense reward of $1$ on every recommendation that receives a user click, which is determined by $p_{click}$ computed in Eq~\ref{supp:eq:reco_click}.

\textbf{Action Set Split}:
The 10,000 products are randomly divided into a 2:1:1 split of train, validation, and test action sets.

\subsection{Chain REAction Tool Environment (CREATE)}
\label{supp:create}

Inspired by the popular video game, \textit{The Incredible Machine}, {Chain REAction Tool Environment (CREATE)} is a physics-based puzzle where the objective is to get a target ball (red) to a goal position (green), as depicted in Figure~\ref{supp:fig:create_env_figure}. Some objects start suspended in the air, resulting in a falling movement when the game starts.
The agent is required to select and place tools to redirect the target ball towards the goal, often using other objects in the puzzle (like the blue ball in Figure~\ref{supp:fig:create_env_figure}).
The agent acts every 40 physics simulation steps to make the task reasonably challenging and uncluttered.
An episode is terminated when the agent accomplishes the goal, or after 30 actions, or when there are no moving objects in the scene, ending the game. 
CREATE was created with the Pymunk 2D physics library~\citep{pymunk} and Pygame physics engine~\citep{pygame}.

CREATE environment features 12 tasks, as shown in Figure~\ref{supp:fig:create_all_tasks}. Results for 3 main tasks are shown in Figure~\ref{fig:baselines},~\ref{fig:ablations} and 9 others in Figure~\ref{supp:fig:add_create}.
Concurrently developed related environments~\citep{allen2019tools, bakhtin2019phyre} focus
on single-step physical reasoning with a few simple polygon tools. In contrast, CREATE supports multi-step RL, features many diverse tools, and requires continuous tool placement.

\textbf{State}: At each time step, the agent receives an 84x84x3 pixel-based observation of the game screen. Here, each originally colored observation is turned into gray-scale and the past 3 frames are stacked channel-wise to preserve velocity and acceleration information in the state.

\textbf{Actions}: In total, CREATE consists of 2,111 distinct tools (actions) belonging to the classes of: ramp, trampoline, lever, see-saw, ball, conveyor belt, funnel, 3-, 4-, 5-, and 6-sided polygon, cannon, fan, and bucket. 2,111 tools are obtained by generating tools of each class with appropriate variations in parameters such as angle, size, friction, or elasticity. The parameters of variation are carefully chosen to ensure that any resulting tool is significantly different from other tools. For instance, no two tools are within 15$^{\circ}$ difference of each other.
There is also a \textit{No-Operation} action, resulting in no tool placement.

The agent outputs in a hybrid action space consisting of (1) the discrete tool
selection from the available tools, and (2) $(x,y)$ coordinates for 
placing the tool on the game screen.

\textbf{Reward}:
CREATE is a sparse reward environment where rewards are given for reaching
the goal, reaching any subgoal once, and making the target ball move in certain
tasks. Furthermore, a small reward is given to continue the episode. There is a penalty for trying to overlap a new tool over existing objects in the scene and an invalid penalty for placing outside the scene.
The agent receives the following reward:
\begin{equation}
\label{supp:eq:create_reward}
    \begin{split}
        R(s, a) = \lambda_{alive} \, + \, \lambda_{Goal} \, \cdot \, \textbf{1}_{Goal} \, \cdot \\
        \lambda_{Subgoal} \, \cdot \, \textbf{1}_{Subgoal} \, \cdot \,
        \lambda_{target \, hit} \, \cdot \, \textbf{1}_{target \, hit} \, + \\
        \lambda_{invalid} \, \cdot \, \textbf{1}_{invalid} \, + \,
        \lambda_{overlap} \, \cdot \, \textbf{1}_{overlap}
    \end{split}
\end{equation}
where $\lambda_{alive} = 0.01$, $\lambda_{Goal} = 10.0$,  $\lambda_{Subgoal} = 2.0$, $\lambda_{target \, hit} = 1$, and $\lambda_{invalid} = \lambda_{overlap} = -0.01$.

\textbf{Action Set Split}:
The tools are divided into a 2:1:1 split of train, validation, and test action
sets. In \textit{Default Split} presented in the main experiments, the tools are split such that the primary parameter (angle for most) is randomly split between training and testing. This ensures that the test tools are considerably different from the training tools in the same class.
The validation set is obtained by randomly splitting the testing set into half.
In \textit{Full Split}, 1,739 of the total tools are divided
into a 2:1:1 split by tool class, as described in
Table~\ref{supp:tab:create_full_split}.

\begin{table}[ht]
\centering
    \begin{tabular}{@{}ll@{}}
    \toprule
         \textbf{Train} & {\begin{tabular}{@{}l@{}}Ramp, Trampoline, Ball, Bouncy Ball,\\ See-saw, Cannon, Bucket\end{tabular}} \\
    \midrule
         {\begin{tabular}{@{}l@{}}\textbf{Validation} \\ \textbf{and Test} \end{tabular}} & {\begin{tabular}{@{}l@{}}Triangle, Bouncy Triangle, Lever, \\ Fan, Conveyor Belt, Funnel\end{tabular}} \\
    \bottomrule
    \end{tabular}
    \vspace{-5pt}
    \caption{Tool classes in the CREATE Full split.}
    \label{supp:tab:create_full_split}
\end{table}

Additionally, we used a total of 7566 tools generated at 3$^{\circ}$ angle differences for analysis experiments to study generalization properties. HVAE was trained as an
oracle encoder over the entire action set, to get action representations suitable for all three analyses. The policy's performance was studied independently by training it on 762 distinct tools with at least 15$^{\circ}$ angle differences and evaluated based on analysis-specific action sampling from the rest of the tools (\eg at least 5$^{\circ}$ apart).

\textbf{Action Observations}: Each tool's observations are obtained by testing
its functionality through scripted interactions with a probe ball. The probe
ball is launched at the tool from various angles, positions, and speeds. The
tool interacts with the ball and changes its trajectory depending on its
properties, \eg a cannon will catch and re-launch the ball in a fixed
direction. Thus, these deflections of the ball can be used to infer the
characteristics of the tool. Examples of these action observations are shown at
\url{https://sites.google.com/view/action-generalization/create}.

The collected action observations have 1024 ball trajectories of length 7 for each tool.
The trajectory is composed of the environment states, which can take the form
of either the 2D ball position (default) or 48x48 gray-scale images. The action
representation module learns to reconstruct the corresponding data mode, either
state trajectories or videos, for obtaining the corresponding action
representations. Different types of action representations used are described
and visualized in Section~\ref{supp:sec:action_rep_vis}.

\subsection{Shape Stacking}
\label{supp:bs}
In Shape Stacking, the agent must place shapes to build a tower as high as possible.
The scene starts with two cylinders of random heights and colors, dropped at random locations on a line, which the agent can utilize to stack towers.
For each action, the agent selects a shape to place and where to place
it. The agent acts every 300 physics simulator steps to give time for placed objects to
settle into a stable position. The episode terminates after 10 shape placements. 

\textbf{State}: The observation at each time step is an 84x84 grayscale image of the shapes lying on the ground. We stack past 4 frames to preserve previous observations in the state.

\textbf{Actions}: The action consists of a discrete selection of the shape to place, the $x$
position on the horizontal axis to drop the shape, and a binary episode termination action. The height of the drop is automatically calculated over the topmost shape, enabling a soft drop.
If a shape has already been placed, trying to place it again does nothing.
There are a total of 810 shapes of classes: triangle,
tetrahedron,
rectangle, cone, cylinder, dome, arch, cube, sphere, and capsule. These shapes are generated by varying the scale and vertical orientation in each shape class. The parameter variance is carefully chosen to ensure all the shapes are sufficiently different from each other.

In Figure~\ref{supp:fig:add_stack}, we compare various hybrid action spaces with shape selection. We study different ways of placing a shape: dropping at a fixed location, or deciding $x$-position, or deciding $(x,y)$-positions. 

\textbf{Reward}:
To encourage stable and tall towers, there is a sparse reward at episode end, for the final height of the topmost shape in the scene, added to the average heights of all $N$ shapes in the scene:
\setlength{\belowdisplayskip}{10pt} \setlength{\belowdisplayshortskip}{10pt}
\setlength{\abovedisplayskip}{10pt} \setlength{\abovedisplayshortskip}{10pt}
\begin{equation}
\label{supp:eq:stack_reward}
    \begin{split}
    R(s) = (\lambda_{top} \max(h_i) + \lambda_{avg} \frac{1}{N} \sum_{i} h_i) \, \cdot \, 
    \textbf{1}_{Done},
  \end{split}
\end{equation}
where $h_i$ is the height of shape $i$ and $\lambda_{top} = \lambda_{avg} =
0.5$. 

\textbf{Action Set Split}: The shapes are divided into a 2:1:1 split of train, validation, and test action sets. In Default Split presented in the main experiments, the shapes are split such that the primary parameter of scale is randomly split between training and testing. This ensures that test tools are considerably different in scale from the train tools in the same class. The validation set is obtained by randomly splitting the test set into half. In Full Split, the split is determined by shape class, as shown in Table~\ref{supp:tab:ss_full_split}. 

\begin{table}[ht]
\centering
    \begin{tabular}{@{}ll@{}}
    \toprule
         \textbf{Train} & {\begin{tabular}{@{}l@{}}Domes, Rectangles, Capsules, Triangles, \\ Arches, Spheres \end{tabular}} \\
    \midrule
         {\begin{tabular}{@{}l@{}}\textbf{Validation} \\ \textbf{and Test} \end{tabular}} & {\begin{tabular}{@{}l@{}}Cylinders, Tetrahedrons, Cubes, Cones, \\ Angled-Rectangles, Angled-Triangles\end{tabular}} \\
    \bottomrule
    \end{tabular}
    \vspace{-10pt}
    \caption{Shape classes in the Shape Stacking Full split.}
    \label{supp:tab:ss_full_split}
\end{table}

\textbf{Action Observations}: In Shape Stacking the functionality of each
action is characterized by the physical appearance of the shape. Thus, the action observations consist of images of the shape from various camera
angles and heights. Each shape has 1,024 observed images of resolution
84x84. Examples of these action observations are shown at
\url{https://sites.google.com/view/action-generalization/shape-stacking}.

\section{Visualizing Action Representations}
\label{supp:sec:action_rep_vis}

\begin{figure*}[!h]
\centering
	\includegraphics[width=\textwidth]{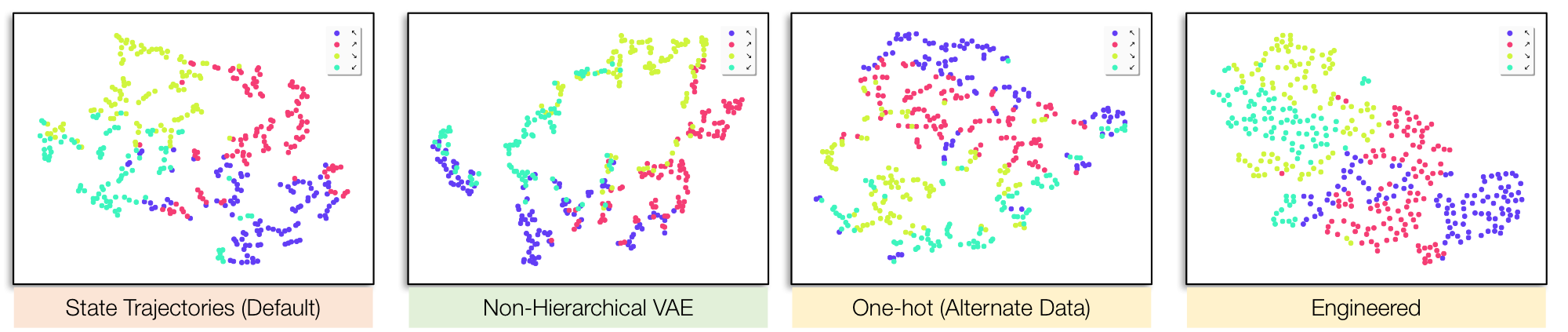}
    \caption{
    t-SNE Visualization of learned skill representation space for Grid World
    environment. Colored by the quadrant that the skill translates the agent to. 
    }
    \label{supp:fig:grid_representations}
\vspace{-5pt}
\end{figure*}

\begin{figure*}[!h]
\centering
	\includegraphics[width=\textwidth]{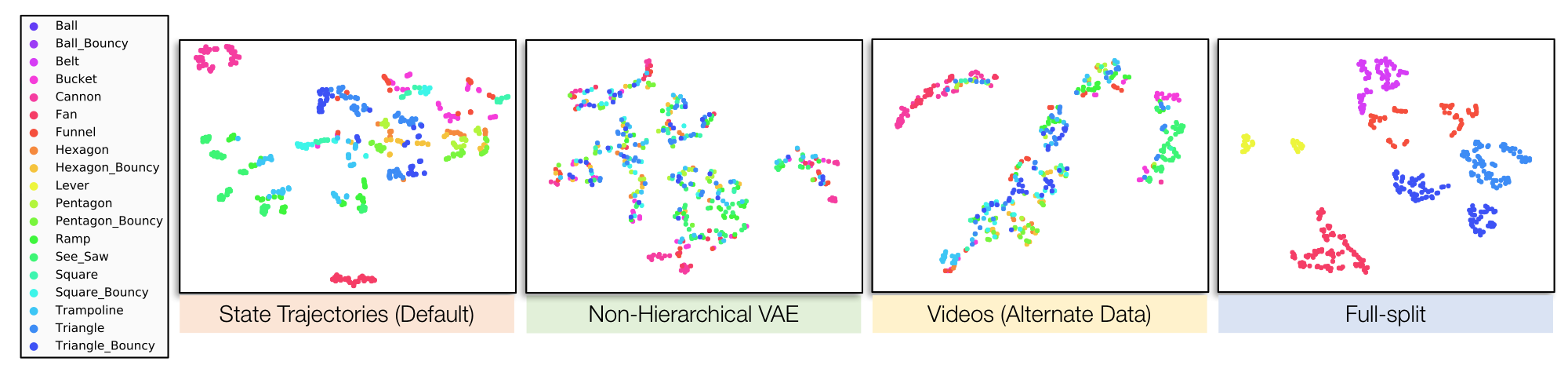}
    \caption{
       t-SNE Visualization of learned tool representation space for CREATE
       environment. Colored by the tool class.
    }
    \label{supp:fig:create_representations}
\vspace{-5pt}
\end{figure*}

\begin{figure*}[!h]
\centering
	\includegraphics[width=\textwidth]{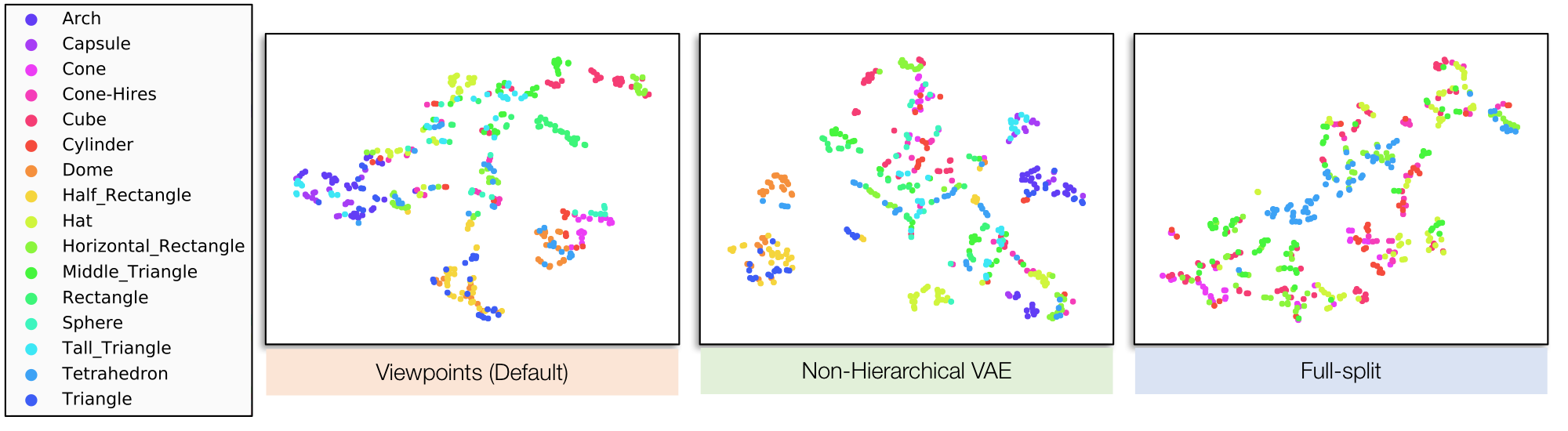}
    \caption{
    t-SNE Visualization of learned action representation space for the Shape-stacking
    environment. Colored by the shape class.
    }
    \label{supp:fig:stack_representations}
\vspace{-5pt}
\end{figure*}

In this work, we train and evaluate a wide variety of action representations based on environments, data-modality, presence or absence of hierarchy in action encoder, and different action splits.
We describe these in detail and provide t-SNE visualizations of the inferred action representations of previously unseen actions.
These visualizations show how our model can extract information about properties of the actions, by clustering similar actions together in the latent space.
Unless mentioned otherwise, the HVAE model is used to produce these representations.

\begin{figure*}[ht]
    \centering
    \begin{subfigure}[t]{0.16\textwidth}
    	\centering
      \frame{\includegraphics[width=\linewidth]{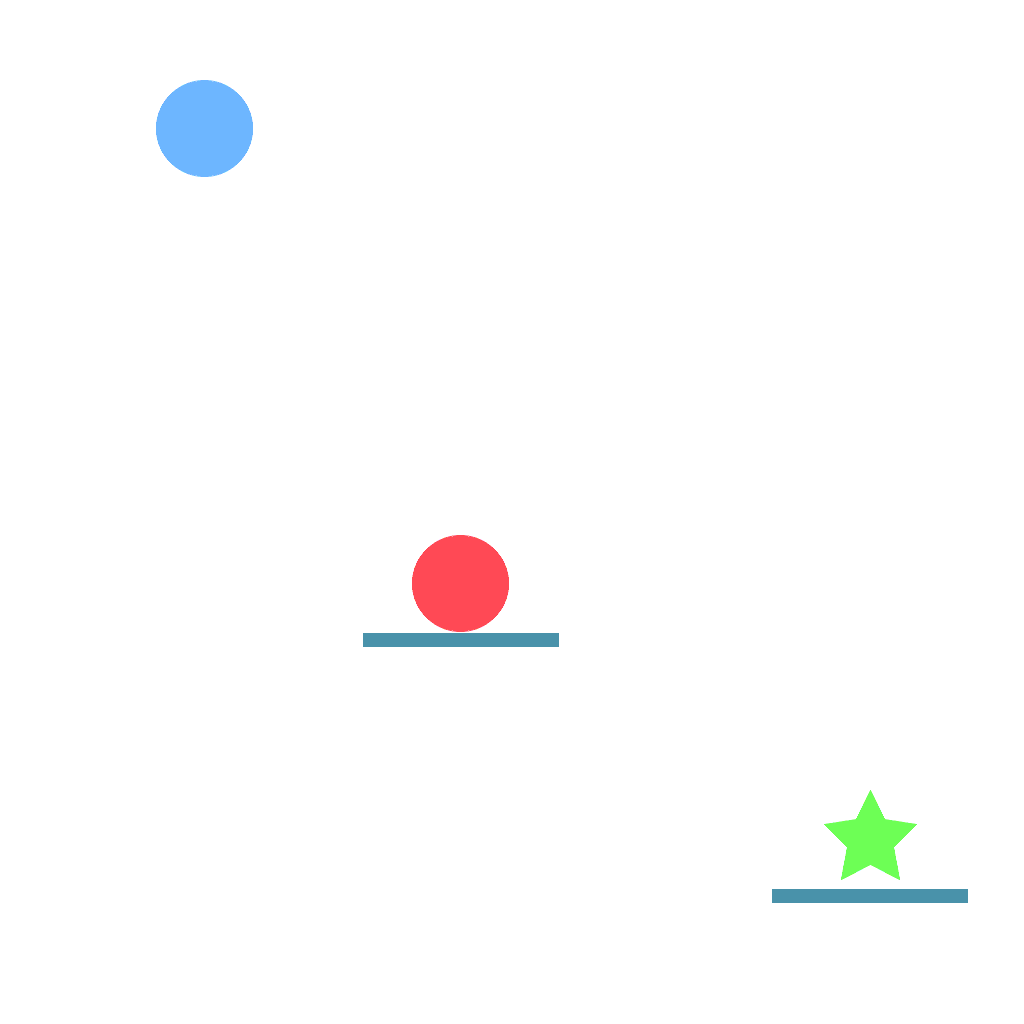}}
      \caption{Push}
    \end{subfigure}
    \begin{subfigure}[t]{0.16\textwidth}
    	\centering
      \frame{\includegraphics[width=\linewidth]{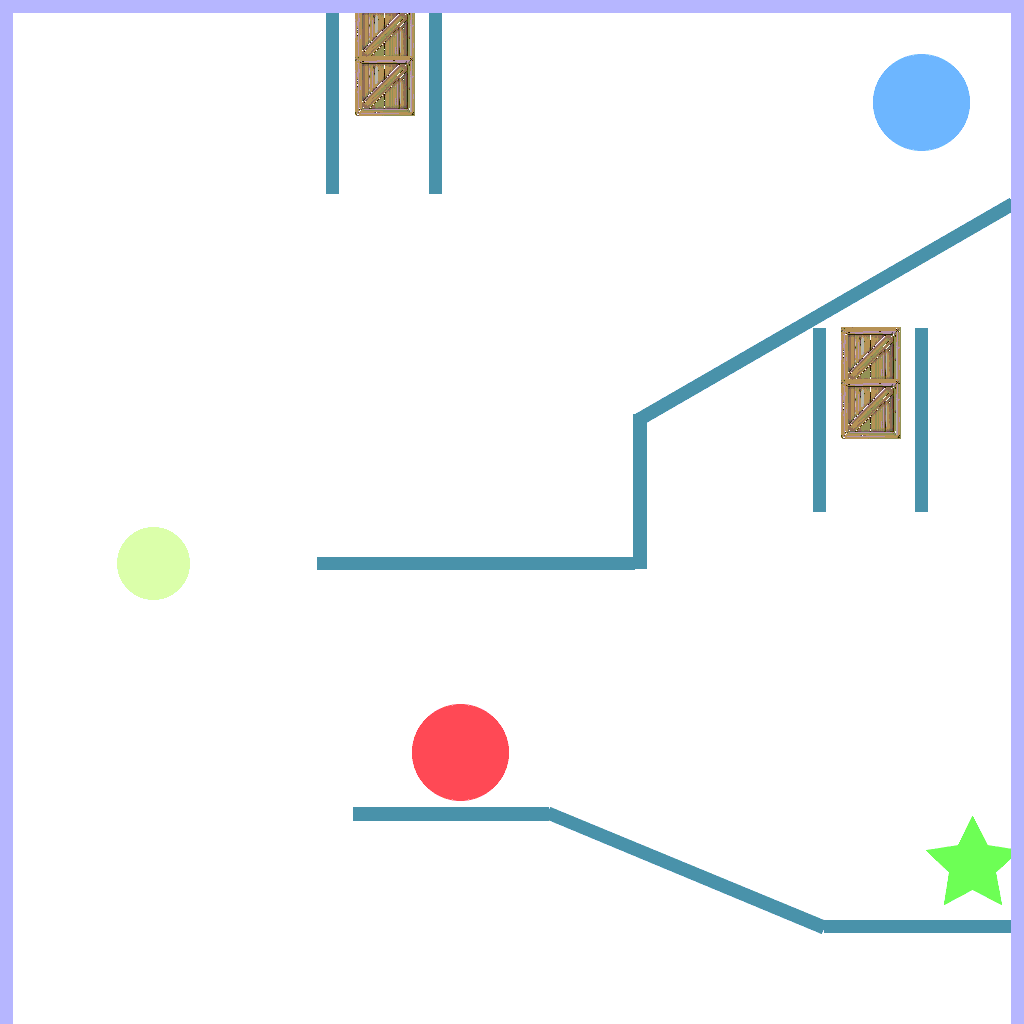}}
      \caption{Obstacle}
    \end{subfigure}
    \begin{subfigure}[t]{0.16\textwidth}
    	\centering
      \frame{\includegraphics[width=\linewidth]{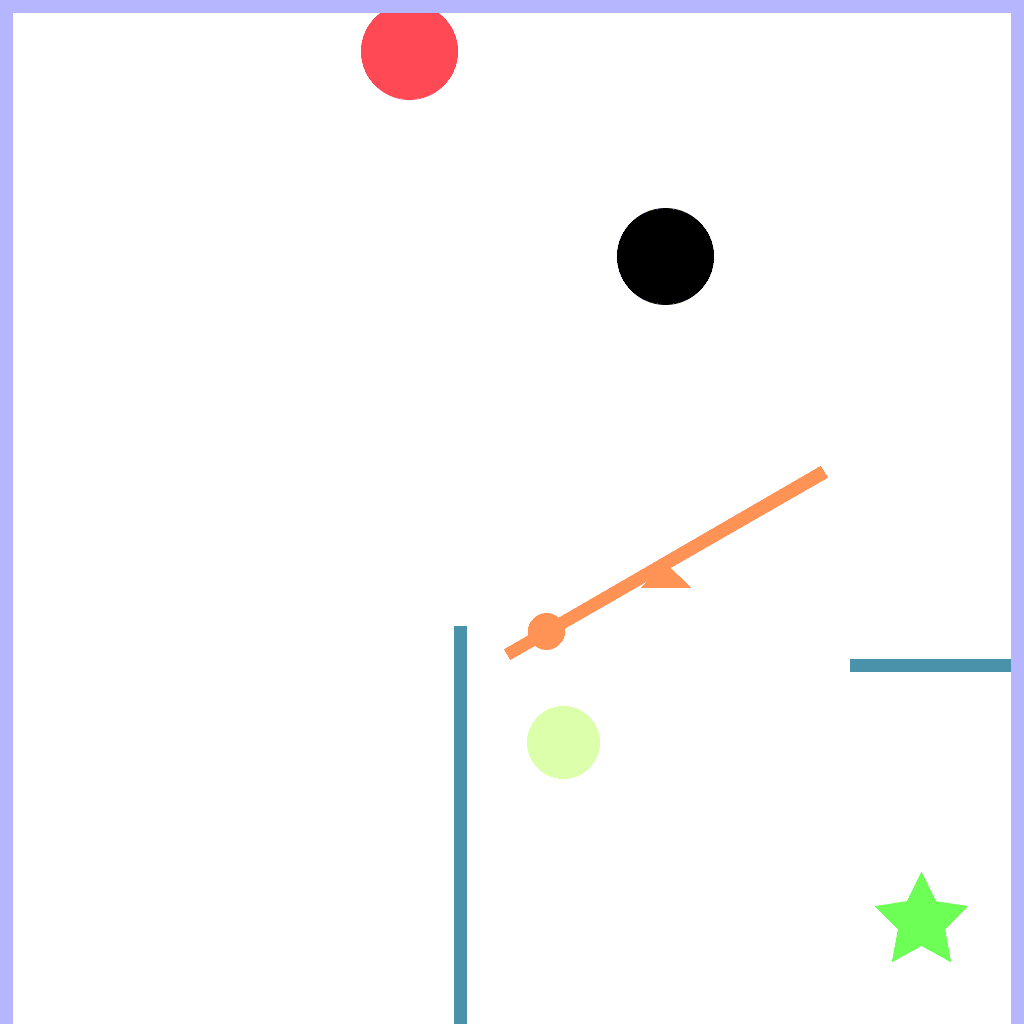}}
      \caption{Seesaw}
    \end{subfigure}
    \begin{subfigure}[t]{0.16\textwidth}
    	\centering
      \frame{\includegraphics[width=\linewidth]{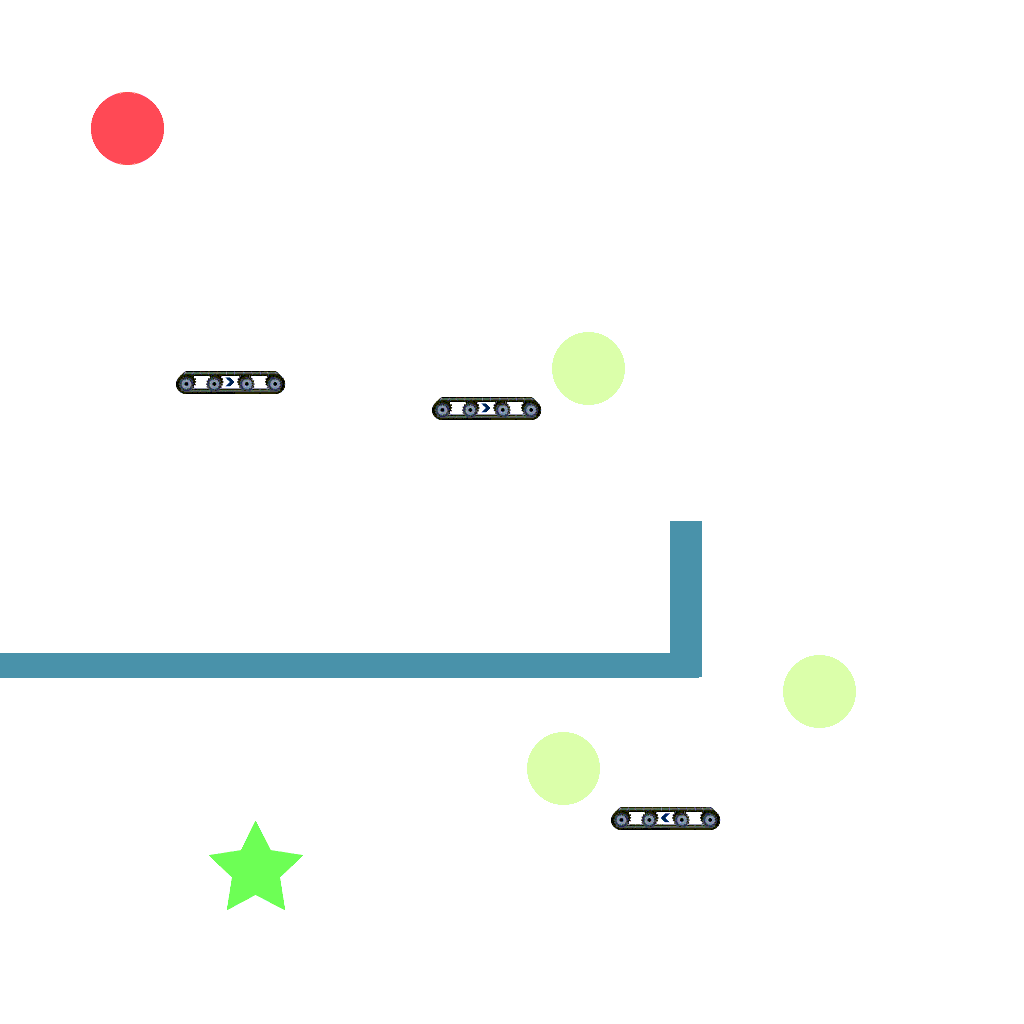}}
      \caption{Belt}
    \end{subfigure}
    \begin{subfigure}[t]{0.16\textwidth}
    	\centering
      \frame{\includegraphics[width=\linewidth]{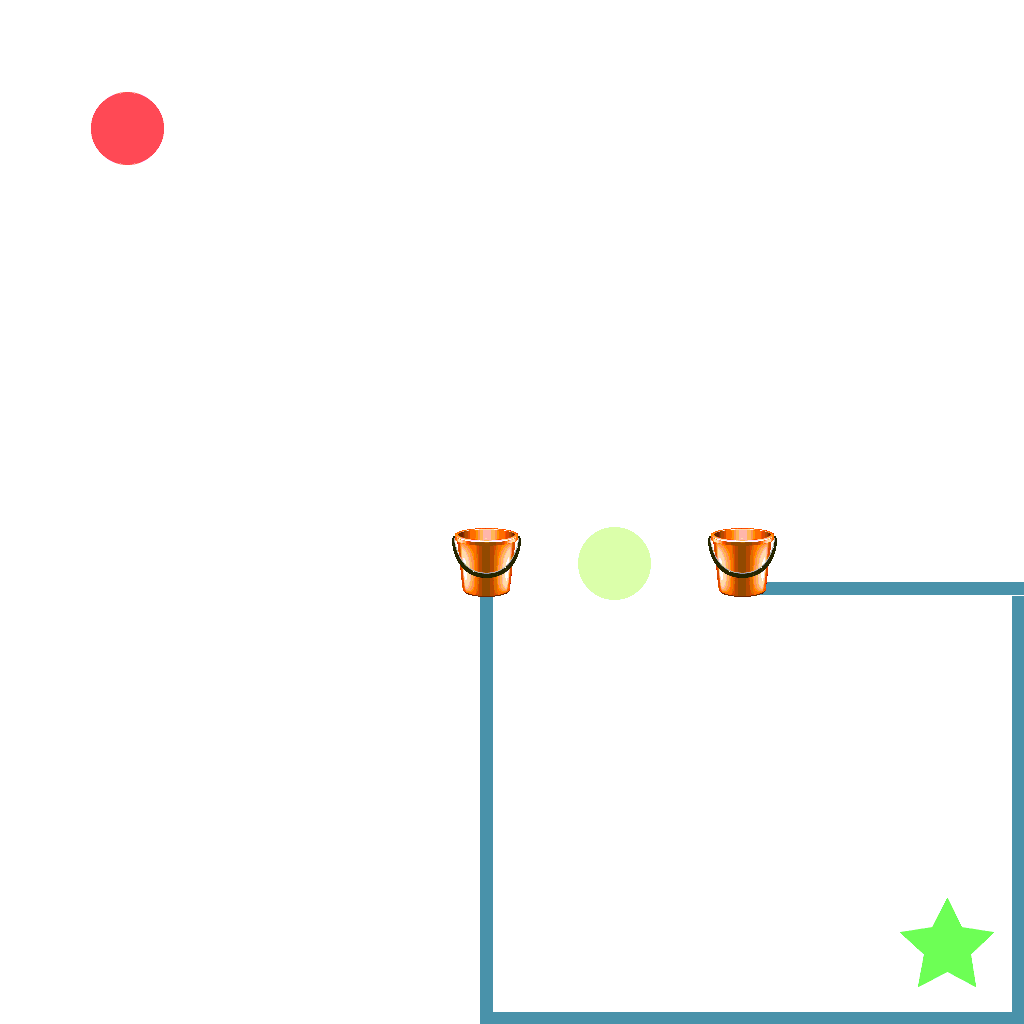}}
      \caption{Bucket}
    \end{subfigure}
    \begin{subfigure}[t]{0.16\textwidth}
    	\centering
      \frame{\includegraphics[width=\linewidth]{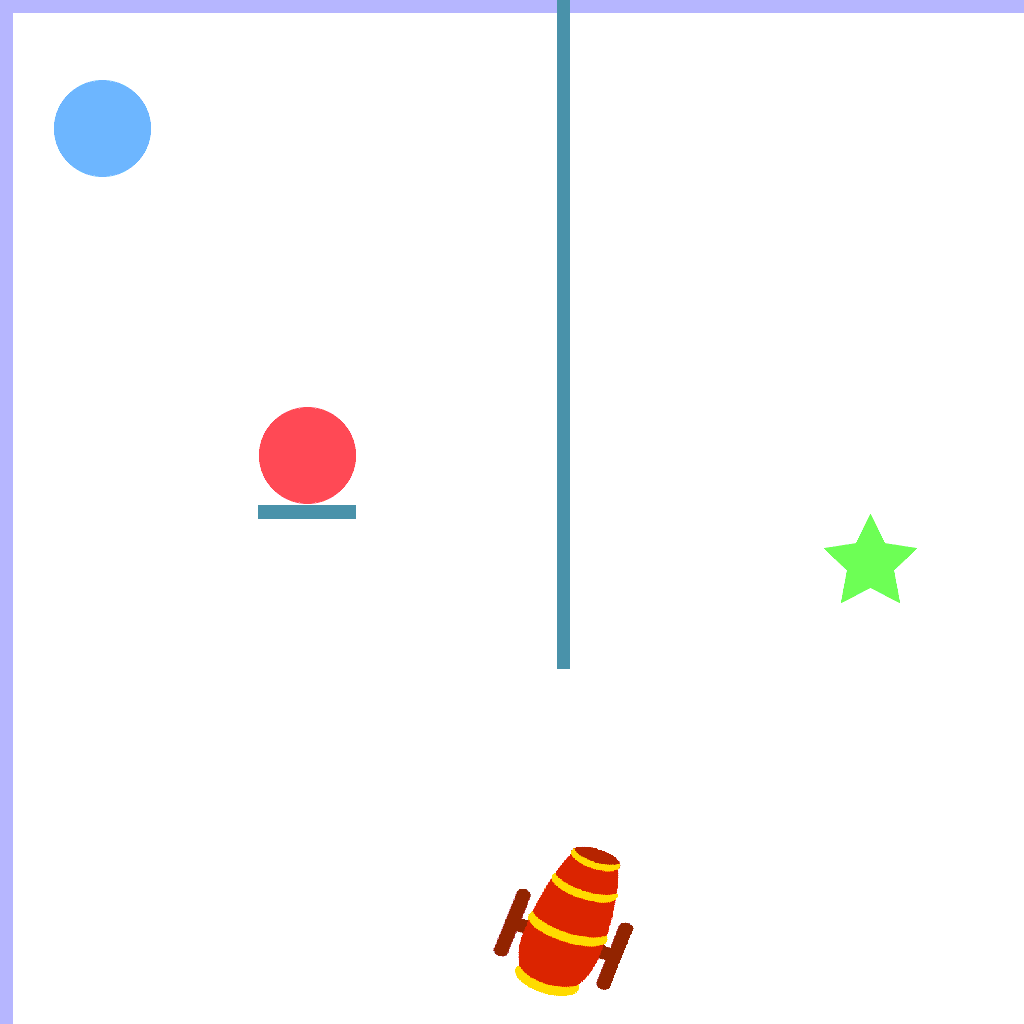}}
      \caption{Cannon}
    \end{subfigure}

    \begin{subfigure}[t]{0.16\textwidth}
    	\centering
      \frame{\includegraphics[width=\linewidth]{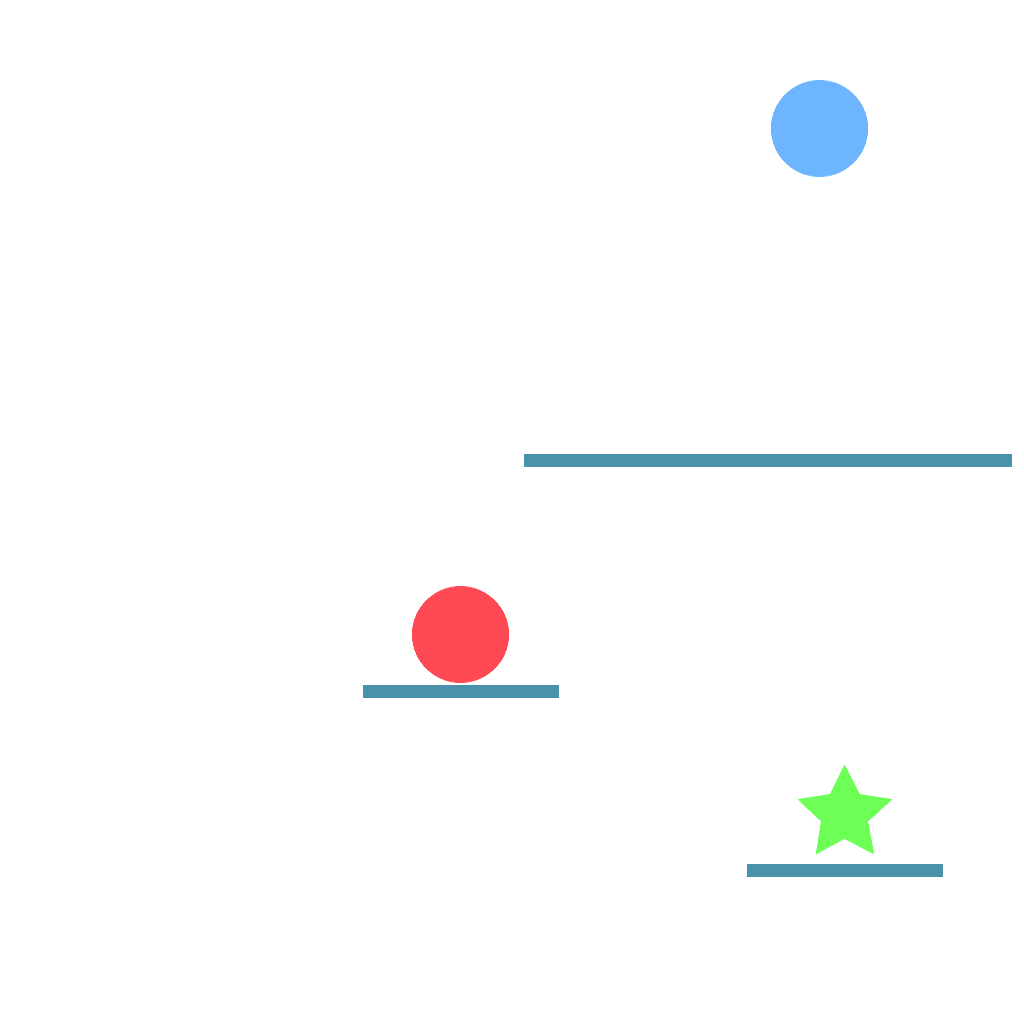}}
      \caption{Navigate}
    \end{subfigure}
    \begin{subfigure}[t]{0.16\textwidth}
    	\centering
      \frame{\includegraphics[width=\linewidth]{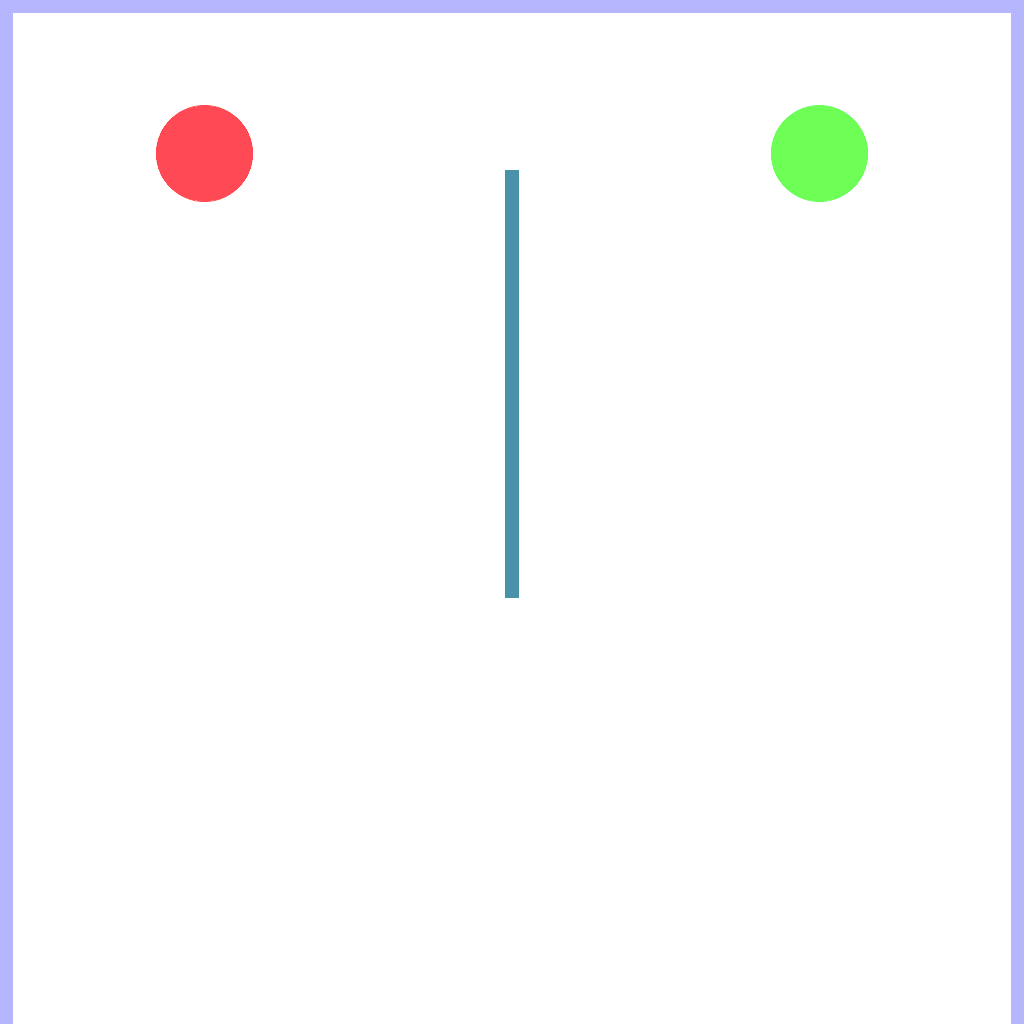}}
      \caption{Collide}
    \end{subfigure}
    \begin{subfigure}[t]{0.16\textwidth}
    	\centering
      \frame{\includegraphics[width=\linewidth]{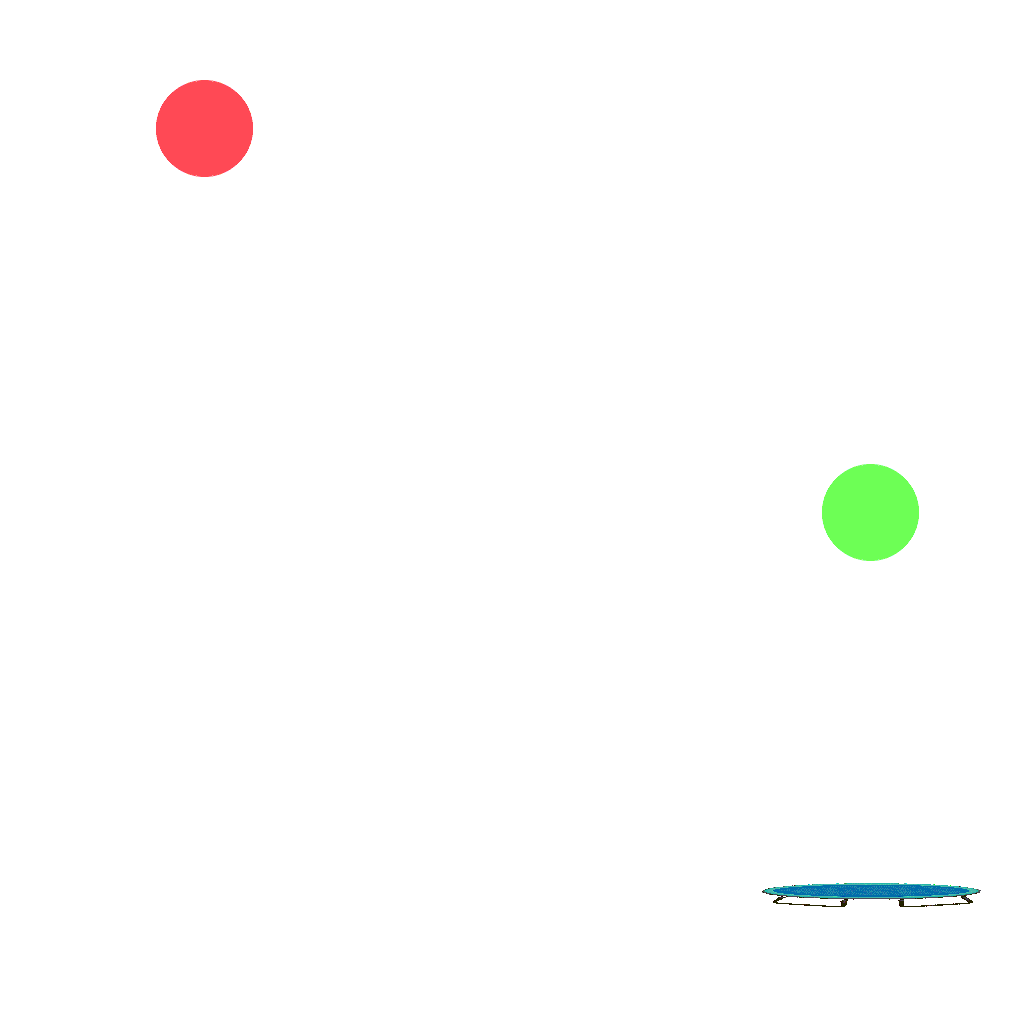}}
      \caption{Moving}
    \end{subfigure}
    \begin{subfigure}[t]{0.16\textwidth}
    	\centering
      \frame{\includegraphics[width=\linewidth]{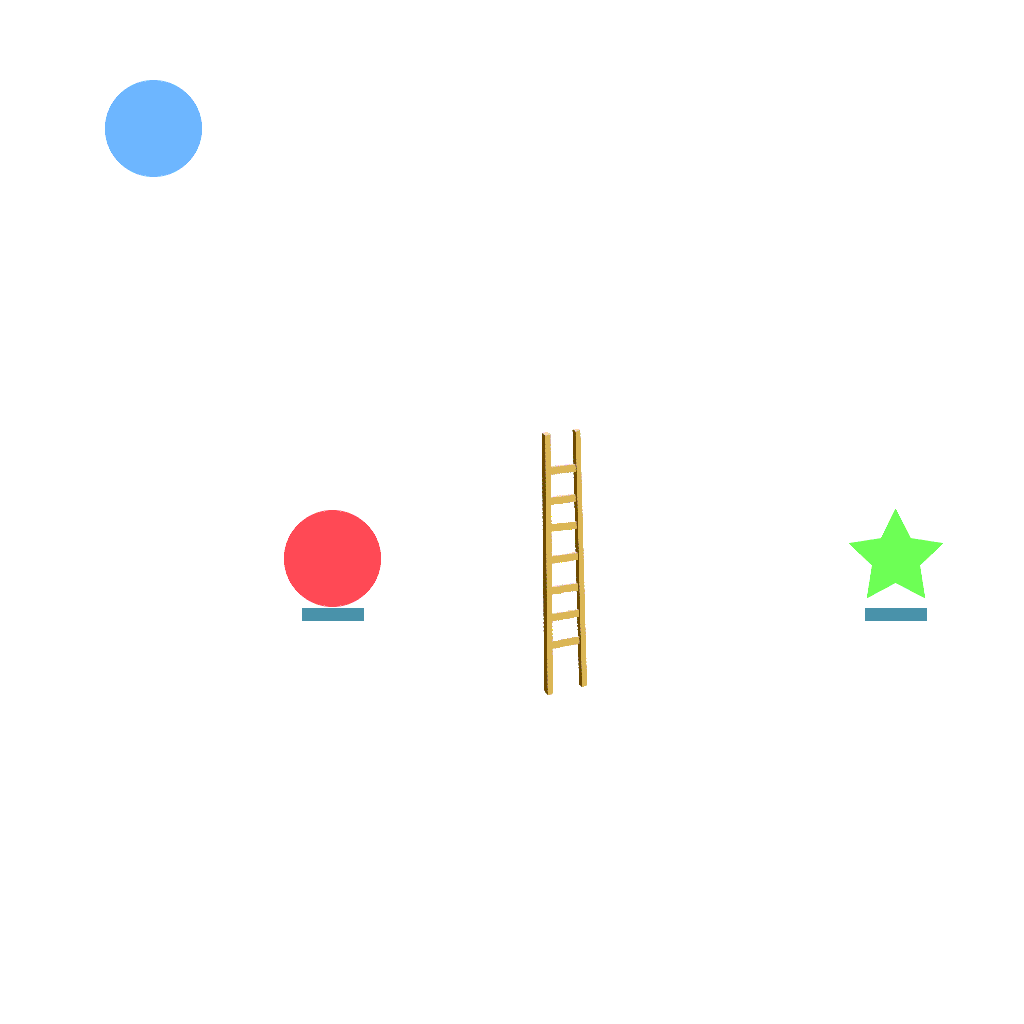}}
      \caption{Ladder}
    \end{subfigure}
    \begin{subfigure}[t]{0.16\textwidth}
    	\centering
      \frame{\includegraphics[width=\linewidth]{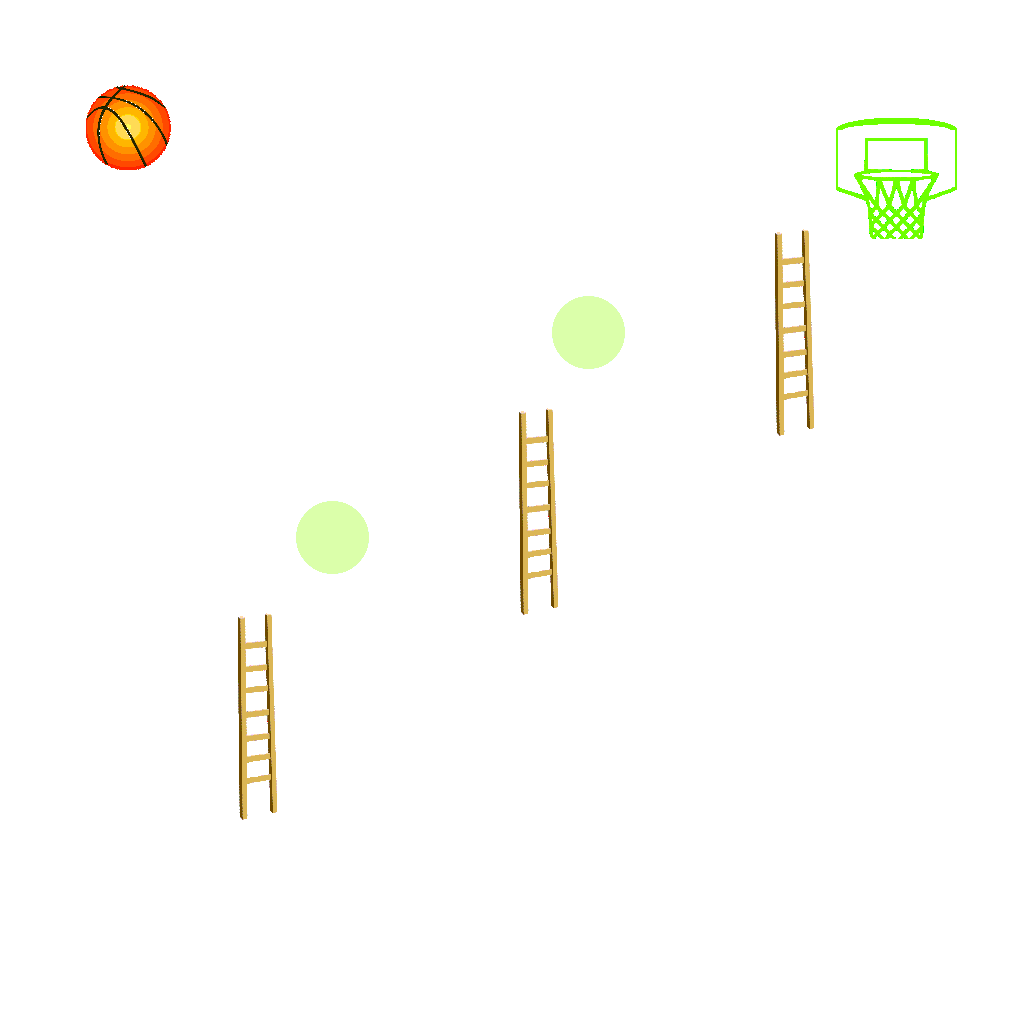}}
      \caption{Basket}
    \end{subfigure}
    \begin{subfigure}[t]{0.16\textwidth}
    	\centering
      \frame{\includegraphics[width=\linewidth]{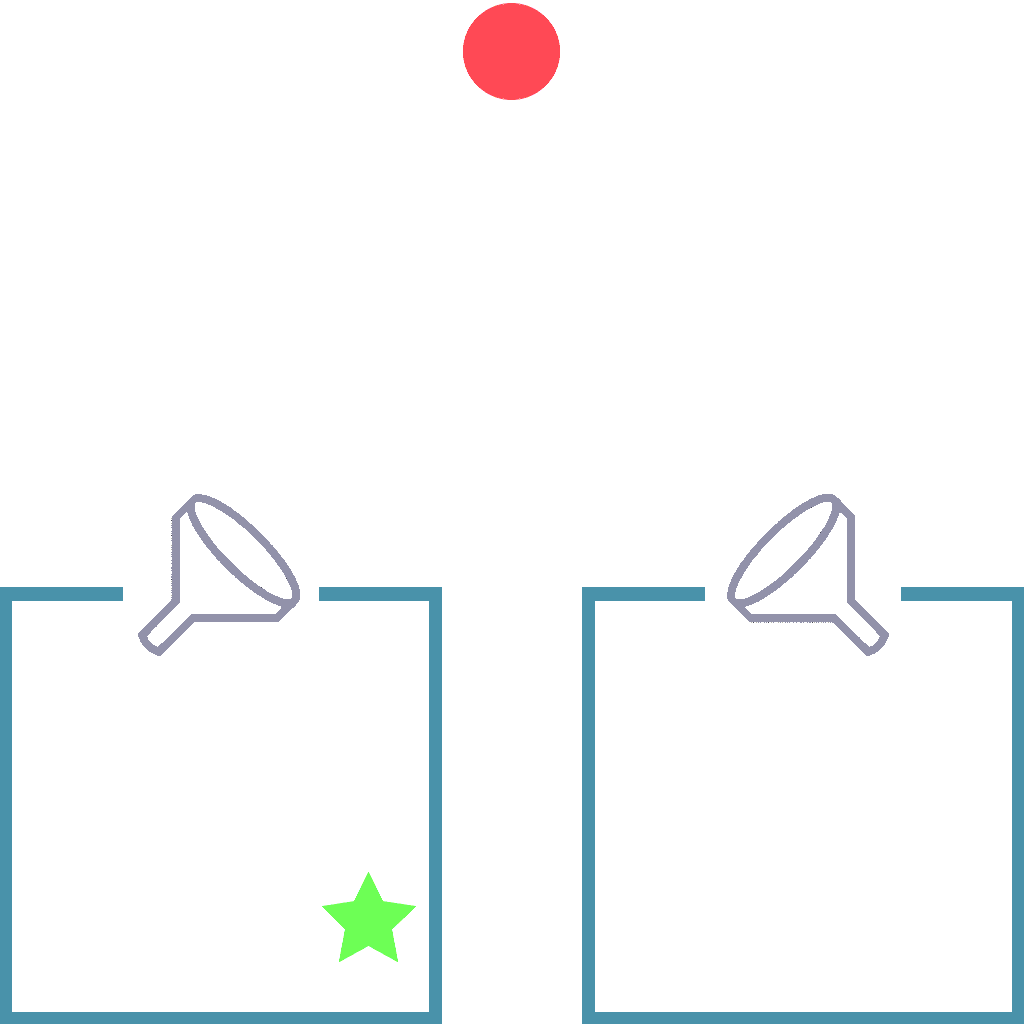}}
      \caption{Funnel}
    \end{subfigure}
    \vspace{-10pt}

    \caption{
      12 CREATE tasks. Complete results on (a) - (c)
      are in Figure~\ref{fig:baselines},~\ref{fig:ablations}, while (d) - (l) are in Figure~\ref{supp:fig:add_create}. 
    }
    \label{supp:fig:create_all_tasks}
\end{figure*}

\begin{figure*}[ht]
    \begin{subfigure}[t]{1\textwidth}
      \centering
      \includegraphics[width=0.32\linewidth]{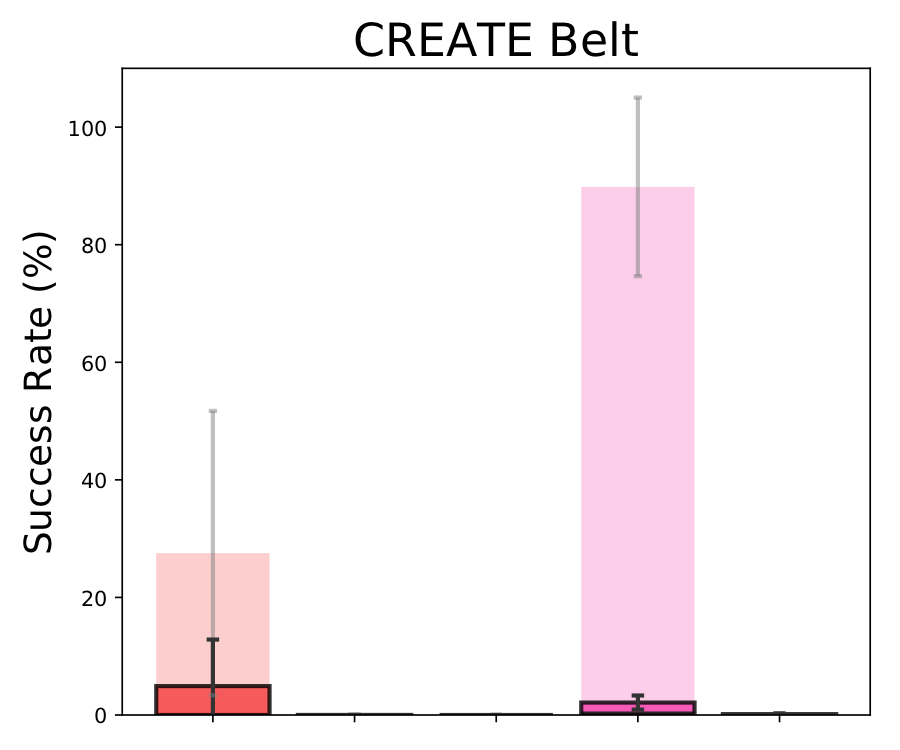}
      \includegraphics[width=0.32\linewidth]{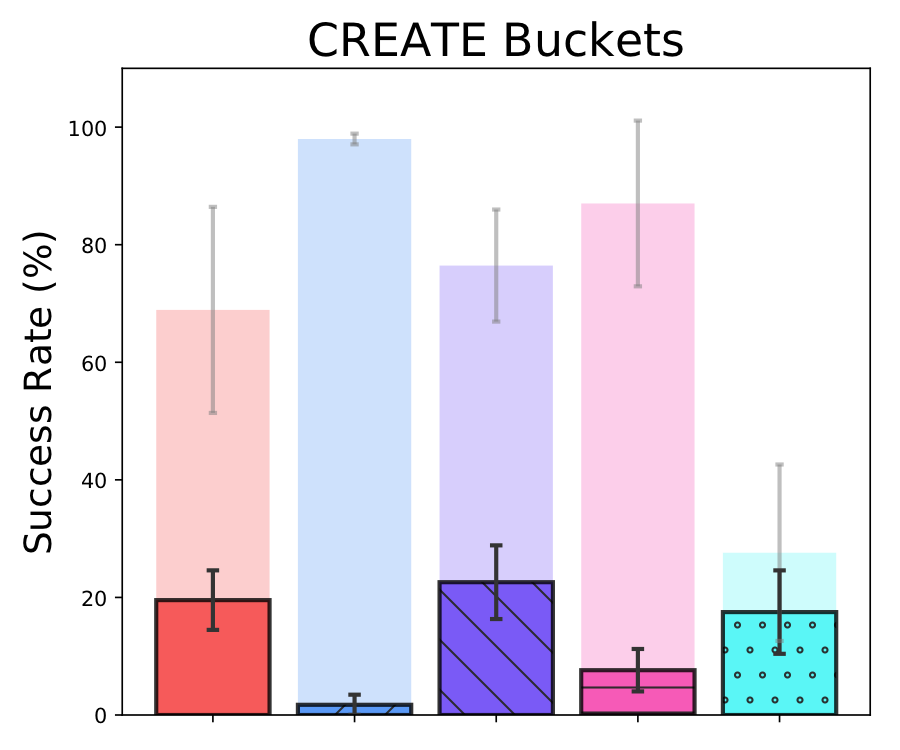}
      \includegraphics[width=0.32\linewidth]{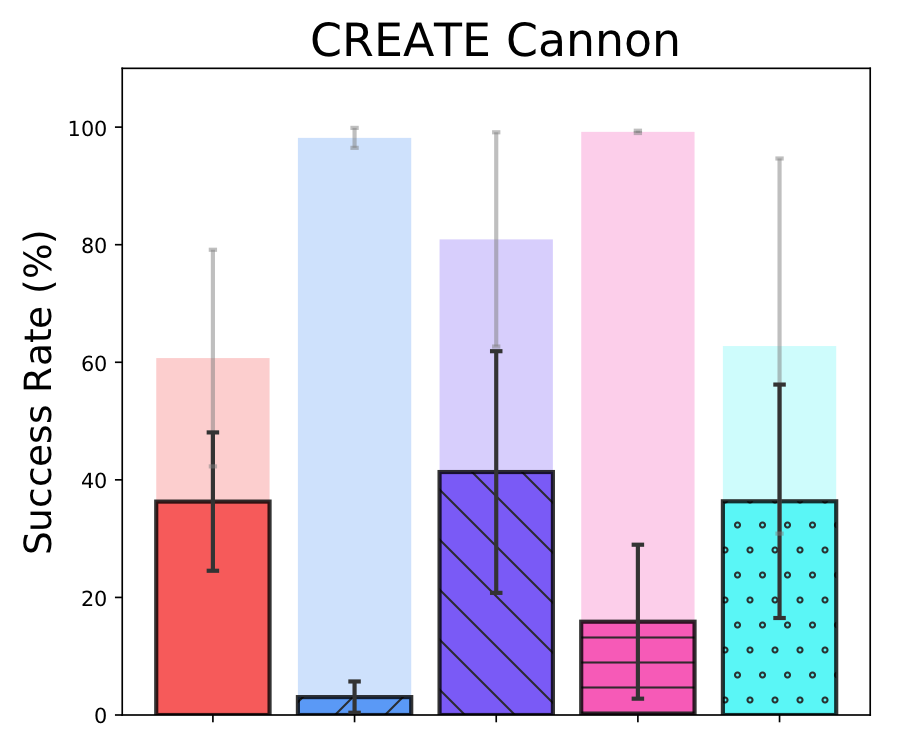}
      \includegraphics[width=0.32\linewidth]{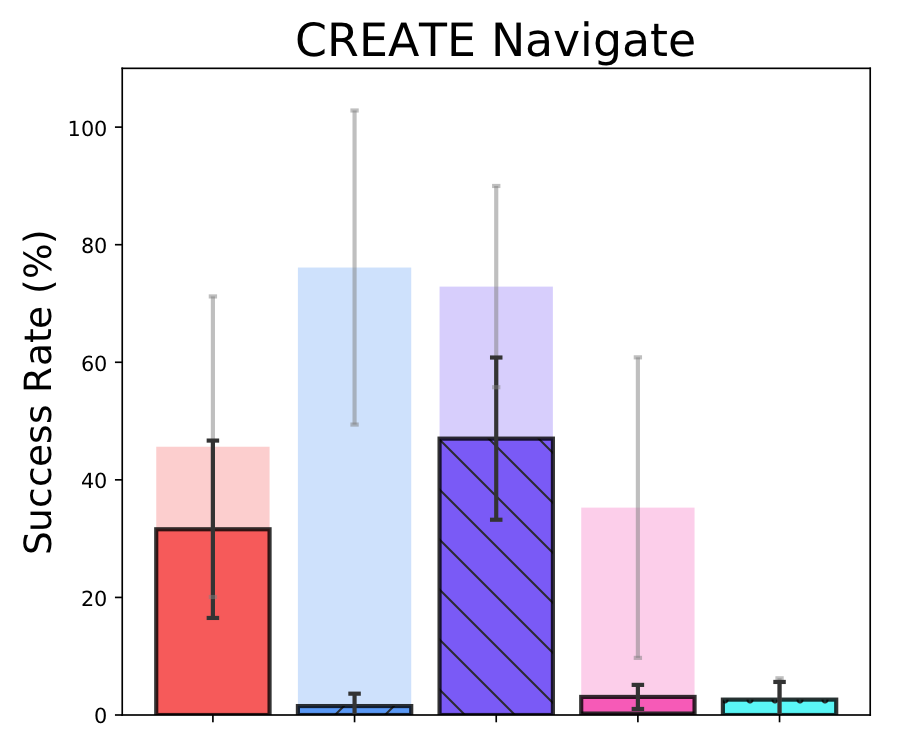}
      \includegraphics[width=0.32\linewidth]{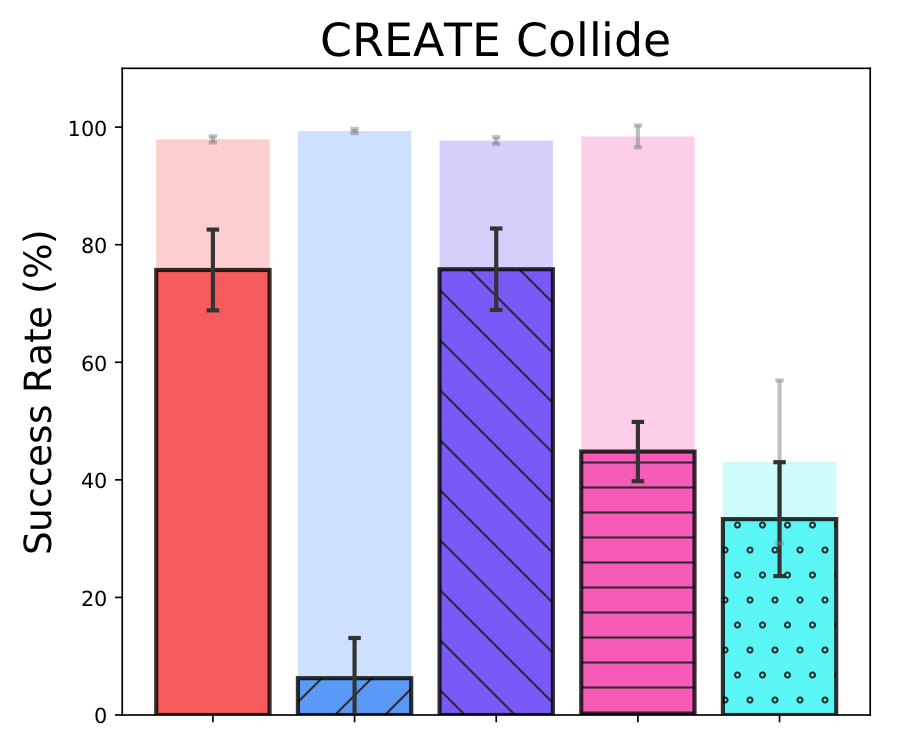}
      \includegraphics[width=0.32\linewidth]{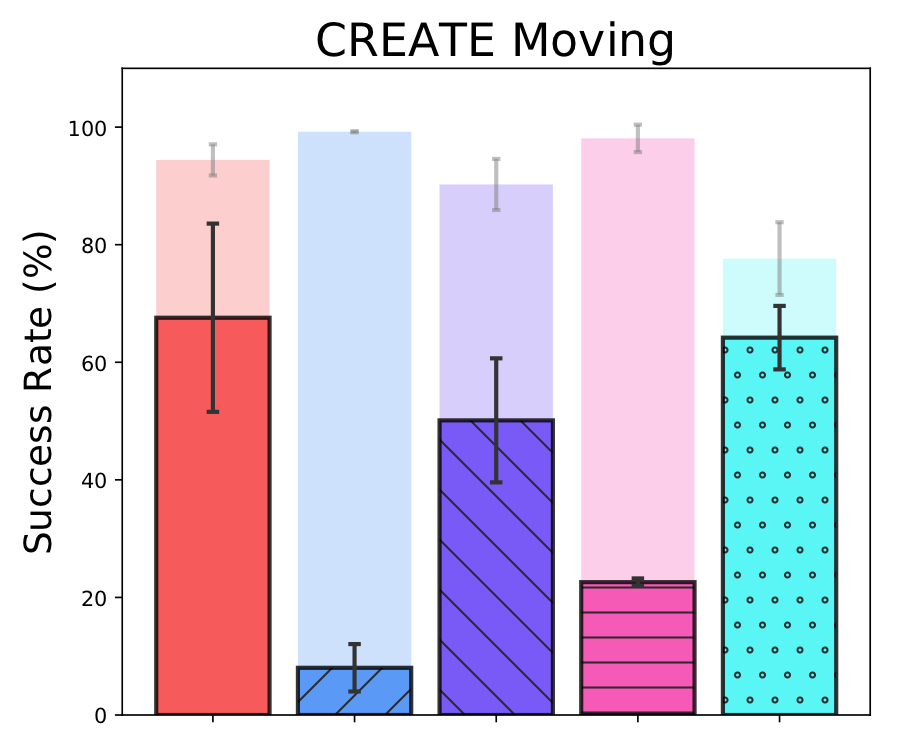}
      \includegraphics[width=0.32\linewidth]{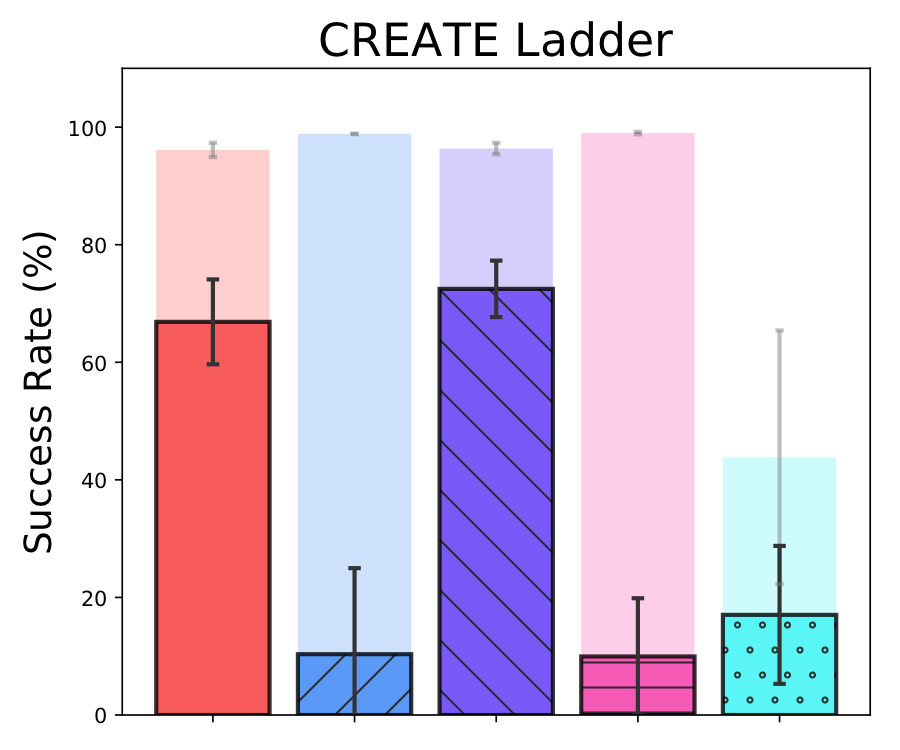}
      \includegraphics[width=0.32\linewidth]{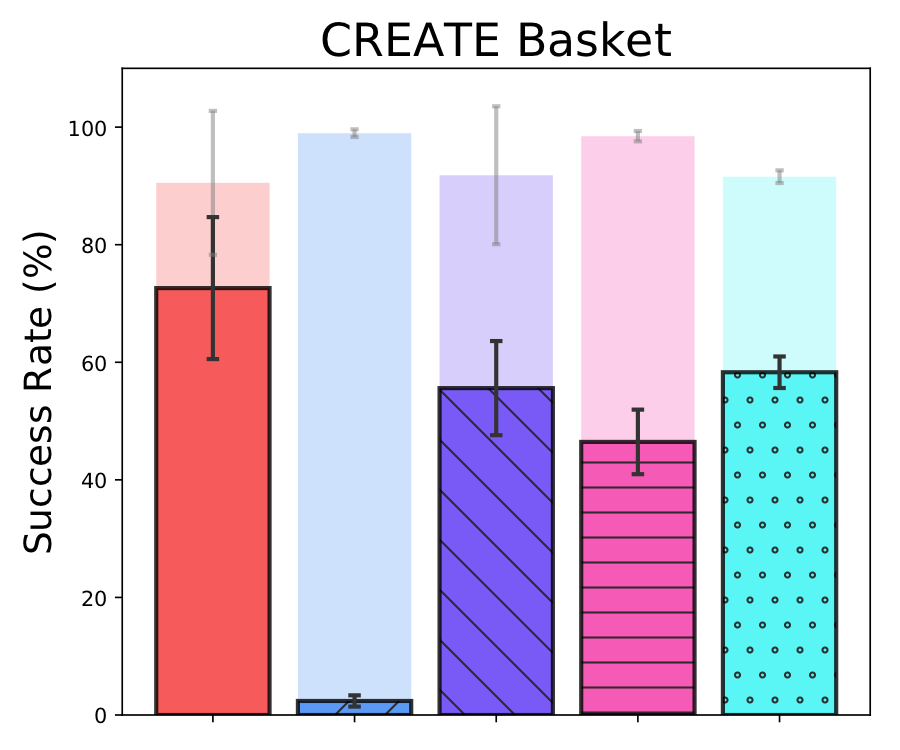}
      \includegraphics[width=0.32\linewidth]{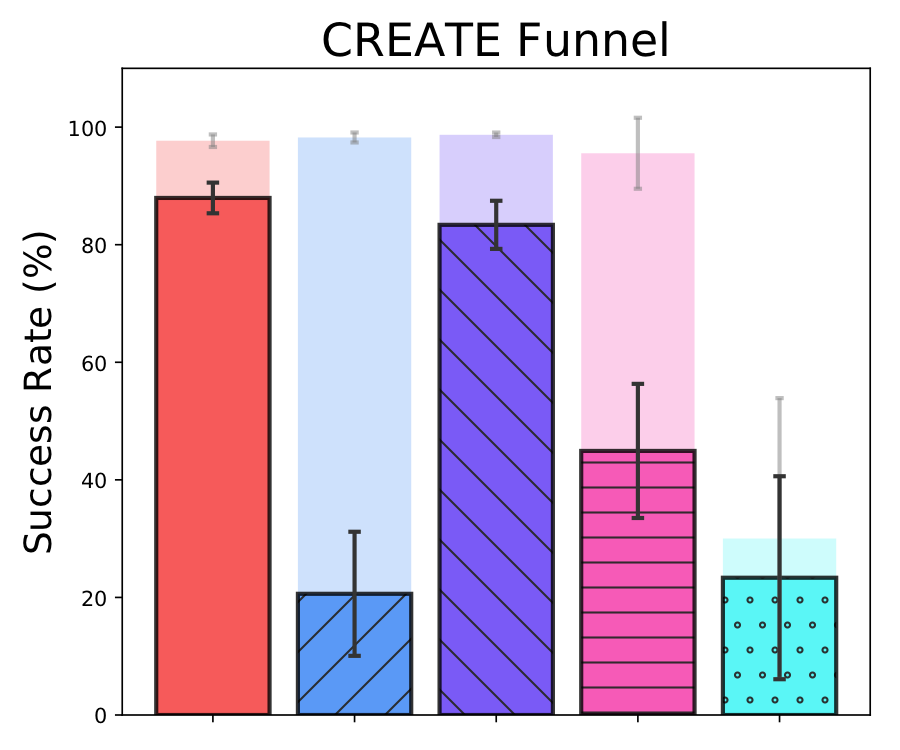}
      \includegraphics[width=0.8\linewidth]{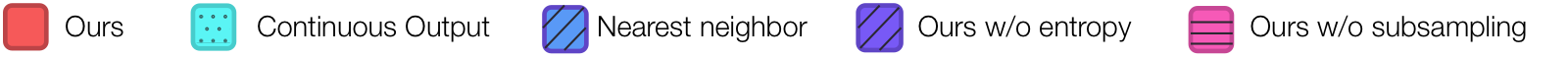}
    \end{subfigure}
    \caption{
        Results on the remaining 9 CREATE tasks with the same evaluation details as the main paper (Figure~\ref{fig:baselines}). We compare our method against all the baselines (Section~\ref{sec:baselines}) and ablations (Section~\ref{sec:ablations}).
      }
    \label{supp:fig:add_create}
\end{figure*}

\begin{figure*}[!ht]
    \centering
    \begin{subfigure}[t]{0.19\textwidth}
    	\includegraphics[width=\textwidth]{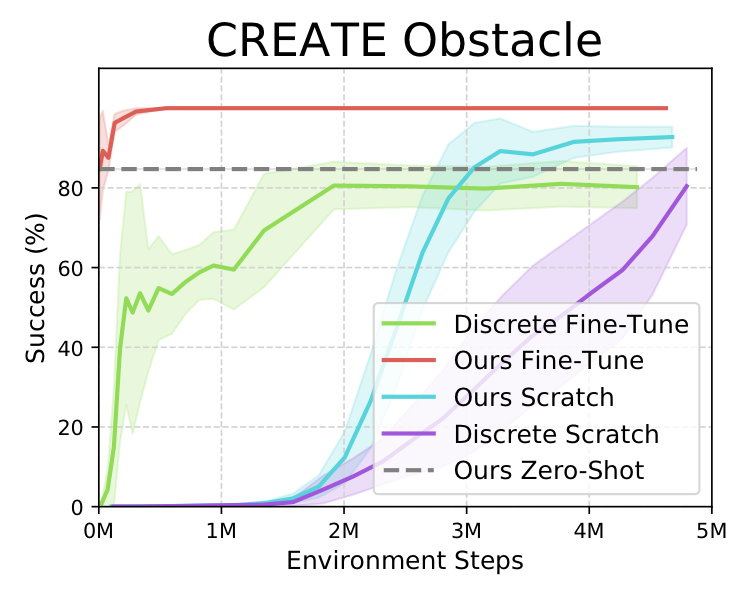}
    \end{subfigure}
    \begin{subfigure}[t]{0.19\textwidth}
    	\includegraphics[width=\textwidth]{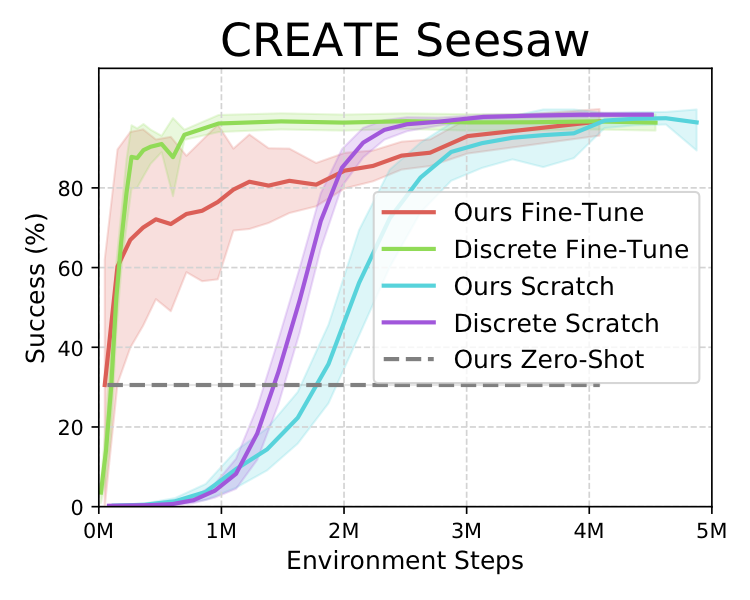}
    \end{subfigure}
    \begin{subfigure}[t]{0.19\textwidth}
    	\includegraphics[width=\textwidth]{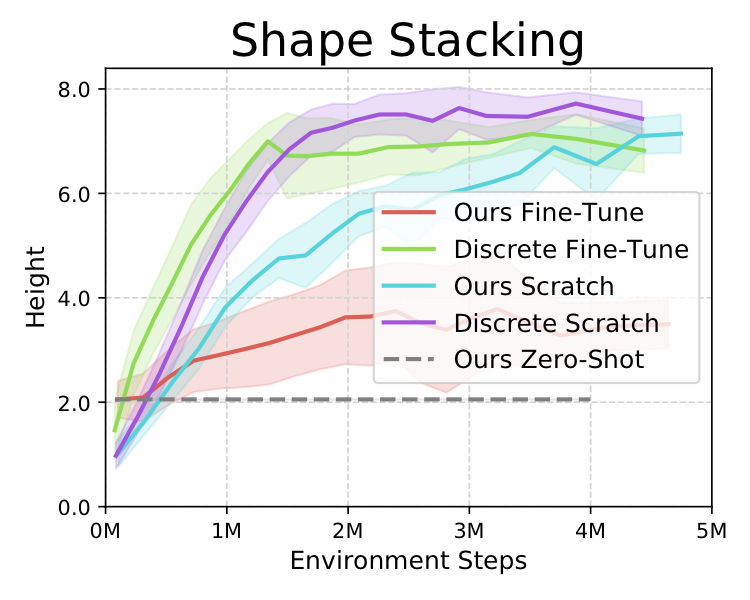}
    \end{subfigure}
    \begin{subfigure}[t]{0.19\textwidth}
    	\includegraphics[width=\textwidth]{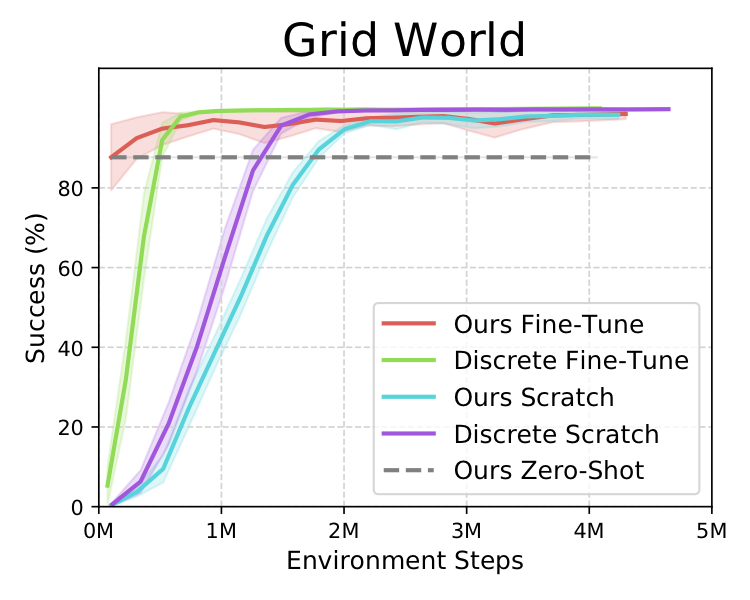}
    \end{subfigure}
    \begin{subfigure}[t]{0.19\textwidth}
    	\includegraphics[width=\textwidth]{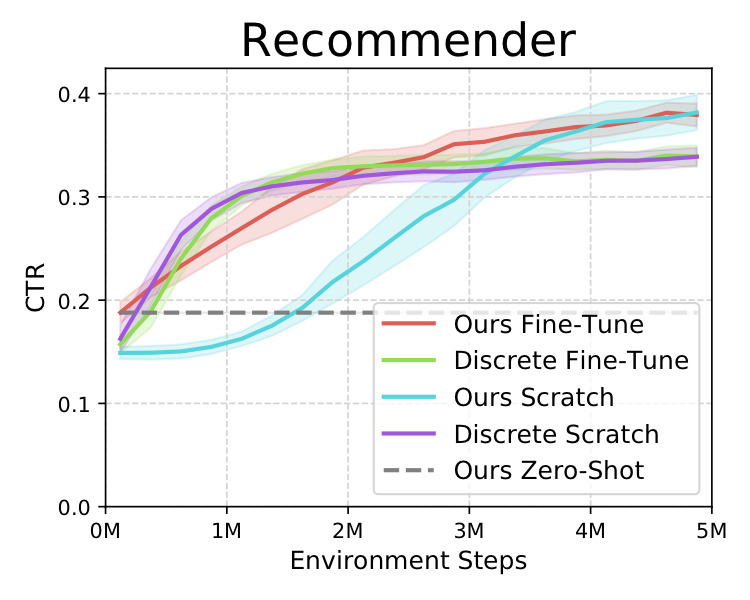}
    \end{subfigure}
    \begin{subfigure}[t]{0.9\textwidth}
      \includegraphics[width=\textwidth]{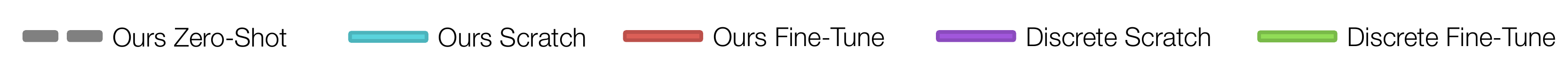}
    \end{subfigure}
    \vspace{-10pt}
    \caption{
      Finetuning or training the policy from scratch on the new action space across the remaining 5 tasks (Figure~\ref{fig:fine_tune} only shows results on CREATE Push). The evaluation settings are the same as described in Section~\ref{sec:fine_tune}.
    }
    \label{supp:fig:all_ft}
\end{figure*}

\textbf{Grid World}: Figure~\ref{supp:fig:grid_representations} shows the inferred action or skill representations in Grid World. The actions are colored by the relative change in the location of the agent after applying the skill. For example, the skill "Up,
Up, Up, Right, Down" would translate the agent to the upper right
quadrant from the origin, hence visualized in red color.
All learned action representations are 16-dimensional.
We plot the following action representations:
\begin{itemize}
[leftmargin=*, noitemsep, topsep=0pt]
    \item State Trajectories (default): HVAE encodes action observations consisting of trajectories of 2D $(x,y)$ coordinates of the agent on the 80x80 grid.
    \item Non-Hierarchical VAE (baseline): A standard VAE encodes all the state-based action observations individually, and then computes the action representation by taking their mean.
    \item One-hot (alternate): State is represented by two 80-dimensional one-hot vectors of the agent's x and y coordinates on the 80x80 grid. Reconstruction is based on a softmax cross-entropy loss over the one-hot observations in the trajectory.
    \item Engineered (alternate): These are 5-dimensional representations containing the ground-truth knowledge of the five moves (up, down, left, right) that constitute a skill. The clustering of our learned representations looks comparable to these oracle representations.
\end{itemize}

\textbf{CREATE}: Figure~\ref{supp:fig:create_representations} shows the inferred action or tool representations in CREATE. The actions are colored by tool class. All action representations are 128-dimensional.
\begin{itemize}
  \item State Trajectories (default): HVAE encodes action observation data composed of $(x,y)$ coordinate states of the probe ball's trajectory.
  \item Non-Hierarchical VAE (baseline): A standard VAE encodes all the state-based action observations individually, and then computes the action representation by taking their mean.
  \item Video (alternate): HVAE encodes action observation data composed of 84x84 grayscale image-based trajectories (videos) of the probe ball interacting with the tool. The data is collected identically as the state case, only the modality changes from state to image frames.
  \item Full Split: HVAE encodes state-based action observations, however, the training and testing tools are from the \textit{Full Split} experiment. The visualizations show that even though training tools are vastly different from evaluation tools, HVAE generalizes and clusters well on unseen tools.
\end{itemize}

\textbf{Shape Stacking}: Figure~\ref{supp:fig:stack_representations} shows the
inferred action representations in Shape Stacking. The shapes are colored
according to shape class. All action representations are 128-dimensional.
\begin{itemize}
  \item Viewpoints (default): HVAE encodes action observations in the form of viewpoints of the shape from different camera angles and positions.
  \item Non-Hierarchical VAE (baseline): A standard VAE encodes all the image-based action observations individually, and then computes the action representation by taking their mean.
  \item Full-split:
   The training and testing tools are from the \textit{Full Split} experiment. Previously unseen shape types are clustered well, showing the robustness of HVAE.
\end{itemize}

\section{Further Experimental Results}
\label{supp:exp}

\subsection{Additional CREATE Results}
\label{supp:exp:additional_create}

Figure~\ref{supp:fig:create_all_tasks} visually describes all the CREATE tasks. The objective is to make the target ball (red) reach the goal (green), which may be fixed or mobile.
Figure~\ref{supp:fig:add_create} demonstrates our method's results on the remaining nine
CREATE tasks (the initial three tasks are in Figure~\ref{fig:baselines},~\ref{fig:ablations}).
Strong training and testing performance on a majority of these tasks shows the robustness of our method.
The developed CREATE environment can be easily modified to generate more such tasks of varying difficulties. Due to the diverse set of tools and tasks, we propose CREATE and our results as a useful benchmark for evaluating action space generalization in reinforcement learning.

\subsection{Additional Finetuning Results}
\label{supp:exp:additional_finetune}
We present additional results of finetuning and training from scratch to adapt to unseen actions across all CREATE Obstacle, CREATE Seesaw, Shape Stacking, Grid World, and Recommender. In the results presented in Figure~\ref{supp:fig:all_ft}, we observe the same trend holds where additional training takes many steps to achieve the performance our method obtains zero-shot.

\subsection{CREATE: No Subgoal Reward}
\label{supp:exp:no_subgoal}
\begin{figure}[ht]
    \centering
    \includegraphics[width=0.7\textwidth]{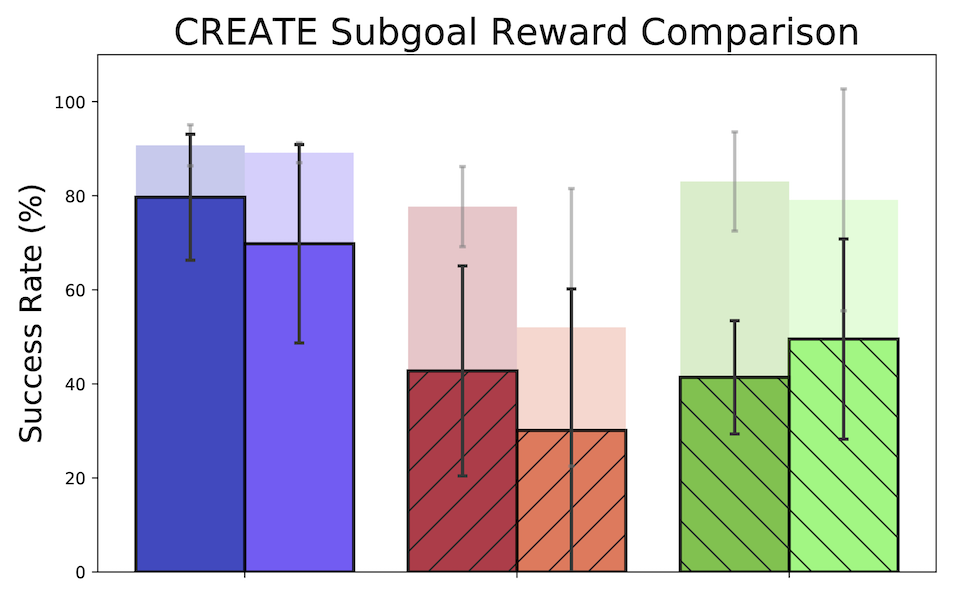}
    \includegraphics[width=0.6\textwidth]{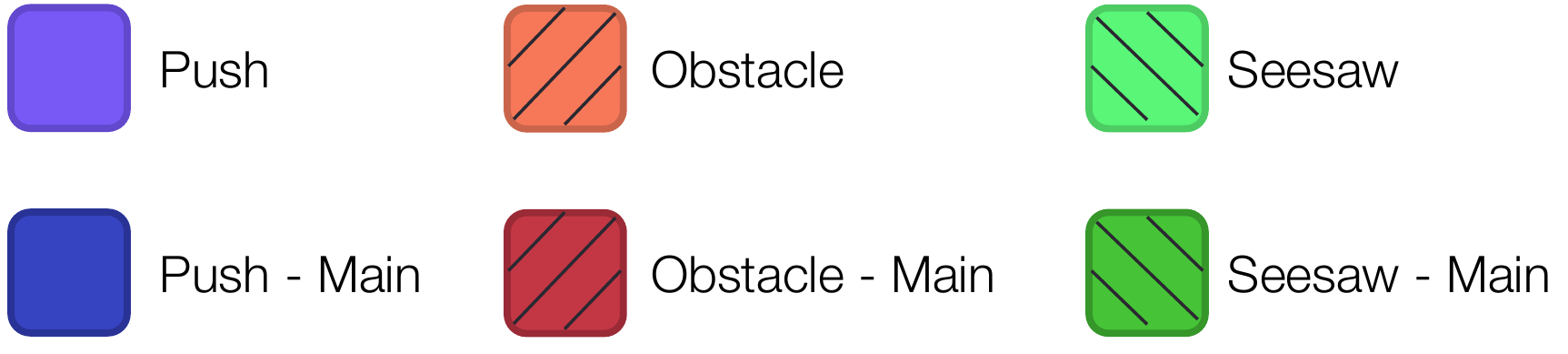}
    \caption{
      Comparison of a version of CREATE that does not use subgoal rewards.
     The ``Main" methods
    are from the main paper using subgoal rewards (Figure~\ref{fig:baselines}).  
    }
    \label{supp:fig:create_less}
\end{figure}

To verify our method's robustness, we also run experiments on a version of the CREATE environment without the subgoal rewards. The results in Figure~\ref{supp:fig:create_less} verify that even without reward engineering, our method exhibits strong generalization, albeit with higher variance in train and testing performance.   

\subsection{Auxiliary Policy Alternative Architecture}
\label{supp:exp:auxiliary_alternate}
\begin{figure}[ht]
    \centering
    \includegraphics[width=0.7\textwidth]{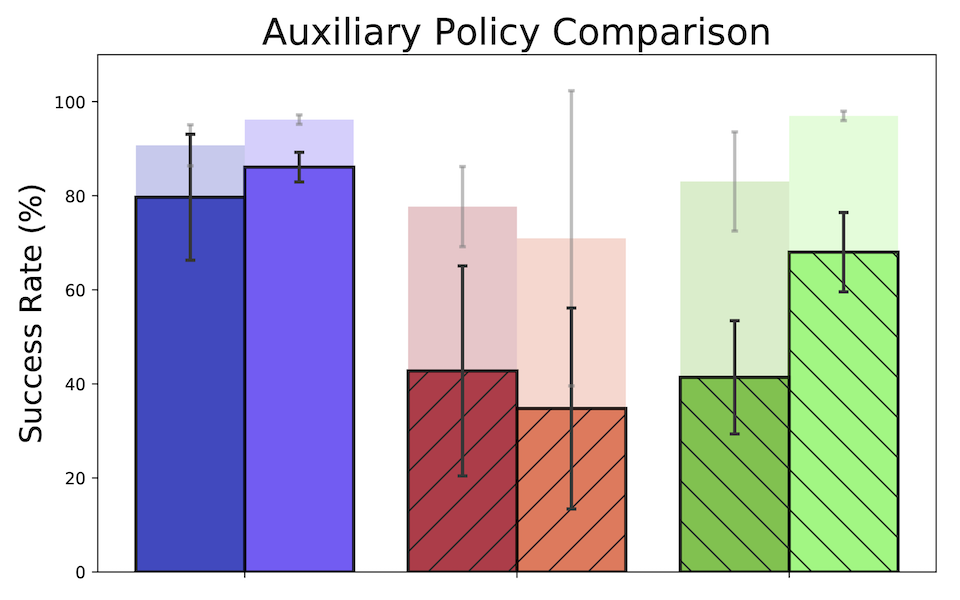}
    \includegraphics[width=0.6\textwidth]{supplementary_images/create_legend.pdf}
    \caption{
    Comparison of an alternative auxiliary network architecture that is
    conditioned on the selected discrete action. The ``Main" results are
    the default results that do not condition the auxiliary policy on the selected action (Figure~\ref{fig:baselines}). 
    }
    \label{supp:fig:create_condaux}
\end{figure}

While in our framework, the auxiliary policy is computed from the state encoding alone, here we compare to also taking the selected discrete-action as input to the auxiliary policy. Comparison of this alternative auxiliary policy to the auxiliary policy from the main paper is shown in Figure~\ref{supp:fig:create_condaux}. There are minimal differences in the average success rates of the two design choices.

\subsection{Fully Observable Recommender System}
\label{supp:sec:fully_obs_reco}
\begin{figure}[ht]
    \centering
	\includegraphics[width=0.8\textwidth]{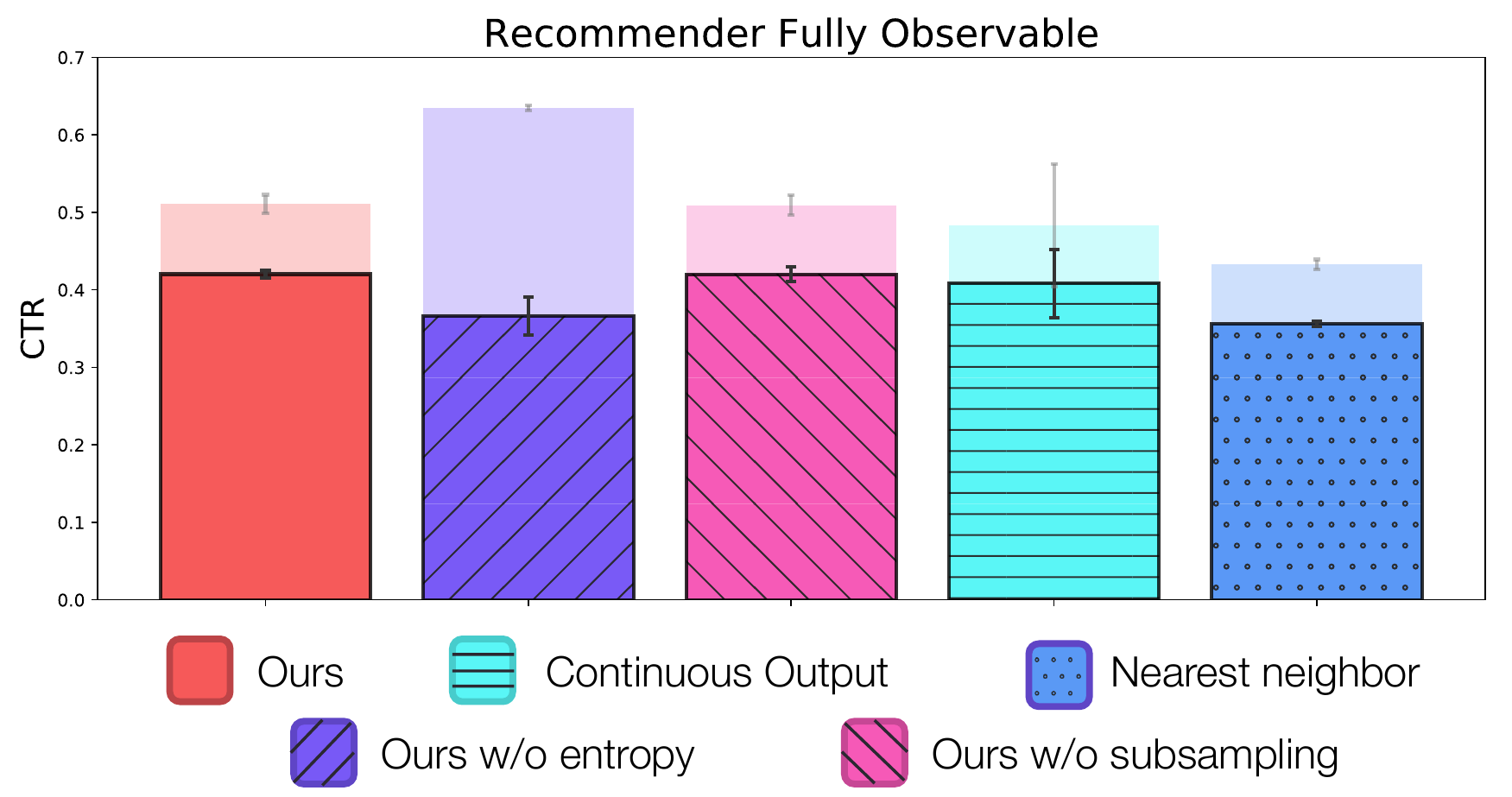}
    \caption{
       Training and testing results on the fully observable version of Recommender System with standard evaluation settings.
    }
    \label{supp:fig:add_reco}
\end{figure}

Figure~\ref{supp:fig:add_reco} demonstrates our method in a fully observable recommender
environment where the product constant $\mu_i$ from Eq.~\ref{supp:eq:reco_click} is also included in the engineered action representation. All methods achieve better
training and generalization performance compared to the original partially observable Recommender System environment. However, full observability is infeasible in practical recommender systems. Therefore, we focus on the partially observed environment in the main results.

\begin{figure}[ht]
    \centering
    \includegraphics[width=\textwidth]{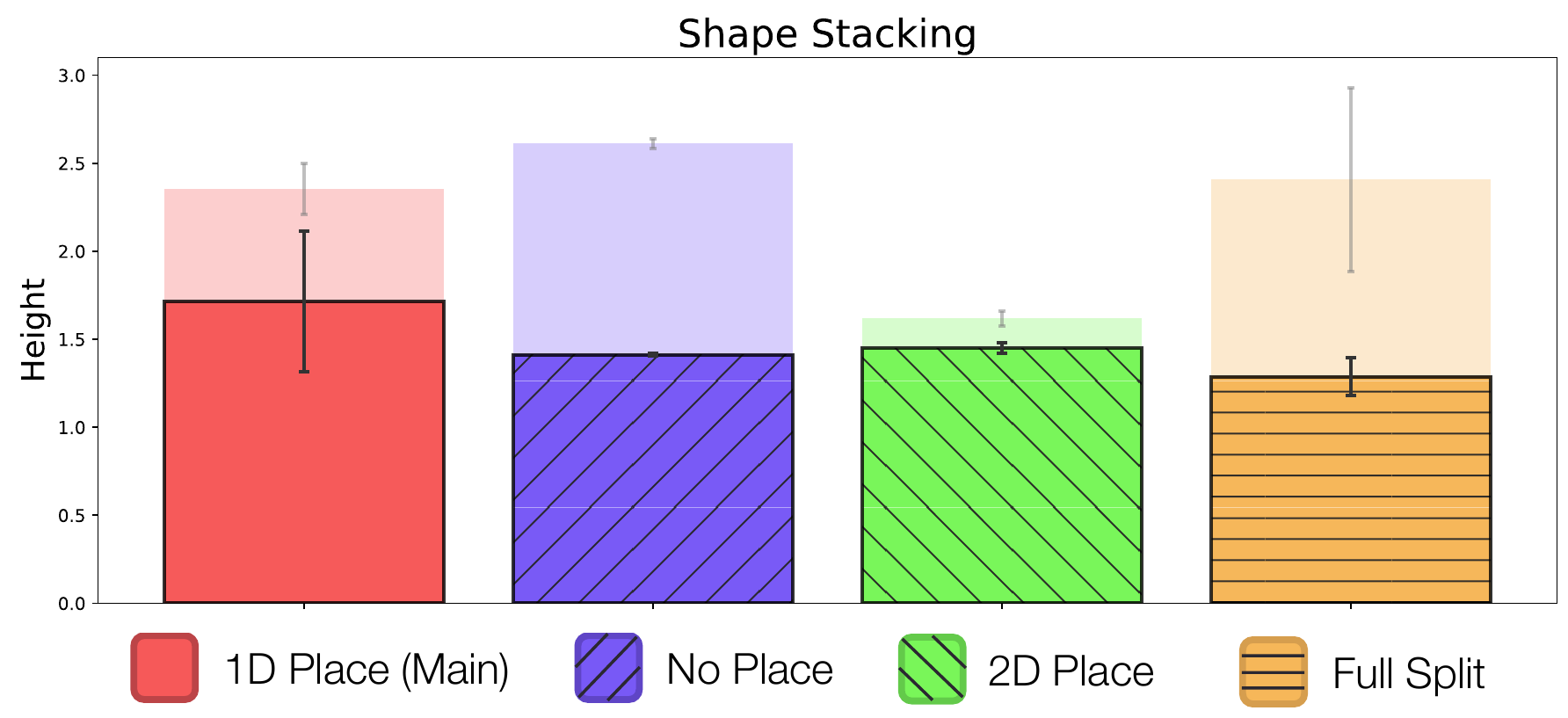}
    \caption{
      Comparing different placement strategies in shape stacking and showing
      performance on the \textit{Full Split} action split. Results are using
      our method with the standard evaluation details. 
    }
    \label{supp:fig:add_stack}
\end{figure}
\subsection{Additional Shape Stacking Results}
\label{supp:exp:additional_shape}
Figure~\ref{supp:fig:add_stack} demonstrates performance on different shape placement strategies in Shape Stacking using our framework.
In \textit{No Place}, the shapes are dropped at the center of the table, and the agent only selects which shape to drop from the available set.
Since there are two randomly placed cylinders on the table, this setting of dropping in the center gives less control to the agent while stacking tall towers.
Thus we report default results on \textit{1D Place}, where the agent outputs in a hybrid action space consisting of shape selection and 1D placement through $x$-coordinate of the dropping location. The $y$-coordinate of the drop is fixed to the center.
Finally, in \textit{2D Place}, the agent decides both $x$ and $y$ coordinates to have more control but makes the task more challenging due to the larger search space.
The evaluation videos of these new settings are available on
\url{https://sites.google.com/view/action-generalization/shape-stacking}.

Figure~\ref{supp:fig:add_stack} also shows the results of our method trained and evaluated on \textit{Full Split} which was introduced in Table~\ref{supp:tab:ss_full_split}. Poor performance on this split could be explained by the policy not seeing enough shape classes during training to be able to generalize well to new shape classes during testing. This is also expected since this split severely breaks the i.i.d. assumption essential for generalization~\citep{bousquet2003introduction}.

\subsection{Learning Curves}
\begin{figure*}[ht]
    \centering
    \begin{subfigure}[t]{1\textwidth}
    	\centering
      \includegraphics[width=0.16\linewidth]{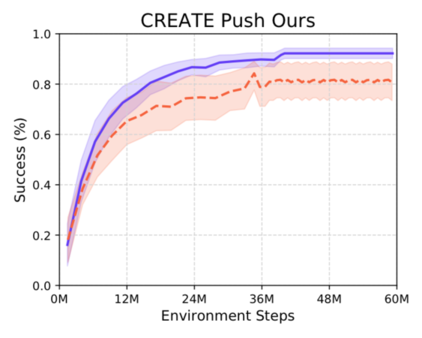}
    	\centering
      \includegraphics[width=0.16\linewidth]{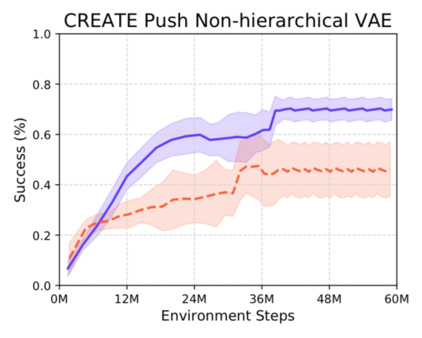}
    	\centering
      \includegraphics[width=0.16\linewidth]{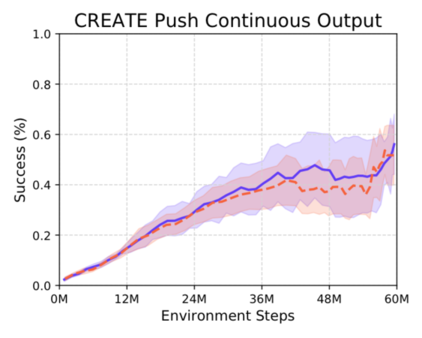}
    	\centering
      \includegraphics[width=0.16\linewidth]{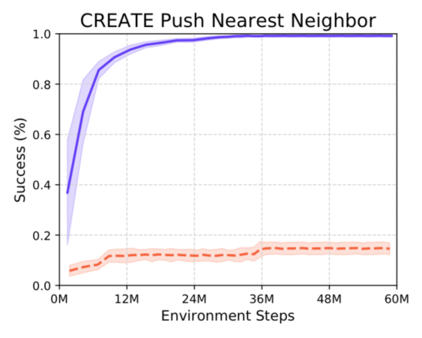}
    	\centering
      \includegraphics[width=0.16\linewidth]{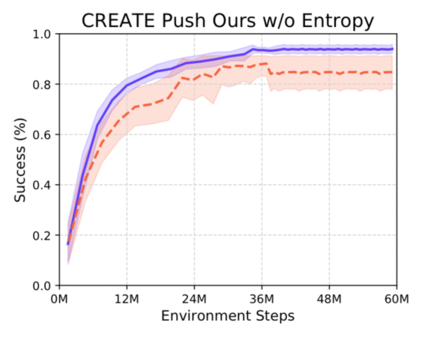}
    	\centering
      \includegraphics[width=0.16\linewidth]{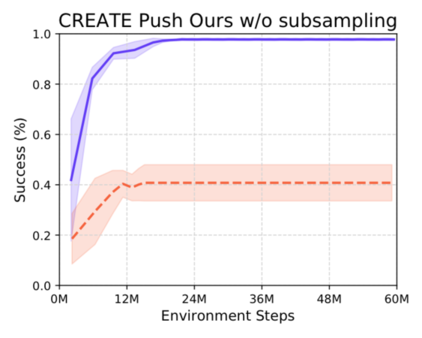}
    \caption{CREATE Push}
    \end{subfigure}

    \begin{subfigure}[t]{\textwidth}
    	\centering
      \includegraphics[width=0.16\linewidth]{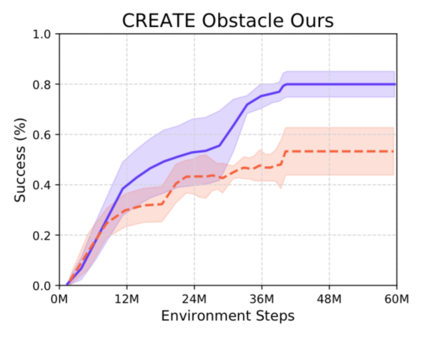}
      \includegraphics[width=0.16\linewidth]{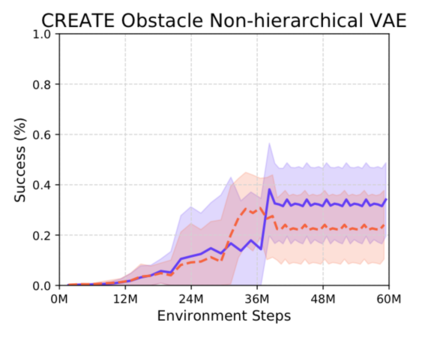}
      \includegraphics[width=0.16\linewidth]{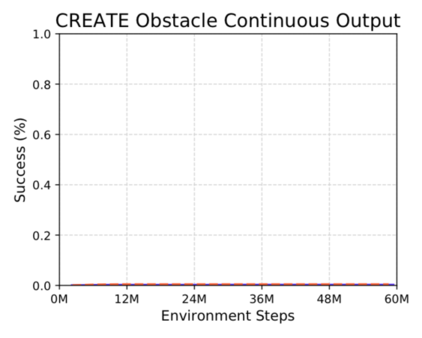}
      \includegraphics[width=0.16\linewidth]{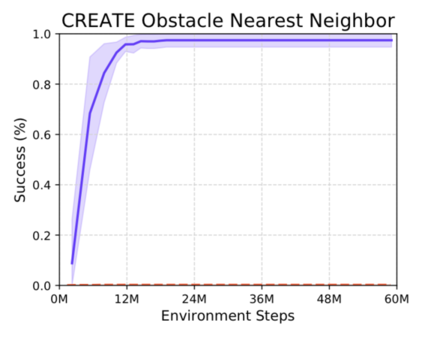}
      \includegraphics[width=0.16\linewidth]{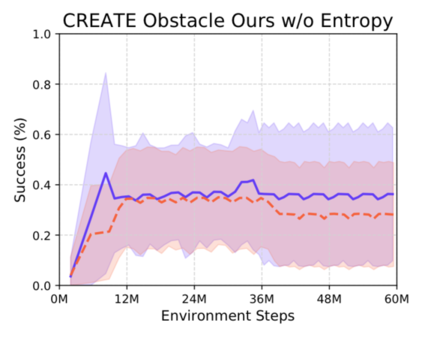}
      \includegraphics[width=0.16\linewidth]{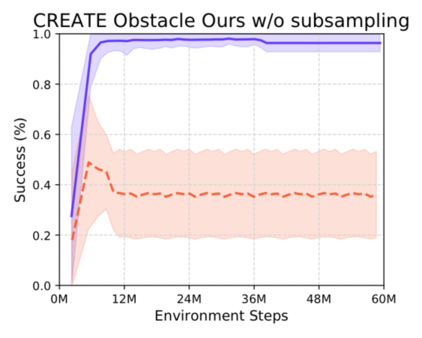}
    \caption{CREATE Obstacle}
    \end{subfigure}

    \begin{subfigure}[t]{\textwidth}
    	\centering
      \includegraphics[width=0.16\linewidth]{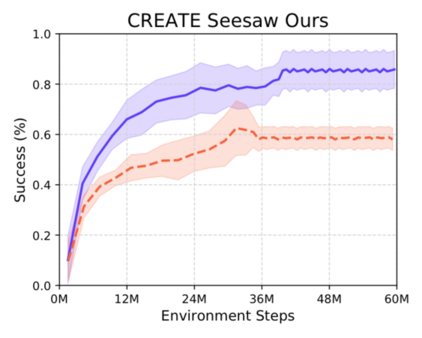}
      \includegraphics[width=0.16\linewidth]{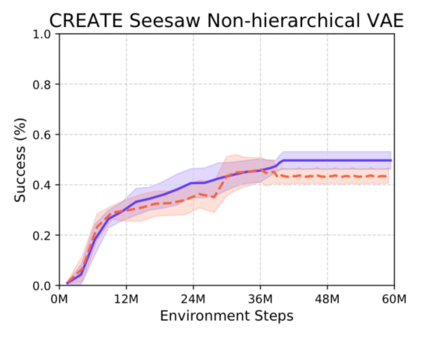}
      \includegraphics[width=0.16\linewidth]{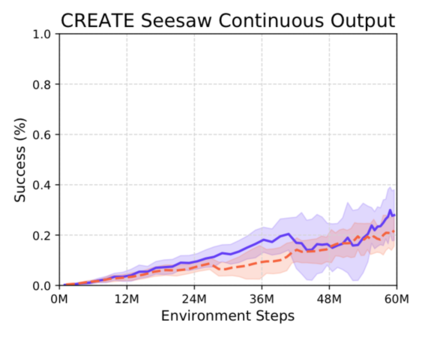}
      \includegraphics[width=0.16\linewidth]{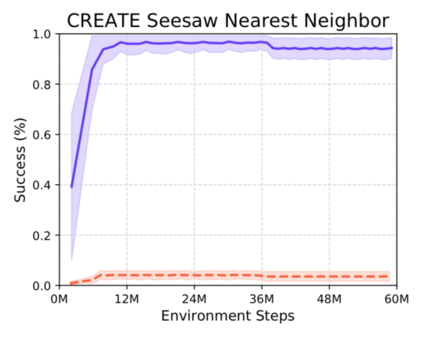}
      \includegraphics[width=0.16\linewidth]{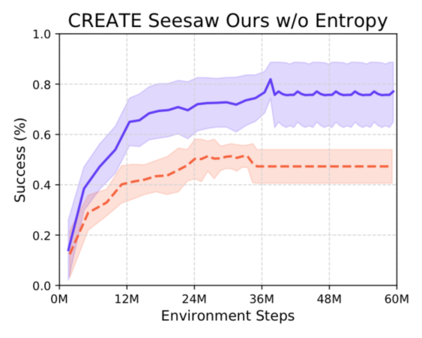}
      \includegraphics[width=0.16\linewidth]{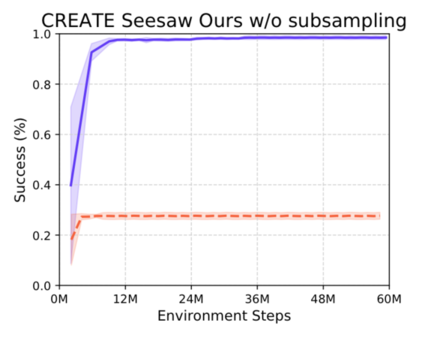}
    \caption{CREATE Seesaw}
    \end{subfigure}

    \begin{subfigure}[t]{\textwidth}
    	\centering
      \includegraphics[width=0.16\linewidth]{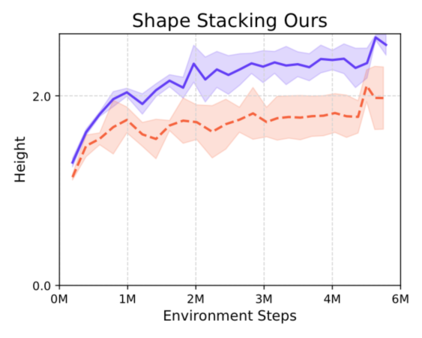}
      \includegraphics[width=0.16\linewidth]{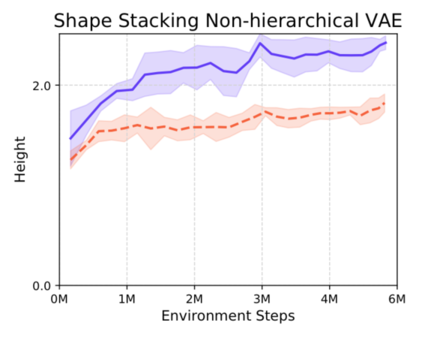}
      \includegraphics[width=0.16\linewidth]{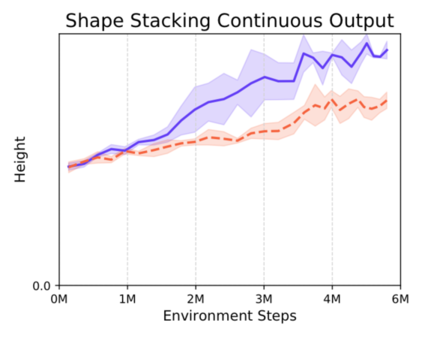}
      \includegraphics[width=0.16\linewidth]{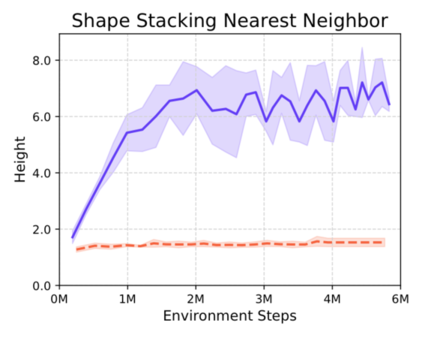}
      \includegraphics[width=0.16\linewidth]{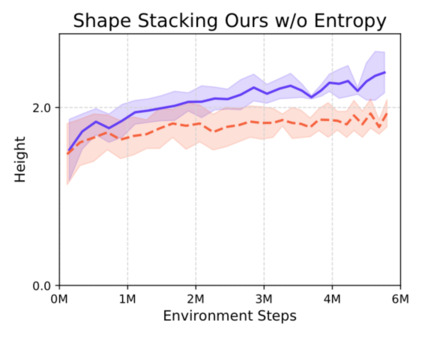}
      \includegraphics[width=0.16\linewidth]{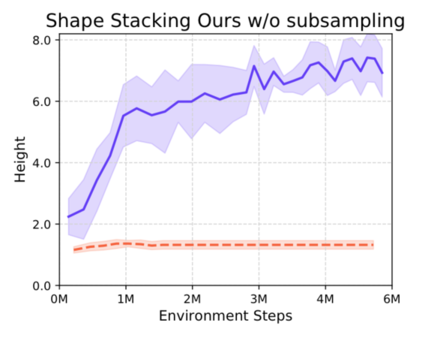}
    \caption{Shape Stacking}
    \end{subfigure}

    \begin{subfigure}[t]{\textwidth}
    	\centering
      \includegraphics[width=0.16\linewidth]{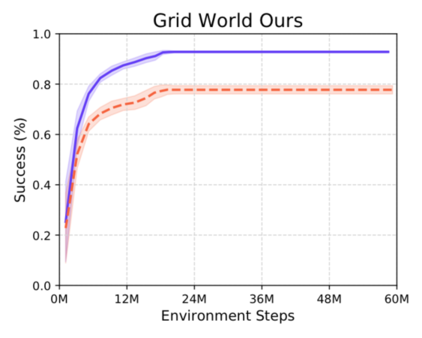}
      \includegraphics[width=0.16\linewidth]{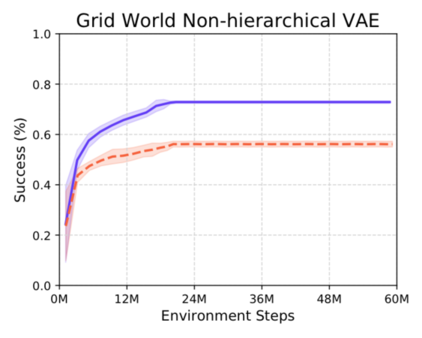}
      \includegraphics[width=0.16\linewidth]{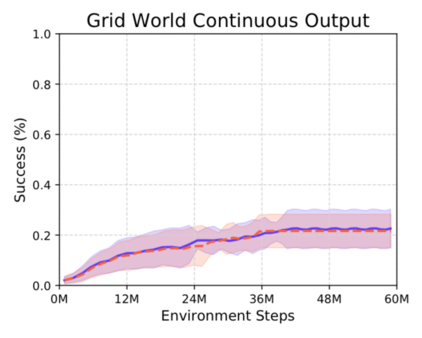}
      \includegraphics[width=0.16\linewidth]{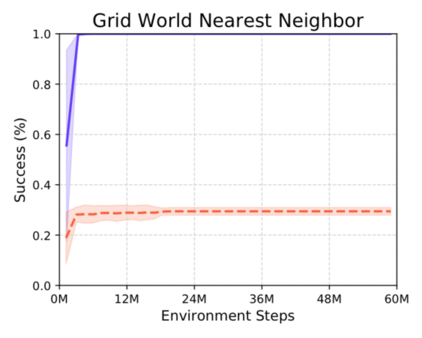}
      \includegraphics[width=0.16\linewidth]{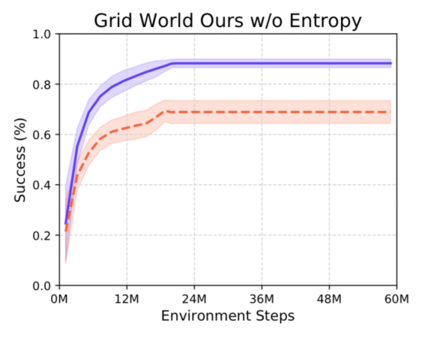}
      \includegraphics[width=0.16\linewidth]{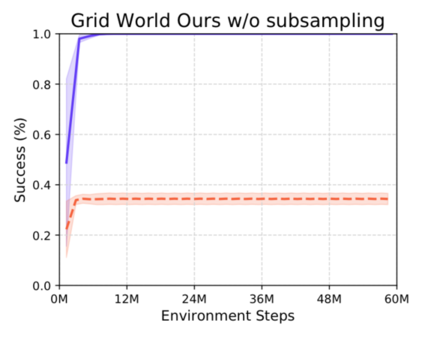}
    \caption{Grid World}
    \end{subfigure}

    \begin{subfigure}[t]{\textwidth}
    	\centering
      \includegraphics[width=0.16\linewidth]{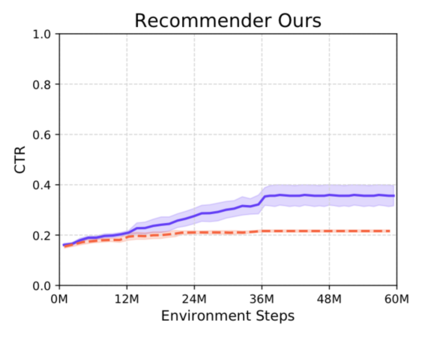}
      \includegraphics[width=0.16\linewidth]{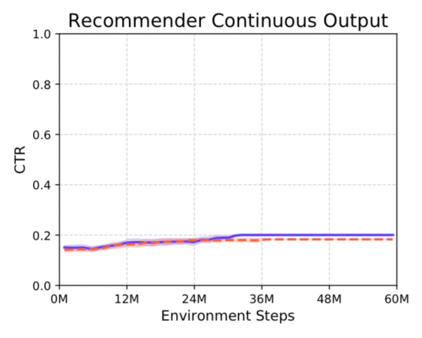}
      \includegraphics[width=0.16\linewidth]{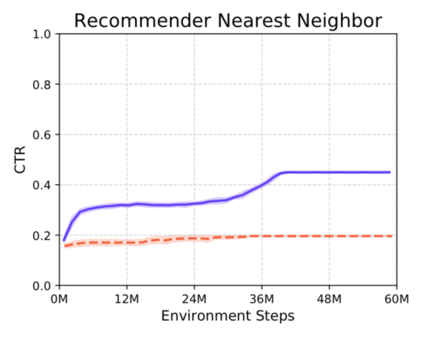}
      \includegraphics[width=0.16\linewidth]{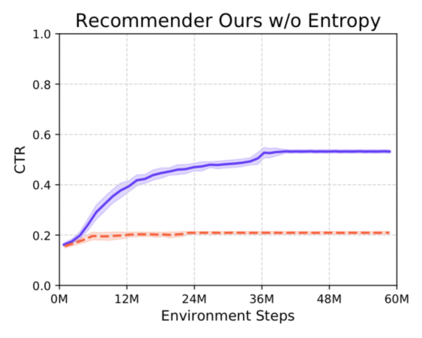}
      \includegraphics[width=0.16\linewidth]{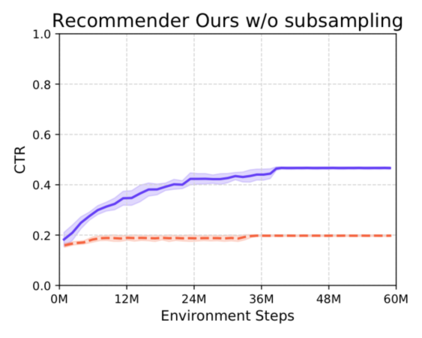}
    \caption{Recommender System}
    \end{subfigure}

    \begin{subfigure}[t]{0.3\textwidth}
    	\centering
      \includegraphics[width=\linewidth]{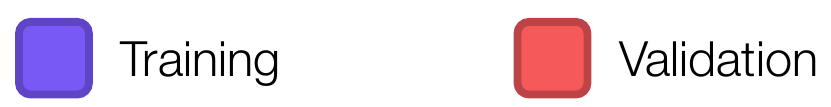}
    \end{subfigure}

    \caption{
      Learning curves for all environments and methods showing performance on both the training and validation
      sets. Each line shows the performance of 5 random seeds (8 for Grid World) as average value and the shaded region as the standard deviation. 
    }
    \label{supp:fig:train_curves}
\end{figure*}

Figure~\ref{supp:fig:train_curves} show the training and validation
performance curves for all methods and environments to contrast the training process
of a policy against the objective of generalization to new actions.
The plots clearly show how the generalization gap varies over the training of the policy.
Ablation curves (last two columns) for some environments depict that an increase in training performance corresponds to a drop in validation performance.
This is attributed to the policy overfitting to the training set of actions, which is often observed in supervised learning. Our proposed regularizing training procedure aims to avoid such overfitting.

\section{Experiment Details}
\label{supp:training}

\begin{table*}[ht]
\centering
\begin{tabular}{lllll}
Hyperparameter & Grid world & Recommender & CREATE & Shape Stacking \\
\midrule
\multicolumn{5}{c}{\textbf{HVAE}} \\
\rule{0pt}{3ex}action representation size & 16 & 16 & 128 & 128 \\
batch size & 128 & -  & 128 & 32 \\
epochs & 10000 & - & 10000 & 5000 \\

\midrule

\multicolumn{5}{c}{\textbf{Policy}} \\
\rule{0pt}{3ex}entropy coefficient & 0.05 & 0.01 & 0.005 & 0.01 \\
observation space & 81 & 16 & $84 \times 84 \times 3 $ & $84\times84\times4 $ \\
actions per episode & 50 & 500 & 50 & 20 \\
total environment steps & $4 \times 10^7$ & $4 \times 10^7$ & $6 \times 10^7$ & $3 \times 10^6$\\
max. episode length & 10 & 100 & 30 & 10 \\
continuous entropy scaling & - & - & 0.1 & 0.1 \\
PPO batch size & 4096 & 2048 & 3072 & 1024 \\

\end{tabular}
\caption{Environment-specific hyperparameters}
\label{supp:tab:hyperparams_specific}
\end{table*}

\subsection{Implementation}
We use PyTorch~\citep{paszke2017automatic} for our implementation, and the
experiments were primarily conducted on workstations with 72-core Intel Xeon
Gold 6154 CPU and 4 NVIDIA GeForce RTX 2080 Ti GPUs.  Each experiment seed
takes about 6 hours (Recommender) to 25 hours (CREATE) to 
converge.  For logging and tracking experiments, we use the Weights \& Biases
tool~\citep{wandb}.  All the environments were developed using the OpenAI Gym
interface~\citep{brockman2016openai}.  The HVAE implementation is based on the
PyTorch implementation of Neural Statistician~\citep{edwards2016towards}, and we
use RAdam optimizer~\citep{liu2019radam}.  For training the policy network, we
use PPO~\citep{PPO, pytorchrl} with the Adam optimizer~\citep{kingma2014adam}.
Further details can be found in the supplementary code.\footnote{Code available at \url{https://github.com/clvrai/new-actions-rl}}

\subsection{Hyperparameters}

The default hyperparameters shared across all environments are shown in Table~\ref{supp:tab:hyperparams_shared} and environment-specific hyperparameters are given in Table~\ref{supp:tab:hyperparams_specific}. We perform linear decay of the learning rate over policy training.

\begin{table}[ht]
\centering
\begin{tabular}{ll}
Hyperparameter & Value \\
\midrule
\multicolumn{2}{c}{\textbf{HVAE}} \\
\rule{0pt}{3ex}learning rate & 0.001 \\
action observations & 1024 \\
MLP hidden layers & 3 \\
$q_\phi$ hidden layer size & 128 \\
default hidden layer size & 64 \\

\midrule
\multicolumn{2}{c}{\textbf{Policy}} \\
\rule{0pt}{3ex}learning rate & 0.001 \\
discount factor & 0.99 \\
parallel processes & 32 \\
hidden layer size & 64 \\
value loss coefficient & 0.5 \\
PPO epochs & 4 \\
PPO clip parameter & 0.1
\end{tabular}
\caption{General Hyperparameters}
\label{supp:tab:hyperparams_shared}
\end{table}

\subsubsection{Hyperparameter Search}

Initial HVAE hyperparameters were inherited from the implementation of \citet{edwards2016towards} and PPO hyperparameters from \citet{pytorchrl}.
The hyperparameters were finetuned to optimize the performance on the held-out validation set of actions.
Certain hyperparameters were sensitive to the environment or the method being trained and were searched for more carefully.

Specifically, entropy coefficient is a sensitive parameter to appropriately balance the ease of reward maximization during training versus the generalizability at evaluation. For each method and environment, we searched for entropy coefficients in subsets of $\{0.0001, 0.001, 0.005, 0.01, 0.05, 0.1 \}$, and selected the best parameter based on the performance on the validation set.
We found PPO batch size to be an important parameter affecting the speed of convergence, convergence value, and variance across seeds. Thus, we searched for the best value in $\{1024, 2048, 3072, 4096\}$ for each environment.
Total environment steps are chosen so all the methods and baselines can run until convergence.

\subsection{Network Architectures}
\subsubsection{Hierarchical VAE}
\label{supp:training:network:HVAE}

\textbf{Convolutional Encoder}:
When the action observation data is in image or video form, a convolution encoder is applied to encode it into a latent state or state-trajectory. Specifically, for CREATE video case, each action observation is a 48x48 grayscale video. Thus, each frame of the video is encoded through a 7-layer convolutional encoder with batch norm~\citep{ioffe2015batch}. Similarly, for Shape Stacking, the action observation is an 84x84 image, that is encoded through 9 convolutional layers with batch norm.

\textbf{Bi-LSTM Encoder}:
When the data is in trajectory form (as in CREATE and Grid World), the sequence of states are encoded through a 2-layer Bi-LSTM encoder. For CREATE video case, the encoded image frames of the video are passed through this Bi-LSTM encoder in place of the raw state vector.
After this step, each action observation is in the form of a 64-dimensional encoded vector.

\textbf{Action Inference Network}:
The encoded action observations are passed through a 4-layer MLP with ReLU activation, and then aggregated with mean-pooling. This pooled vector is passed through a 3-layer MLP with ReLU activation, and then 1D batch-norm is applied. This outputs the mean and log-variance of a Gaussian distribution $q_\phi$, which represents the entire action observation set, and thus the action. This is then used to sample an action latent to condition reconstruction of individual observations.

\textbf{Observation Inference Network}:
The action latent and individual encoded observations are both passed through linear layers and then summed up, and followed by a ReLU nonlinearity. This combined vector is then passed through two 2-layer MLPs with ReLU followed by a linear layer, to output the mean and log-variance of a Gaussian distribution, representing the individual observation conditioned on the action latent. This is used to sample an observation latent, which is later decoded back while being conditioned on the action latent.

\textbf{Observation Decoder}:
The sampled observation latent and its action latent are passed through linear layers, summed and then followed by a nonlinearity.
For non-trajectory data (as in Shape Stacking), this vector is then passed through a 3-layer MLP with ReLU activation to output the decoded observation's mean and log-variance (i.e. a Gaussian distribution).
For trajectory data (as in CREATE and Grid World),
the initial ground truth state of the trajectory is first encoded with a 3-layer MLP with ReLU. Then an element-wise product is taken with the action-observation combined vector. The resulting vector is then passed through an LSTM network to produce the latents of future states of the trajectory. Each future state latent of the trajectory goes through a 3-layer MLP with ReLU, to result in the mean and log-variance of the decoded trajectory observation (i.e. a Gaussian distribution).

\textbf{Convolutional Decoder}:
If the observation was originally an image or video, then the mean of the reconstructed observation is converted into pixels through a convolutional decoder consisting of 2D convolutional and transposed-convolutional layers.
For the case of video input, the output of the convolutional decoder is also channel-wise augmented with with a 2D pixel mask. This mask is multiplied with the mean component of the image output (i.e. log-variance output stays the same), and then added to the initial frame of the video. This is the temporal skip connection technique~\citep{ebert2017self}, which eases the learning process with high-dimensional video observation datasets.

Finally, the reconstruction loss is computed using the Gaussian log-likelihood of the input observation data with respect to the decoded distribution.

\subsubsection{Policy Network}
\textbf{State Encoder $f_\omega$}:
When the input state is in image-form (channel-wise stacked frames in CREATE and Stacking), $f_\omega$ is implemented with a 5-layer convolutional network, followed by a linear layer and ReLU activation function.
When the input is not an image, we use 2-layer MLP with tanh activation to encode the state.

\textbf{Critic Network $V$}:
For image-based states, the output of the state encoder $f_\omega$ is passed through a linear layer to result in the value function of the state. This is done to share the convolutional layers between the actor and critic.
For non-image states, we use 2-layer MLP with tanh activation, followed by a linear layer to get the state's value.

\textbf{Utility Function $f_\nu$}:
Each available action's representation $c$ is passed through a linear layer and then concatenated with the output of the state encoder. This vector is fed into a 2-layer MLP with ReLU activation to output a single logit for each action. The logits of all the available actions are then stacked and input to a Categorical distribution. This acts as the policy's output and is used to sample actions, compute log probabilities, and entropy values.

\textbf{Auxiliary Policy $f_\chi$}:
The output of the state encoder is also separately used to compute auxiliary action outputs.
For CREATE and Shape Stacking, we have a 2D position action in $[-1, 1]$. For such constrained action space, we use a Beta distribution whose $\alpha$ and $\beta$ are computed using linear layers over the state encoding. Concretely, $\alpha = 1 + \text{softplus} (\text{fc}_\alpha(f_\omega(s))$ and $\beta = 1 + \text{softplus} (\text{fc}_\beta(f_\omega(s))$, to ensure their values lie in [1, $\infty$]. This in turn ensures that the Beta distribution is unimodal with values constrained in [0,1] (as done in~\citep{chou2017improving}), which we then convert to [-1,1].
The Shape Stacking environment also has a binary termination action for the agent. This is implemented by passing the state encoding through a linear layer which outputs two logits (for continuation/termination) of a Categorical distribution. 
The auxiliary action distributions are combined with the main discrete action Categorical distribution from $f_\nu$. This overall distribution is used to sample hybrid actions, compute log probabilities, and entropy values. Note, the entropy value of the Beta distribution is multiplied by a scaling factor of 0.1, for better convergence.

\end{document}